\pdfoutput=1

\documentclass[11pt]{article}

\usepackage[preprint]{acl}

\usepackage{times}
\usepackage{latexsym}
\usepackage{makecell}
\usepackage{booktabs}
\usepackage{multirow}
\usepackage{multicol}
\usepackage{amssymb}
\usepackage{pifont}
\usepackage{tcolorbox}
\usepackage[table, dvipsnames]{xcolor}
\usepackage{enumitem}

\newcommand{\cmark}{\ding{51}}%
\newcommand{\xmark}{\ding{55}}%
\usepackage[T1]{fontenc}

\usepackage[utf8]{inputenc}
\usepackage{bbding}
\usepackage{siunitx}

\definecolor{lightcoral}{rgb}{0.94, 0.5, 0.5}
\definecolor{lightgreen}{rgb}{0.56, 0.93, 0.56}

\usepackage{microtype}

\usepackage{inconsolata}

\usepackage{graphicx}

\title{Investigating Language and Retrieval Bias in Multilingual Previously~Fact-Checked Claim Detection
}

\author{
 \textbf{Ivan Vykopal\textsuperscript{\textdagger,1,2}},
 \textbf{Antonia Karamolegkou\textsuperscript{\textdagger,3}},
 \textbf{Jaroslav Kopčan\textsuperscript{2}},
 \textbf{Qiwei Peng\textsuperscript{3}},
\\
 \textbf{Tomáš Javůrek\textsuperscript{2}},
 \textbf{Michal Gregor\textsuperscript{2}} and
 \textbf{Marián Šimko\textsuperscript{2}}
\\
 \textsuperscript{1}Brno University of Technology,
 \\
 \textsuperscript{2}Kempelen Institute of Intelligent Technologies,
 \\
 \textsuperscript{3}University of Copenhagen,
 \\
 \textsuperscript{\textdagger}Contributed equally \\
\\
}

\begin{document}
\maketitle
\begin{abstract}
Multilingual Large Language Models (LLMs) offer powerful capabilities for cross-lingual fact-checking. However, these models often exhibit \textit{language bias}, performing disproportionately better on high-resource languages such as English than on low-resource counterparts. We also present and inspect a novel concept - \textit{retrieval bias}, when information retrieval systems tend to favor certain information over others, leaving the retrieval process skewed.  In this paper, we study language and retrieval bias in the context of Previously Fact-Checked Claim Detection (PFCD). We evaluate six open-source multilingual LLMs across 20 languages using a fully multilingual prompting strategy, leveraging the AMC-16K dataset. By translating task prompts into each language, we uncover disparities in monolingual and cross-lingual performance and identify key trends based on model family, size, and prompting strategy. Our findings highlight persistent bias in LLM behavior and offer recommendations for improving equity in multilingual fact-checking. To investigate retrieval bias, we employed multilingual embedding models and look into the frequency of retrieved claims. Our analysis reveals that certain claims are retrieved disproportionately across different posts, leading to inflated retrieval performance for popular claims while under-representing less common ones.
\end{abstract}






\section{Introduction}

Recent advances in multilingual Large Language Models (LLMs) have enabled powerful capabilities in multilingual and cross-lingual natural language understanding. Recent models support reasoning and generation across dozens or even hundreds of languages, powering applications ranging from machine translation to multilingual retrieval. Among these, fact-checking applications have benefited significantly: LLMs are increasingly used to identify claims, verify them, or even identify whether a given claim has already been verified~\cite{chen2023combatingmisinformationagellms, vykopal2024generativelargelanguagemodels}.

Despite these advances, multilingual LLMs exhibit persistent and often severe disparities in performance across different languages. This phenomenon, referred to as \emph{language bias}, occurs when models systematically favor high-resource languages—such as English, Chinese, or Spanish—over underrepresented ones, leading to uneven model behavior and outcomes \citep{hu2024survey}. Language bias is especially problematic in fact-checking contexts, where accurate performance in low-resource languages is critical to equitable access to trustworthy information. These disparities are rooted in the pretraining data, where high-resource languages dominate, and are amplified during model alignment and instruction tuning \citep{yang-etal-2024-language-bias}.

\begin{figure}
    \centering
    \includegraphics[width=\linewidth]{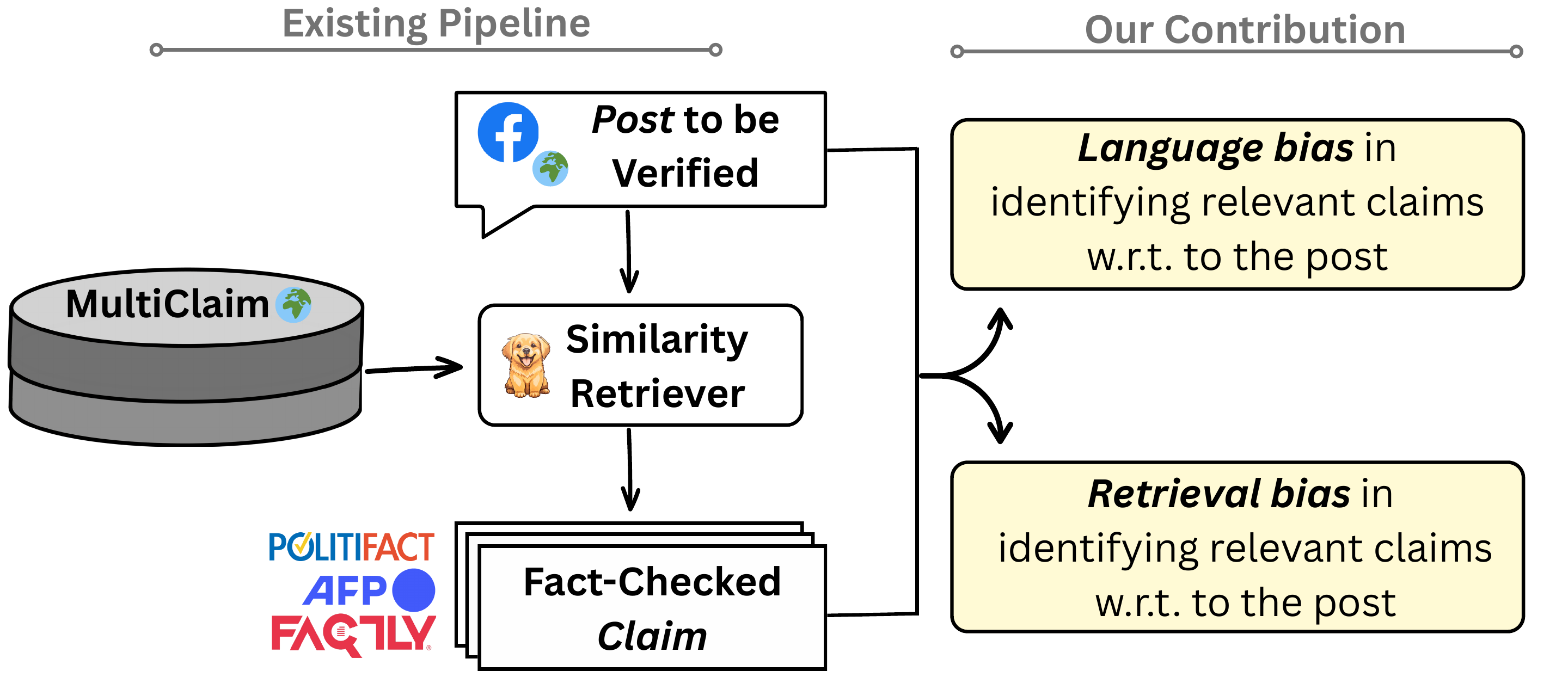}
    \caption{Our contribution and suggested analysis in the existing PFCD pipeline.}
\label{fig:overal-pipeline}
\end{figure}

In addition to language bias, we focus on what we call \emph{retrieval bias}. Whereas prior work on factuality in LLMs typically refers to the correctness of generated content, we define retrieval bias as the tendency of a model to retrieve certain fact-checked claims more frequently than others, independent of their actual relevance. In PFCD, this means the model might systematically favor or recall a subset of popular or easily phrased debunks, regardless of whether they truly match the input post. Retrieval bias can arise from differences in phrasing, claim popularity, or language-induced artifacts: for example, claims that appear more often in training data or have generic wording may be retrieved more often. This phenomenon is distinct from generation accuracy and has not been explicitly characterized in previous work. 

In this paper, we systematically investigate both language bias and retrieval bias in multilingual PFCD. We build upon the \textit{\textbf{AMC-16K}} dataset~\cite{vykopal2025largelanguagemodelsmultilingual}, a subset of \textit{\textbf{MultiClaim}} \citep{pikuliak2023multiclaim}, which contains 16,000 annotated pairs of social media posts and corresponding fact-checked claims in 20 languages, covering both monolingual and cross-lingual scenarios. A key novelty of our study is the use of \textit{fully multilingual prompting}: all prompts and task instructions are translated into the respective input languages, enabling direct comparison with English-only prompting strategies. Our contribution and the analysis are shown in Figure~\ref{fig:overal-pipeline}.
We assess six state-of-the-art open-source LLMs (\texttt{Qwen3 ${8B}$} and \texttt{$14B$}, \texttt{Llama3.1 ${8B}$} and \texttt{$70B$}, \texttt{Gemma3 ${12B}$} and \texttt{$27B$}) on \textit{\textbf{AMC-16K}}. Our experiments use various prompting strategies (zero-shot, zero-shot with task description, few-shot, and chain-of-thought) to measure their robustness.\footnote{Code and data are available at: \url{https://github.com/kinit-sk/llms-biases}.} 

We find that (1) most models exhibit significant language bias: performance often drops on non-English inputs and in languages less represented during training, consistent with \citet{yang-etal-2024-language-bias} and \citet{huang-etal-2023-languages}, and (2) models also exhibit retrieval bias: certain claims are retrieved far more often, typically those with simpler phrasing or higher prevalence, indicating that the models are influenced by claim distribution as well as relevance. These findings highlight the need to account for both language and retrieval biases when developing and evaluating multilingual PFCD systems.

\section{Background}

\subsection{Fact-Checked Claim Detection}


The task of retrieving previously fact-checked claims (PFCD) was introduced by \citet{shaar-etal-2020-known}, who developed the first benchmark and demonstrated its utility for reducing redundant verification efforts. Since then, research has expanded PFCD to multilingual settings with researchers providing new data sources \citep{kazemi-etal-2021-claim} or organizing shared tasks with multilingual tracks \citep{nakov-etal-2022-overview, semeval2025task7}. Recently, \citet{pikuliak-etal-2023-multilingual} introduced MultiClaim, a large multilingual dataset covering 39 languages. Building upon the MultiClaim, \citet{vykopal2024generativelargelanguagemodels} released \textit{AMC-16K}, a curated subset with 16K claim–post pairs across 20 languages. PFCD systems have evolved from embedding-based ranking to entailment-style models \citep{shaar-etal-2022-assisting} and, more recently, LLM-based generation or reranking \citep{zheng-etal-2024-evidence, pisarevskaya-zubiaga-2025-zero}. However, most studies still focus on English or a few high-resource languages, leaving performance disparities largely unexplored.

\subsection{Language Bias}

Language bias in multilingual NLP refers to systematic disparities in model performance across languages. \citet{hu2024survey} provide a comprehensive review of such disparities, noting that LLMs consistently favor high-resource languages due to imbalanced training corpora and optimization strategies. Empirical work by \citet{yang-etal-2024-language-bias} shows that retrieval success, classification, and summarization tasks all suffer from reduced performance in low-resource languages. Several mitigation strategies have been proposed, including balanced pretraining \citep{liu-etal-2020-multilingual}, multilingual alignment \citep{conneau2020unsupervised}, and synthetic data augmentation \citep{hedderich-etal-2021-survey}. Despite this, most multilingual models continue to perform best in English.

\subsection{Retrieval Bias}

We introduce the term retrieval bias to describe a tendency of claim-matching models to favor certain fact-checked claims over others, leading to a retrieval bias in PFCD. In our context, retrieval bias manifests when a model systematically retrieves some claims (e.g., those with high semantic similarity, profile topics, or generic phrasing) more often. For example, if a model frequently matches unrelated posts to a well-known debunk about vaccine safety simply because that debunk appears often in data or is phrased generically, this indicates retrieval bias. Such biases can arise from the uneven distribution of claims in the training or retrieval corpus: popular or frequently paraphrased claims are more “accessible” to the model. They may also result from language-driven effects if certain claims translate more naturally into the prompt language. To our knowledge, this specific bias has not been explicitly defined in PFCD research, so we formalize it here. We will analyze retrieval bias by measuring the frequency with which each claim is retrieved (or scored as relevant) by the model, controlling for actual ground-truth relevance. If some claims are chosen disproportionately often across different posts, this signals retrieval bias. Understanding this bias is important because it can inflate apparent performance if models “over-retrieve” popular claims, and it can also highlight blind spots for less common claims. Retrieval bias is conceptually related to opularity bias in recommender systems (RS), where algorithms tend to favor already popular items (e.g., widely streamed songs or frequently purchased products) at the expense of long-tail items \citep{Klimashevskaia2023survey}. In both cases, system outputs become skewed toward certain content, inflating apparent performance while masking blind spots for rarer or more specific instances. Whereas RS research has developed formal metrics and mitigation strategies for popularity bias \citep{Abdollahpouri2021UserCentered}, retrieval bias in PFCD has not been explicitly defined or systematically studied.




\section{Language Bias}
\label{sec:langauge-bias}

To systematically examine the language bias, we evaluate the behavior of multilingual LLMs on the task of detecting previously fact-checked claims. Our analysis considers both monolingual and cross-lingual scenarios across 20 languages, enabling us to explore how LLMs address linguistic diversity.

\begin{figure}
    \centering
    \includegraphics[width=\linewidth]{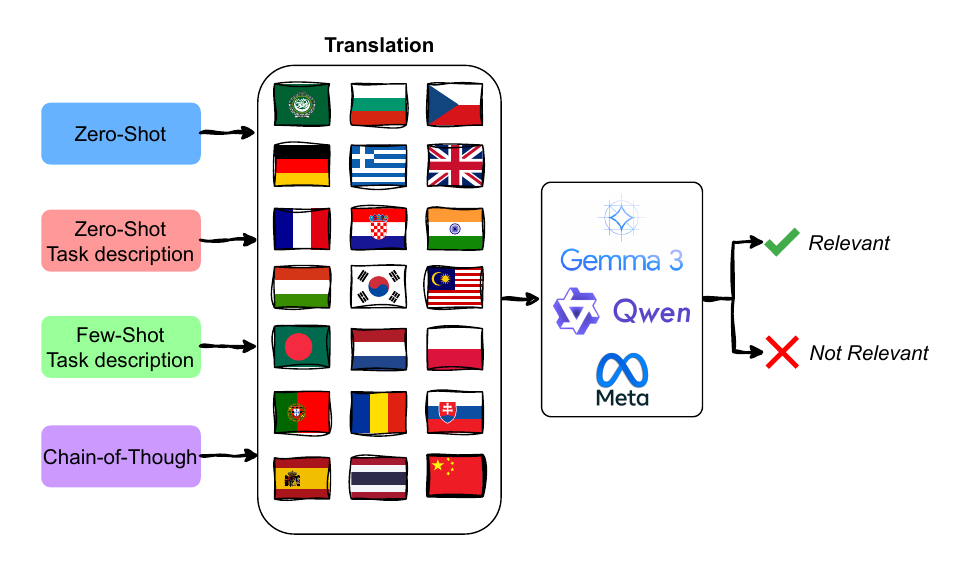}
    \caption{A pipeline for language bias experiments. First, we translate instructions into 21 languages and then prompt LLMs with translated instructions on the entire dataset using all four prompting strategies.}
    \label{fig:llm-biases}
\end{figure}

\subsection{Methodology}

Our experimental design builds upon the methodology introduced by~\citet{vykopal2025largelanguagemodelsmultilingual}, who evaluated seven multilingual LLMs across five prompting strategies in both monolingual and cross-lingual settings. However, their experiments relied only on English-language instructions, regardless of the language used in posts or fact-checked claims. To address this limitation, we extend their work by translating prompt instructions into the 20 languages represented in the dataset (see Figure~\ref{fig:llm-biases}). This allows us to asses how multilingual prompting impact model performance and to analyze language biases in LLM behavior.

\paragraph{Dataset.}

We employed the \textbf{\textit{AMC-16K}} dataset~\cite{vykopal2025largelanguagemodelsmultilingual} as the basis for our experiments. The dataset is a subset of the \textbf{\textit{MultiClaim}} dataset~\cite{pikuliak-etal-2023-multilingual}, consisting of 16,000 pairs of social media posts and corresponding fact-checked claims. These pairs are divided into two settings, monolingual and cross-lingual. In the \textit{monolingual setting}, the post and the fact-checked claim are in the same language, with 8K samples spanning 20 languages. In the \textit{cross-lingual setting}, the post and the fact-checked claim are in different languages, comprising another 8K pairs that cover 20 diverse language combinations.

\paragraph{Large Language Models.}

We selected six multilingual LLMs from three model families: \texttt{Qwen3}, \texttt{Llama3.1} and \texttt{Gemma3}. Each model family is represented by two variants with a different model size. All selected models are open-source, similarly to~\citet{vykopal2025largelanguagemodelsmultilingual}, to ensure the reproducibility of our experiments. Table~\ref{tab:models} provides details on the number of parameters, supported languages and whether we used "thinking" mode for a specific model. The "thinking" mode represents a reasoning-enhanced inference that is offered by \texttt{Qwen3} models. However, for our experiments, we leveraged the "thinking" mode, only with the smallest \texttt{Qwen3} model, due to computational reasons. By experimenting with "thinking" mode, we aim to assess whether enabling internal reasoning has an effect on the performance and the bias in LLMs.

\begin{table}
\resizebox{\columnwidth}{!}{%
\begin{tabular}{llrcl}
\toprule
\textbf{Model} & \textbf{\# Params} & \multicolumn{1}{l}{\textbf{\# Langs}} & \textbf{Thinking} & \textbf{Citation} \\
\midrule
\texttt{Qwen3} & 8B & 100 & \checkmark & \citet{yang2025qwen3technicalreport} \\ 
\texttt{Qwen3} & 14B & 100 & \xmark & \citet{yang2025qwen3technicalreport} \\
\texttt{Llama3.1 Instruct} & \makecell[l]{8B\\70B} & 8 & \xmark & \citet{grattafiori2024llama3herdmodels} \\
\texttt{Gemma3 IT} & \makecell[l]{12B\\27B} & 130 & \xmark & \citet{gemmateam2025gemma3technicalreport} \\
\bottomrule
\end{tabular}
}
\caption{Overview of multilingual LLMs used in our experiments.}
\label{tab:models}
\end{table}

\paragraph{Prompting Strategies.}

To examine how different prompting strategies affect LLMs' performance across languages, we adopted techniques introduced by~\citet{vykopal2025largelanguagemodelsmultilingual}. We explore four main approaches: \textit{(1) zero-shot}; \textit{(2) zero-shot with task description}; \textit{(3) few-shot with task description}; and \textit{(4) chain-of-thought}. In \textbf{zero-shot prompting}, the LLM receives only a post and a fact-checked claim without any task explanation, with a simple question to determine the relevance. \textbf{Zero-Shot with task description} adds a brief instruction to clarify and describe the task. \textbf{Few-shot with task descriptions} supplements this with 10 labeled demonstrations to guide the LLM. Finally, \textbf{chain-of-though (CoT)} prompting encourages step-by-step reasoning before producing a decision.

We excluded the \textbf{crosslingual-thought prompting (XLT)}~\cite{huang-etal-2023-languages} strategy used in the original study, as it instructs LLMs to translate both the post and the fact-checked claim into English. Our aim is to assess the model's multilingual abilities without translation as an intermediate step, which could mask language-specific biases.

To support fair multilingual evaluation, we translated all prompt templates and task descriptions into 19 non-English languages present in the \textbf{\textit{AMC-16K}} dataset and Chinese, resulting in instructions across 21 languages, while the data cover 20. The translations were generated using the GPT-4.1 API\footnote{https://openai.com/api/} and then manually reviewed by native or proficient speakers to ensure syntactic accuracy and semantic fidelity.

\paragraph{Evaluation.}

To evaluate the capabilities of LLMs for detecting previously fact-checked claims, each LLM was instructed to produce a binary label, determining the relevance between social media posts and fact-checked claims. Given the class imbalance in the dataset, where only 16\% of the pairs are relevant, we use the \textit{Macro F1} as the main evaluation metric, as it balances performance across both relevant and irrelevant classes.

To further analyze LLM's behavior, similarly to~\citet{vykopal2025largelanguagemodelsmultilingual}, we also report the \textit{True Negative Rate (TNR)}, which captures how well the LLM identifies irrelevant pairs, and the \textit{False Negative Rate (FNR)}, which reflects the frequency of missed relevant pairs.

\subsection{Experiments and Results}

In this section, we present findings on language bias in LLMs for the task of detecting previously fact-checked claims. We evaluated model performance in two experimental setups: \textit{monolingual} and \textit{cross-lingual}. For each setting, we report and discuss key observations based on the results.

\begin{table*}[]
\resizebox{\textwidth}{!}{%
\small
\begin{tabular}{l||rrrr|r||rrrr|r||rrrr|r}
\toprule
 & \multicolumn{5}{c||}{\textbf{Monolingual}} & \multicolumn{5}{c||}{\textbf{Cross-lingual - post-language}} & \multicolumn{5}{c}{\textbf{Cross-lingual - claim-language}} \\ \midrule
\textbf{Model} & \multicolumn{1}{c}{\textbf{ZS}} & \multicolumn{1}{c}{\textbf{\begin{tabular}[c]{@{}c@{}}ZS\\ + Task\end{tabular}}} & \multicolumn{1}{c}{\textbf{\begin{tabular}[c]{@{}c@{}}FS\\ + Task\end{tabular}}} & \multicolumn{1}{c|}{\textbf{CoT}} & \multicolumn{1}{c||}{\textbf{Avg.}} & \multicolumn{1}{c}{\textbf{ZS}} & \multicolumn{1}{c}{\textbf{\begin{tabular}[c]{@{}c@{}}ZS\\ + Task\end{tabular}}} & \multicolumn{1}{c}{\textbf{\begin{tabular}[c]{@{}c@{}}FS\\ + Task\end{tabular}}} & \multicolumn{1}{c|}{\textbf{CoT}} & \multicolumn{1}{c||}{\textbf{Avg.}} & \multicolumn{1}{c}{\textbf{ZS}} & \multicolumn{1}{c}{\textbf{\begin{tabular}[c]{@{}c@{}}ZS\\ + Task\end{tabular}}} & \multicolumn{1}{c}{\textbf{\begin{tabular}[c]{@{}c@{}}FS\\ + Task\end{tabular}}} & \multicolumn{1}{c|}{\textbf{CoT}} & \multicolumn{1}{c}{\textbf{Avg.}} \\ \midrule
\multirow{2}{*}{\texttt{Qwen3 8B}} & 69.70 & 77.81 & 70.70 & 83.07 & 75.32 & 60.25 & 70.40 & 63.84 & 76.65 & 67.79 & 60.25 & 70.40 & 63.84 & 76.65 & 67.79 \\
& \cellcolor{lightcoral}{-6.01} & \cellcolor{lightcoral}{-1.50} & \cellcolor{lightgreen}{0.76} & \multicolumn{1}{r|}{\cellcolor{lightcoral}{-6.31}} & \cellcolor{lightcoral}{-3.26} & \cellcolor{lightcoral}{-4.97} & \cellcolor{lightcoral}{-1.75} & \cellcolor{lightgreen}{0.51} & \cellcolor{lightcoral}{-8.52} & \cellcolor{lightcoral}{-3.68} & \cellcolor{lightcoral}{-1.42} & \cellcolor{lightcoral}{-2.40} & \cellcolor{lightgreen}{0.98} & \cellcolor{lightcoral}{-4.28} & \cellcolor{lightcoral}{-1.78} \\
\midrule
\multirow{2}{*}{\texttt{Qwen3 14B}} & 73.40 & 77.08 & 74.72 & \textbf{85.15} & 77.59 & 64.12 & 70.99 & 69.29 & \textbf{78.24} & 70.66 & 64.12 & 70.99 & 69.29 & \textbf{78.24} & 70.66 \\
 & \cellcolor{lightcoral}{-9.23} & \cellcolor{lightgreen}{3.19} & \cellcolor{lightgreen}{0.39} & \multicolumn{1}{r|}{\cellcolor{lightcoral}{-4.82}} & \cellcolor{lightcoral}{-2.62} & \cellcolor{lightcoral}{-7.67} & \cellcolor{lightcoral}{-0.30} & \cellcolor{lightgreen}{0.58} & \cellcolor{lightcoral}{-5.69} & \cellcolor{lightcoral}{-3.27} & \cellcolor{lightcoral}{-4.94} & \cellcolor{lightgreen}{0.81} & \cellcolor{lightcoral}{-0.38} & \cellcolor{lightcoral}{-3.11} & \cellcolor{lightcoral}{-1.91} \\
\midrule
\multirow{2}{*}{\texttt{Gemma3 12B}} & 56.44 & 58.72 & 70.07 & 58.90 & 61.03 & 52.86 & 52.22 & 65.92 & 50.78 & 55.45 & 52.86 & 52.22 & 65.92 & 50.78 & 55.45 \\
& \cellcolor{lightcoral}{-3.67} & \cellcolor{lightgreen}{6.36} & \cellcolor{lightgreen}{0.26} & \multicolumn{1}{r|}{\cellcolor{lightgreen}{1.99}} & \cellcolor{lightgreen}{1.23} & \cellcolor{lightcoral}{-4.91} & \cellcolor{lightgreen}{3.18} & \cellcolor{lightgreen}{0.90} & \cellcolor{lightgreen}{0.42} & \cellcolor{lightcoral}{-0.10} & \cellcolor{lightcoral}{-2.08} & \cellcolor{lightgreen}{2.60} & \cellcolor{lightcoral}{-1.31} & \cellcolor{lightgreen}{0.85} & \cellcolor{lightgreen}{0.02} \\
\midrule
\multirow{2}{*}{\texttt{Gemma3 27B}} & 59.35 & 55.54 & 72.72 & 52.41 & 60.00 & 53.69 & 49.54 & 64.94 & 46.37 & 53.64 & 53.69 & 49.54 & 64.94 & 46.37 & 53.64 \\
& \cellcolor{lightgreen}{1.64} & \cellcolor{lightgreen}{10.85} & \cellcolor{lightgreen}{0.83} & \multicolumn{1}{r|}{\cellcolor{lightgreen}{5.60}} & \cellcolor{lightgreen}{4.73} & \cellcolor{lightcoral}{-3.41} & \cellcolor{lightgreen}{3.61} & \cellcolor{lightcoral}{-1.12} & \cellcolor{lightgreen}{3.24} & \cellcolor{lightgreen}{0.58} & \cellcolor{lightgreen}{0.50} & \cellcolor{lightgreen}{4.16} & \cellcolor{lightgreen}{0.83} & \cellcolor{lightgreen}{2.09} & \cellcolor{lightgreen}{1.90} \\
\midrule
\multirow{2}{*}{\texttt{Llama3.1 8B}} & 50.46 & 71.13 & 51.12 & 67.18 & 59.97 & 45.53 & 66.48 & 50.56 & 58.82 & 55.35 & 45.53 & 66.48 & 50.56 & 58.82 & 55.35 \\
& \cellcolor{lightcoral}{-1.55} & \cellcolor{lightcoral}{-7.20} & \cellcolor{lightgreen}{4.41} & \multicolumn{1}{r|}{\cellcolor{lightcoral}{-2.34}} & \cellcolor{lightcoral}{-1.67} & \cellcolor{lightcoral}{-2.65} & \cellcolor{lightcoral}{-5.72} & \cellcolor{lightgreen}{3.97} & \cellcolor{lightcoral}{-3.98} & \cellcolor{lightcoral}{-2.10} & \cellcolor{lightgreen}{0.12} & \cellcolor{lightcoral}{-1.31} & \cellcolor{lightgreen}{1.92} & \cellcolor{lightgreen}{1.05} & \cellcolor{lightgreen}{0.45} \\
\midrule
\multirow{2}{*}{\texttt{Llama3.1 70B}} & 73.97 & 74.38 & 54.04 & 75.48 & 69.47 & 65.57 & 65.77 & 54.34 & 67.66 & 63.33& 65.57 & 65.77 & 54.34 & 67.66 & 63.33 \\
& \cellcolor{lightcoral}{-5.18} & \cellcolor{lightgreen}{0.48} & \cellcolor{lightgreen}{0.67} & \multicolumn{1}{r|}{\cellcolor{lightcoral}{-1.98}} & \cellcolor{lightcoral}{-1.50} & \cellcolor{lightgreen}{0.45} & \cellcolor{lightgreen}{3.00} & \cellcolor{lightcoral}{-2.82} & \cellcolor{lightcoral}{-1.88} & \cellcolor{lightcoral}{-0.31} & \cellcolor{lightcoral}{-1.42} & \cellcolor{lightgreen}{1.46} & \cellcolor{lightcoral}{-0.74} & \cellcolor{lightcoral}{-2.37} & \cellcolor{lightcoral}{-0.77} \\ \midrule
\multirow{2}{*}{\textit{Average}} & 63.89 & 69.11 & 65.56 & 70.67 & 67.23 & 57.00 & 62.57 & 61.48 & 63.09 & 61.04 & 57.00 & 62.57 & 61.48 & 63.09 & 61.04 \\
& \cellcolor{lightcoral}{-4.00} & \cellcolor{lightgreen}{2.03} & \cellcolor{lightgreen}{1.22} & \multicolumn{1}{r|}{\cellcolor{lightcoral}{-1.31}} & \cellcolor{lightcoral}{-0.51} & \cellcolor{lightcoral}{-3.86} & \multicolumn{1}{r}{\cellcolor{lightgreen}{0.34}} & \multicolumn{1}{r}{\cellcolor{lightgreen}{0.34}} & \multicolumn{1}{r|}{\cellcolor{lightcoral}{-2.74}} & \cellcolor{lightcoral}{-1.48} & \multicolumn{1}{r}{\cellcolor{lightcoral}{-1.54}} & \multicolumn{1}{r}{\cellcolor{lightgreen}{0.89}} & \multicolumn{1}{r}{\cellcolor{lightgreen}{0.22}} & \multicolumn{1}{r|}{\cellcolor{lightcoral}{-0.96}} & \cellcolor{lightcoral}{-0.35} \\ \midrule
\multirow{2}{*}{\makecell[l]{\texttt{Qwen3 8B}\\\texttt{(thinking)}}} & \textbf{77.19} & \textbf{82.74} & \textbf{81.27} & - & \textbf{80.40} & \textbf{66.29} & \textbf{74.43} & \textbf{73.22} & - & \textbf{71.31} & \textbf{66.29} & \textbf{74.43} & \textbf{73.22} & - & \textbf{71.31} \\
& \cellcolor{lightcoral}{-2.50} & \cellcolor{lightcoral}{-1.25} & \cellcolor{lightcoral}{-0.96} & \multicolumn{1}{r|}{-} & \cellcolor{lightcoral}{-0.93} & \cellcolor{lightcoral}{-2.28} & \cellcolor{lightcoral}{-1.45} & \cellcolor{lightcoral}{-0.15} & - & \cellcolor{lightcoral}{-1.29} & \cellcolor{lightcoral}{-1.09} & \cellcolor{lightcoral}{-2.08} & \cellcolor{lightgreen}{0.21} & - & \cellcolor{lightcoral}{-0.99} \\ \bottomrule
\end{tabular}}
\setlength{\fboxsep}{1pt}
\caption{Macro F1 performance across 20 languages and language combinations for multiple models in three settings: monolingual, cross-lingual with post-language instructions, and cross-lingual with claim-language instructions. The first row per model shows absolute performance with English instructions; the second row shows the difference when using the target language for instruction. ZS = zero-shot; FS = few-shot; Task = using Task description. \colorbox{lightgreen}{Green} cells indicate improvements; \colorbox{lightcoral}{red} cells indicate declines. The highest absolute score for each column is in \textbf{bold}.}
\label{tab:overall-results}
\end{table*}

\paragraph{Monolingual Settings.}
In Table~\ref{tab:overall-results}, \textit{Monolingual} column shows the average performance differences when using target-language instruction and English instructions, averaged across 20 languages for each model and prompting strategies. 

\texttt{Gemma3 27B} consistently preferred target-language instructions across all prompting strategies. This LLM showed positive performance differences in all settings when the instruction was provided in the target languages, suggesting its strong multilingual capabilities.

Prompting strategies that include task description or few-shot examples reduce language bias. Across most LLMs, using target-language instructions in zero-shot settings resulted in strongly negative performance differences, indicating a bias towards English. However, adding task-specific context narrowed these differences, moving performance closer to that of the English baseline, but still achieving a positive increase towards target-language instructions. This reveals that enhanced prompting helps mitigate language-specific performance gaps and reduces bias.

\texttt{Qwen} models showed strong language bias in zero-shot and CoT prompting. Both \texttt{Qwen3 8B} and \texttt{Qwen3 14B} had large negative differences when using the target language for instructions in zero-shot and CoT settings, indicating a bias toward English instructions. These large deviations suggest that the \texttt{Qwen} models are less effective at processing multilingual prompts unless additional context or demonstrations are provided.

\paragraph{Cross-Lingual Settings.} 

In the cross-lingual settings (see Table~\ref{tab:overall-results}, cross-lingual columns), we compared the post-language and claim-language instruction against the English-language instruction. For post-language instruction, we investigate the difference between using the language of the post for the instruction compared to English instruction, while in claim-language instruction, we are using the language of the fact-checked claim.

For the post-language setup, we found that few-shot prompting combined with a task description reduced language bias and led to positive performance gains across most LLMs. This suggests that instruction clarity, especially when combined with the same language as the input post, can improve model performance and steer the language bias. In addition, \texttt{Gemma3} models showed strong robustness to instruction language, with consistently improved performance when using the post language in all strategies except zero-shot. 

In the claim-language instruction setting, we observed that zero-shot prompting generally led to a negative impact on performance, indicating a bias toward English when no additional context was provided. While adding a task description or a few-shot examples helped mitigate this effect, the resulting improvements were maller than those observed in the post-language setup. These findings suggest that, compared to post-language prompts, using the claim language for instructions offers less benefit overall, particularly in low-context settings.

\paragraph{Thinking.}

In our experiments, we also considered "thinking mode" for the \texttt{Qwen3 8B} model and evaluated whether thinking affects language bias. We found that enabling the thinking mode in \texttt{Qwen3 8B} did not improve performance when using target-language instructions in a monolingual setting. In general, the thinking mode benefited the English-language instructions, suggesting a potential bias in how reasoning capabilities are tuned across languages. However, in a zero-shot setting, the average performance decrease was less severe when thinking was enabled (-6 without thinking vs. -2.5 with thinking), indicating a partial mitigation of the negative effect. 

For the cross-lingual, we observed that the thinking mode decreased the language bias in LLMs (average performance was decreased by around 1.8 and 2.9 times). When including the task description along with the demonstrations, the mitigation of the language bias was the highest, resulting only in marginal differences in Macro F1. Overall, these results demonstrate that activating thinking reduces the performance gap between English and target-language instructions, highlighting its potential for mitigating language bias.

\section{Retrieval Bias}
\label{sec:retrieval-bias}

We define retrieval bias as the model's tendency to retrieve certain fact-checked claims more frequently than others, regardless of their actual relevance to the post. To examine retrieval bias, we use both quantitative and qualitative methods. First measure retrieval success using standard metrics. Then, we analyze how frequently different claims are retrieved in the top-$K$ results to identify retrieval biases. Finally, we perform topic modeling to explore the main themes of these frequently retrieved claims.

\subsection{Methodology}

\begin{figure}[h]
    \centering
    \includegraphics[width=\linewidth]{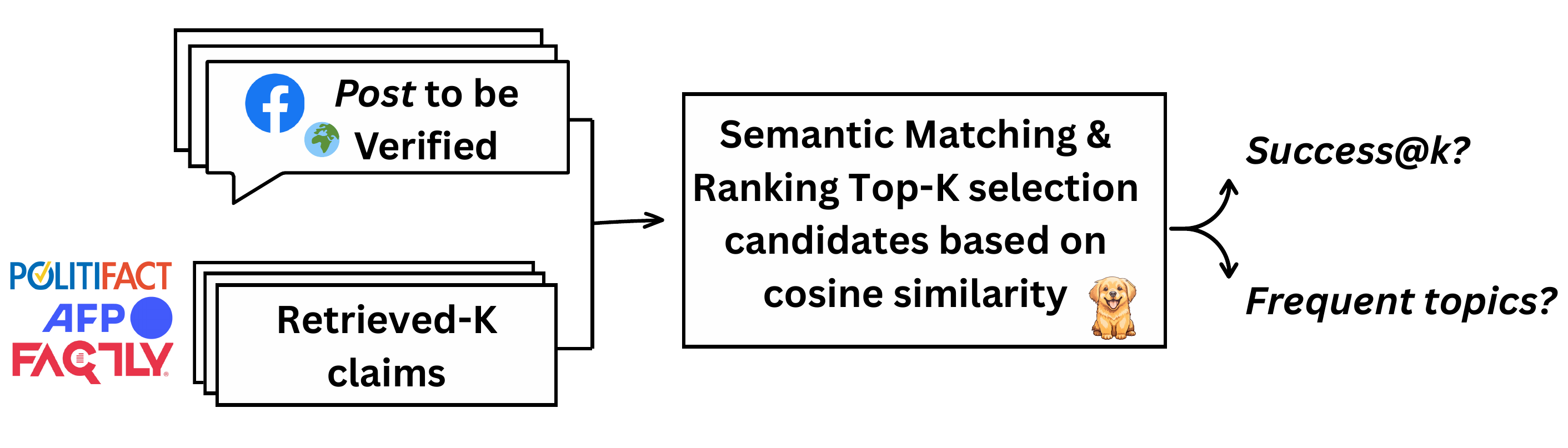}
    \caption{The retriever takes a post as a query, searches through a database of previously fact-checked claims to identify and rank the most relevant claims based on cosine similarity, returning top-K candidates. We then calculate the success of the retrieval and potentially frequently retrieved claims. 
    }
\label{fig:retrieval-biases}
\end{figure}

We follow \citep{pikuliak2023multiclaim} and frame the retrieval task as a semantic matching and ranking problem, where the utilized embedding model must rank all available claims for each given post. We base our experiments on the test split of the \textbf{\textit{MultiClaim}} dataset, and employ \texttt{T5} \citep{ni-etal-2022-large} and \texttt{Multilingual E5} \citep{wang2024multilingual} as embedding models. 
These models serve as our retrieval system by encoding both posts (queries) and claims (candidates) into dense vector representations, which are then ranked using cosine similarity scores. 
The split consists of 1,239 post-claim pairs. 
For each post, we compute claim embeddings, calculate cosine similarities with the post embedding, and rank by similarity to retrieve the top-K candidates, as depicted in Figure \ref{fig:retrieval-biases}.
To ensure computational efficiency while maintaining comprehensive coverage, we have limited the overall retrieval only to the top 20 claims for each post text.
In particular, we analyze two key aspects: first, the retrieval success (both quantitatively and qualitatively) and second, the retrieval topic distribution.


\textbf{Retrieval Success.} We begin by evaluating retrieval performance using \textit{Success@K}, \textit{Mean Average Precision (MAP)}, and \textit{Mean Reciprocal Rank (MRR)}. \textit{Success@K} measures the proportion of posts for which at least one gold-relevant claim appears within the top-$K$ retrieved candidates. \textit{MAP} captures the average precision across all posts, emphasizing the rank positions of relevant claims and providing a holistic view of ranking quality. \textit{MRR} focuses on the position of the first correct (gold-relevant) claim by averaging the reciprocal of its rank. These metrics are computed using the gold claim-post mapping, following the evaluation protocol of \citet{pikuliak2023multiclaim}. To further analyze retrieval quality and biases, we then examine the distribution of retrieved claims across posts to identify whether certain fact-checked claims are retrieved disproportionately often, regardless of their relevance. Finally, to better understand thematic patterns in these frequently retrieved claims, we apply topic modeling techniques, which reveal dominant topics and potential biases influencing retrieval behavior.

\textbf{Topic distribution.} 
To understand the thematic distribution of the retrieved claims and identify potential retrieval biases, we apply topic modeling on the claim contents. Specifically, we use \emph{Semantic Signal Separation (S³)} via the TurfTopic modeling framework~\cite{Kardos2025}, which leverages semantic embeddings rather than raw text. This allows for more nuanced topic discovery in multilingual and noisy claim data. We first encode the retrieved claim texts using the \texttt{paraphrase-multilingual-MiniLM-L12-v2} model from the SentenceTransformers library~\cite{reimers-gurevych-2019-sentence}. This model generates dense sentence embeddings capturing semantic similarity across languages and contexts. The embeddings are then input to the Semantic Signal Separation model, which decomposes the embedding space into a predefined number of topics (in our case, 30). The model is fit using a fit-transform procedure that returns topic assignment probabilities for each claim. We assign each claim to the topic with the highest probability, producing a topic distribution that reflects the major semantic themes in the retrieved data. This topic distribution enables qualitative analysis of frequent claim types and highlights potential thematic retrieval biases in the system.

\subsection{Experiments and Results}

In this section, we present findings from our retrieval bias analysis. Building on the methodology described above, we evaluate retrieval behavior across two dimensions.


\paragraph{Retrieval Success.} 



The results of retrieval success across the different metrics are presented in Table~\ref{tab:retrieval_models}. The \texttt{Multilingual E5} model slightly outperforms \texttt{T5} at Success@1 (38.8\% vs. 36.7\%) and Success@20 (88.5\% vs. 87.5\%), with marginal gains in MAP and MRR as well. The scores at Success@10 are consistent with previous works that use a subset of the \textbf{\textit{MultiClaim}} dataset \citep{pikuliak2023multiclaim,semeval2025task7}. This indicates that both models achieve comparable retrieval quality, with \texttt{Multilingual E5} providing a modest advantage in early retrieval effectiveness.

\begin{table}[h]
\centering
\resizebox{\columnwidth}{!}{%
\begin{tabular}{lcccccc}
\toprule
\textbf{Model} & \textbf{@1} & \textbf{@3} & \textbf{@5} & \textbf{@10} & \textbf{@20} & \textbf{MAP / MRR} \\
\midrule
\texttt{T5}           & 36.72\% & 64.41\% & 74.01\% & 83.21\% & 87.49\% & 51.23 / 52.43 \\
\texttt{Multilingual E5} & 38.80\% & 63.40\% & 74.40\% & 82.90\% & 88.50\% & 52.69 / 53.98 \\
\bottomrule
\end{tabular}%
}
\caption{Success@K and Mean Average Precision (MAP)/Mean Reciprocal Rank (MRR) for dense retrieval models on \textbf{\textit{MultiClaim}}.}
\label{tab:retrieval_models}
\end{table}

\begin{figure}
    \centering
    \includegraphics[width=0.8\linewidth]{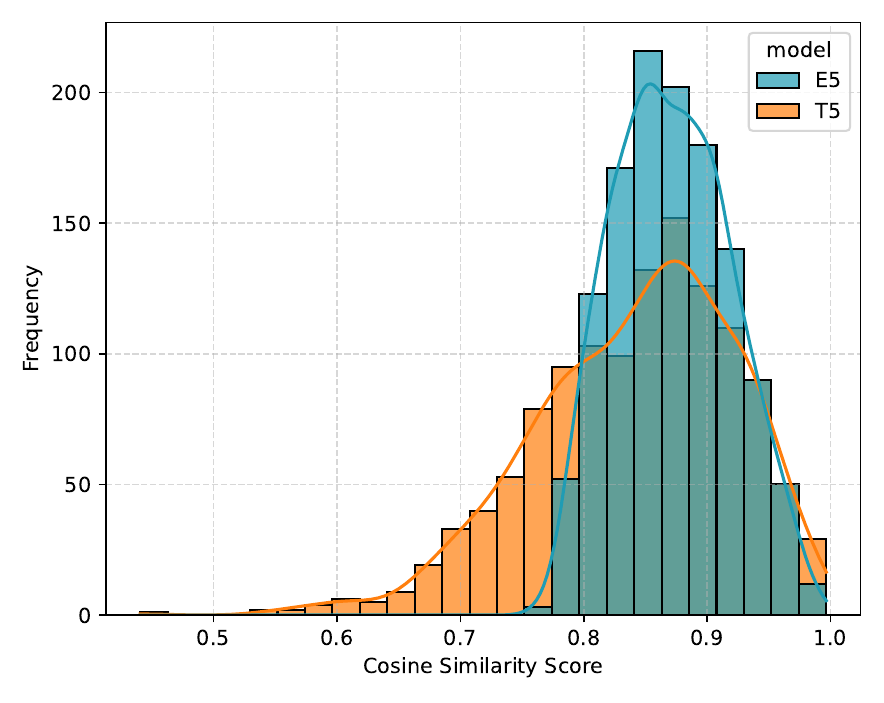}
    \caption{Distribution of Top-1 Cosine Similarity Scores.}
    \label{fig:cosine}
\end{figure}

To better understand the behavior of dense retrievers beyond aggregate performance metrics, we analyze the similarity scores used to rank candidate claims. Specifically, we examine the cosine similarity distributions of top-1 retrieved claims for both \texttt{T5} and \texttt{Multilingual E5}. This allows us to investigate whether the similarity scores are high for the top-1 retrieved claim as shown in Figure \ref{fig:cosine}. We can see a distribution between 0.8 and 1.0 for \texttt{Multilingual E5} and 0.6 and 1.0 for \texttt{T5}. This relatively narrow band of high scores can lead to confirmation-style matches, where retrieved fact-checks are "close enough" semantically but potentially off-topic or unable to verify the post. High similarity does not always imply relevance, indicating that embedding-based retrieval alone is insufficient.

Apart from a quantitative analysis, we also conduct a qualitative inspection of the most frequently retrieved top-1 claims by the \texttt{Multilingual E5} model. Table~\ref{tab:top1_claims} presents the ten claims that most often appeared in the top-1 retrieval position across posts. While the majority are well-formed factual statements, typically addressing political events or health-related misinformation, we also observe two noisy or low-information entries: a truncated JavaScript snippet and a Google Drive link. These cases highlight that, despite generally favoring semantically rich and interpretable claims at rank 1, the model can still surface non-claim-like artifacts. The presence of such items suggests potential vulnerabilities in embedding-based retrieval when encountering malformed inputs. For comparison, we report the top-20 most frequent retrieved claims (regardless of rank) in the Appendix Table \ref{tab:top20claims}, where such noisy entries are even more prominent.




\begin{table}[h]
\centering
\resizebox{\columnwidth}{!}{%
\small
\begin{tabular}{p{5.5cm} r r}
\toprule
\textbf{Claim (truncated)} & \textbf{Freq} & \textbf{CosSim} \\
\midrule
(function(d, s, id) \{ var js, fjs... & 12 & 0.811 \\
New Australian law passed in Parliament... & 6 & 0.904 \\
Trump tweets on South Africa’s unemployment... & 5 & 0.901 \\
https://drive.google.com/file/d/1TF... & 5 & 0.799 \\
South Korean protesters bombed with... & 5 & 0.909 \\
A message is being widely shared on... & 5 & 0.898 \\
Report quotes Duterte saying COVID-19... & 5 & 0.842 \\
Accurate representation of vaccinations... & 5 & 0.876 \\
Alexandre Trudeau, the brother of Canadian... & 4 & 0.946 \\
Pictures show Azerbaijan has destroyed... & 4 & 0.910 \\
\bottomrule
\end{tabular}}
\caption{Top-1 retrieved claims: truncated content with frequency (freq) and average cosine similarity score (CosSim).}
\label{tab:top1_claims}
\end{table}

Lastly, to further improve retrieval outcomes, we explore using LLMs to classify each claim–post pair as \textit{relevant} or \textit{not}. As LLMs do not return a ranked list but only binary predictions, we report \textbf{Top-1 relevance classification accuracy}—i.e., the fraction of posts for which the top-scoring LLM-predicted relevant claim aligns with the gold label. Results are shown in Table~\ref{tab:llm-retriev}. 

\begin{table}[h]
\centering
\resizebox{0.65\columnwidth}{!}{%
\begin{tabular}{lc}
\toprule
\textbf{Model} & \textbf{Top-1 Accuracy} \\
\midrule
\texttt{T5}                 & 36.72\%  \\
\texttt{Multilingual E5}    & 38.80\% \\
\texttt{Qwen3 8B}           & 63.12\%  \\
\texttt{Gemma3 12B}         & 42.80\%  \\
\texttt{Llama3.1 8B}        & 38.86\% \\
\bottomrule
\end{tabular}%
}
\caption{Top-1 relevance classification accuracy (or success@1) for LLMs and retrieval models, showing that LLMs can help improve the success of the retrieval. }
\label{tab:llm-retriev}
\end{table}

Incorporating LLM-based relevance filtering improves these scores, demonstrating the model's ability to mitigate retrieval bias by deprioritizing frequent but irrelevant claims.

\paragraph{Topic Analysis.}

The results of the topical analysis are presented in Table~\ref{tab:top_topics}, revealing that some claims are retrieved disproportionately often, indicating a bias toward specific themes. To investigate this skew, we applied topic modeling using \texttt{turftopic} on the retrieved claims. The top 10 topics include high-profile and globally salient issues such as COVID-19, Ukraine, and viral videos. These topics are both frequent in the dataset and more likely to be retrieved, often independent of the input post’s specific content. We further visualize these topic clusters using a t-SNE projection of the claim embeddings in the Appendix Figure~\ref{fig:topics}


\begin{table}[t]
\centering
\small
\resizebox{\columnwidth}{!}{%
\begin{tabular}{p{5cm}c}
\toprule
\textbf{Topic Keywords} & \textbf{Count} \\
\midrule
COVID, flu, PCR testing & 170 \\
Ukraine, Kyiv, Russia & 139 \\
Videos, footage, shows & 128 \\
Nigeria, Kenya, Africa & 118 \\
Kidnapping, jail, trafficking & 117 \\
Coronavirus, viral content & 114 \\
Bushfires, Australia, New Zealand & 110 \\
Assam, BJP, Taliban, Banerjee & 109 \\
Hydroxychloroquine, FDA, graphene & 107 \\
Beirut, Palestine, Israel, explosion & 106 \\
\bottomrule
\end{tabular}%
}
\caption{Top 10 most frequent topics among retrieved claims based on TurfTopic modeling.}
\label{tab:top_topics}
\end{table}

\section{Discussion}

Our analysis highlights two central sources of bias in PFCD: \textit{language bias} and \textit{retrieval bias}. These biases affect model performance in distinct but interconnected ways and have compounding effects on the fairness and effectiveness of multilingual fact-checking systems.

\paragraph{Language Bias and Its Impact.}

Language bias emerges as systematic performance disparities across languages, with models consistently favoring high-resource languages such as English. Despite using multilingual prompting as suggested by previous works, we observe that models like \texttt{Qwen3} and \texttt{Llama3.1} perform significantly worse on low-resource languages, particularly under zero-shot settings. We find that language bias can be partially mitigated through enhanced prompting strategies. Few-shot prompting and task descriptions significantly reduce performance gaps between English and target-language instructions. Moreover, enabling reasoning-enhanced inference (``thinking mode'') further narrows this gap, particularly in monolingual settings. This suggests that both contextual clarity and internal reasoning mechanisms help models better generalize across languages.

The implications of language bias are profound: users in underrepresented linguistic communities are less likely to benefit from accurate fact-checking, exacerbating global information inequities. Moreover, the performance gap persists even when semantic equivalence is maintained across translations, suggesting that model alignment and pretraining corpora remain skewed toward dominant languages.

\paragraph{Retrieval Bias and Its Impact.}

Retrieval bias refers to the model's tendency to favor certain claims during retrieval, independent of their actual relevance. Our analysis shows that some claims—especially those with generic phrasing and high topical prevalence are disproportionately retrieved. This is evident in the topic modeling results, where high-frequency topics (e.g., COVID-19, Ukraine, viral videos) dominate the top retrieved claims. We quantify retrieval bias using \textit{Success@K}, measuring the likelihood that a relevant claim appears among the top-$K$ retrieved results. Without LLM reranking, the success@1 is only 38.8\%, improving to 88.5\% at $K=20$. However, when we incorporate LLM-based relevance filtering, success@1 improves, indicating that LLMs can help mitigate retrieval bias by reranking based on semantic relevance rather than frequency or phrasing alone. This bias can lead to reduced topical coverage and overlooked relevant claims, especially for underrepresented narratives. As a result, the effectiveness of fact-check retrieval systems may be compromised in diverse real-world settings.


\paragraph{Interplay and Broader Implications.} Language and retrieval biases are not isolated phenomena. Language bias can amplify retrieval bias when certain claims are more accessible or better represented in specific languages. Conversely, retrieval bias can mask language bias by over-representing high-frequency claims that are easier to match across languages. Together, these biases challenge the equity and robustness of multilingual PFCD systems. Addressing them requires not only architectural and training improvements but also evaluation frameworks that explicitly account for both semantic relevance and linguistic diversity.




\section{Conclusion}
This study investigates language and retrieval biases in multilingual LLMs when applied to the task of PCFD. Through a systematic evaluation across 20 languages and six models, we show that most models perform best in English and suffer measurable drops in accuracy when operating in low-resource languages. Our results also suggest that the prompt language and prompting strategy influence performance, with few-shot prompting and translated task descriptions helping mitigate some of the bias. We also identify a distinct form of retrieval bias, where certain claims—often generic, high-profile, or frequently occurring—are retrieved disproportionately. This can distort evaluation metrics and limit the diversity of retrieved evidence. However, we find that LLM-based reranking improves semantic alignment and helps mitigate this effect.

We hope this analysis provides a foundation for more equitable multilingual fact-checking systems. Future directions could explore deeper linguistic analysis of failures, adaptive prompting strategies, and fairness-aware retrieval objectives. Beyond technical solutions, interdisciplinary approaches—such as sociolinguistic audits of retrieved claims—may offer new lenses on model behavior, revealing how cultural framing and linguistic nuance shape retrieval outcomes.


\section*{Limitations}

Despite covering 20 languages, our study remains limited by the scope of the \textbf{\textit{AMC-16K}} dataset and the selection of language pairs. Some regions and scripts (e.g., indigenous languages, South Asian scripts beyond Devanagari) are underrepresented. Additionally, our evaluation assumes task equivalence across languages, which may not fully capture cultural or linguistic nuances.

We also only test six open-source models, which may not reflect the capabilities of proprietary LLMs. Finally, while we apply multilingual prompting, we do not consider code-switching or mixed-language input, which may be more common in real-world scenarios.

\section*{Acknowledgments}

This research was partially supported by \textit{DisAI - Improving scientific excellence and creativity in combating disinformation with artificial intelligence and language technologies}, a project funded by Horizon Europe under \href{https://doi.org/10.3030/101079164}{GA No.101079164}, by the \textit{European Union NextGenerationEU} through the Recovery and Resilience Plan for Slovakia under the project No. 09I01-03-V04-00007. and by the Ministry of Education, Youth and Sports of the Czech Republic through the e-INFRA CZ (ID:90254).

Antonia Karamolegkou was supported by the Onassis Foundation - Scholarship ID: F ZP 017-2/2022-2023'.

\section*{Ethical Consideration}

\paragraph{Intended Use.} 

All experiments in this work were conducted using publicly available, research-focused datasets: \textbf{\textit{MultiClaim}}~\citep{pikuliak2023multiclaim} and its subset \textbf{\textit{AMC-16K}}~\citep{vykopal2025largelanguagemodelsmultilingual}. These datasets consist of social media posts and fact-checking information collected from public sources. Since both datasets are focused on the fact-checking domain, social media posts and fact-checked claims include harmful content. Our goal is to promote fairness and transparency in multilingual model evaluation, particularly for low-resource languages that are historically underrepresented in NLP systems.

\paragraph{Usage of AI Assistants.} We have used the AI assistant for grammar checks and sentence structure improvements. We have not used AI assistants in the research process beyond the experiments detailed in the Methodology in Section \ref{sec:langauge-bias} and \ref{sec:retrieval-bias}. 




\bibliography{anthology,custom}

\appendix

\section{Computational Resources}

For our experiments, we leveraged a computational infrastructure consisting of A40 PCIe 40GB and H100 NVL 94GB NVIDIA GPUs, while our experiments ran in parallel on multiple GPUs. In total, our experiments required approximately 3250 GPU hours.

\section{Language Bias}

In this section, we provide additional findings and results on the identification of language biases in LLMs. We extend the prompting setup by \citep{vykopal2025largelanguagemodelsmultilingual} and translate the prompts into different languages as described in the main paper. We provide the system prompt and zeroshot prompt template used in our experiments in Figures \ref{fig:prompts} below.

\begin{figure*}[tb!]
\begin{tcolorbox}[
    colback=white,
    colframe=brown
]
\begin{verbatim}
    
You are a fact-checker responsible for determining the relevance of previously 

debunked claims to a given social media post. Your task is to assess if the claim 

is relevant to the post and whether it is possible to infer main statements from 

the post from the debunked claim. Based on your analysis, provide one of the 

following answers:

- 'Yes' if the claim is relevant to the social media post.

- 'No' if the debunked claim is not relevant to the social media post.	

\end{verbatim}
\end{tcolorbox}
\begin{tcolorbox}[
    colback=white,
    colframe=blue
]
\begin{verbatim}
    
Claim: {document}
Post: {query}

Is the claim relevant to the social media post?  Respond with a single word, 

either ""Yes"" or ""No"", in English only.	

\end{verbatim}
\end{tcolorbox}

\caption{System prompt used for the zero-shot English setting. This prompt was translated into different languages using the GPT-4.1 API as detailed in the main paper.}
\label{fig:prompts}
\end{figure*}

\subsection{Experimental Setup}

For all experiments for the language biases, we used the same hyperparameter settings during inference. In particular, we set the temperature to 0 to ensure mostly deterministic outputs and to minimize variability due to sampling. However, for the experiments with thinking mode with the \texttt{Qwen3 8B}, we adopted the recommended configuration provided by the model's authors. Specifically, we set the temperature to 0.6, TopP to 0.95, TopK to 20, and MinP to 0.

\subsection{Monolingual Results}

The comparison of absolute performance differences across the three combinations: monolingual, cross-lingual with the instruction in the post language and cross-lingual with the instruction in the claim language is shown in Table~\ref{tab:overall-results-absolute}. The full results comparing the use of target language and English instructions across all models and prompting strategies in monolingual settings are presented in Table~\ref{tab:monolingual-all}.

\paragraph{Overall Trends.}

In the monolingual setting, we observed several patterns that emerged regarding the effect of using target language instructions versus English. While the overall differences were modest, a slightly greater number of model-prompting combinations performed better with target language instructions, suggesting \textbf{a subtle positive bias toward using the target language}. However, zero-shot prompting consistently led to poorer results when instructions were not in English (except \texttt{Gemma3 27B}), highlighting the dominance of English in the zero-shot setting without additional context.

\paragraph{Language Families.}

\textit{Germanic} languages, such as Dutch and German, tended to benefit from the target language instructions, with average improvements of around 4\% in macro F1. Similarly, \textit{Slavic} languages, such as Slovak, Czech, Polish, Serbo-Croatian, and Bulgarian, showed gains of approximately 2.5-3.5\% on average. In contrast, \textit{Sino-Tibetan} and other language families showed uniformly negative results with target language instructions, with some drops reaching up to 14\% for Burmese, pointing to a pronounced bias toward English for really low-resource languages.

\paragraph{Writing Script.}

Script also played a significant role. Languages written in Latin script were more likely to benefit from target language instructions, whereas those using non-Latin scripts tended to perform worse, particularly under zero-shot and CoT prompting. There were exceptions, such as Arabic, Bulgarian, and Serbo-Croatian, despite using non-Latin scripts, that still demonstrated performance improvements across multiple settings and models. In addition, larger models generally demonstrated greater gains from target language instructions, indicating that scale may help mitigate English-centric bias. These trends likely reflect the pertaining data distribution, which is heavily skewed toward Latin-script and English-language content.

\subsection{Cross-lingual Results}

\paragraph{Post-Language Instructions.}

Detailed results for the cross-lingual setting, comparing post-language and English-language instructions, are provided in Table~\ref{tab:post-langauge}.

We identified several findings when the instructions were given in the language of the post. Zero-shot prompting generally led to weaker performance compared to English-language instruction, reinforcing the strong default bias toward English in multilingual LLMs. However, the \texttt{Qwen3} models, for instance, demonstrated improvements with post-language instructions, but only when few-shot demonstrations were provided, suggesting that these models benefit more from additional context when processing non-English instructions. On the other hand, \texttt{Gemma3 12B} achieved higher performance with the post-language instruction in most prompting strategies (except zero-shot), reducing the bias between post-language and English-language instructions.

Additionally, language similarity also played a role. Pairs of linguistically related languages, such as Slovak and Czech and Polish and Serbo-Croatian, often achieved better performance with post-language instructions. This indicates that shared linguistic structures and vocabulary may help models better align the instruction with the input content in such cases.

\paragraph{Claim-Language Instructions.}

Results for the cross-lingual setting using claim-language instructions, where the instruction is given in the language of the claim (i.e., the second language in each pair), are presented in Table~\ref{tab:claim-language}.

\begin{table}[]
\centering
\resizebox{\columnwidth}{!}{%
\begin{tabular}{l|rrrr|r}
\toprule
\textbf{Model} & \textbf{ZS} & \multicolumn{1}{c}{\textbf{\begin{tabular}[c]{@{}c@{}}ZS\\ + Task\end{tabular}}} & \multicolumn{1}{c}{\textbf{\begin{tabular}[c]{@{}c@{}}FS\\ + Task\end{tabular}}} & \textbf{CoT} & \textbf{Avg.}\\
\midrule
\texttt{Qwen3 8B} & \cellcolor{lightcoral}{-18.40} & \cellcolor{lightcoral}{-9.41} & \cellcolor{lightgreen}{2.44} & \cellcolor{lightcoral}{-3.63} & \cellcolor{lightcoral}{-7.25}\\
\texttt{Qwen3 14B} & \cellcolor{lightcoral}{-20.46} & \cellcolor{lightcoral}{-3.44} & \cellcolor{lightgreen}{2.93} & \cellcolor{lightcoral}{-4.37} & \cellcolor{lightcoral}{-6.34} \\
\texttt{Gemma3 12B} & \cellcolor{lightcoral}{-11.07} & \cellcolor{lightcoral}{-5.44} & \cellcolor{lightgreen}{2.99} & \cellcolor{lightcoral}{-2.25} & \cellcolor{lightcoral}{-3.94}\\
\texttt{Gemma3 27B} & \cellcolor{lightcoral}{-2.11} & \cellcolor{lightgreen}{5.18} & \cellcolor{lightgreen}{3.91} & \cellcolor{lightgreen}{4.00} & \cellcolor{lightgreen}{2.75} \\
\texttt{Llama3.1 8B} & \cellcolor{lightcoral}{-12.02} & \cellcolor{lightcoral}{-25.06} & \cellcolor{lightcoral}{-2.31} & \cellcolor{lightcoral}{-8.60} & \cellcolor{lightcoral}{-12.00} \\
\texttt{Llama3.1 70B} & \cellcolor{lightcoral}{-11.22} & \cellcolor{lightgreen}{0.26} & \cellcolor{lightcoral}{-3.86} & \cellcolor{lightcoral}{-8.02} & \cellcolor{lightcoral}{-1.7} \\
\bottomrule
\end{tabular}}
\setlength{\fboxsep}{1pt}
\caption{Macro F1 score differences when using \textit{Chinese} vs. \textit{English} language for instructions across prompting strategies, evaluated on all annotated pairs (16K). Negative values (\colorbox{lightcoral}{red}) indicate lower performance with Chinese prompts; positive values (\colorbox{lightgreen}{green}) indicate improvements.}
\label{tab:zho-vs-eng}
\end{table}

\paragraph{Chinese vs English.}

To investigate language bias between \textit{Chinese} and \textit{English}, we compared model performance when instructions were provided in \textit{Chinese} versus \textit{English}. As shown in Table~\ref{tab:zho-vs-eng}, most models exhibited lower performance when prompted in Chinese, particularly under zero-shot (ZS) and zero-shot with task description (ZS +Task) settings. This suggests that English prompts remain more effective for guiding model behavior, mostly due to the model's pre-training data and instruction tuning being predominantly in English. Only the \texttt{Gemma3 27B} demonstrated improvements with Chinese prompts across 3 prompting strategies, resulting in an average improvement of 2.75. This highlights that expanding and diversifying the pertaining data to include more languages may lead to a trade-off: while it can improve performance in underrepresented languages, it may slightly degrade performance in English due to a more distributed linguistic focus.

\subsection{Thinking}

To further investigate the role of reasoning processes in instruction understanding, we conducted an ablation study on the impact of the thinking mode across different prompting strategies. In the monolingual setting, the results are shown in Table~\ref{tab:thinking-monolingual}, where we compare performance with and without "thinking mode" across 20 languages. To extend this analysis to multilingual scenarios, we also evaluated the effect of the thinking mode in cross-lingual settings. Table~\ref{tab:thinking-post-language} presents the results when using post-language instructions (the first language in each pair), while Table~\ref{tab:thinking-claim-language} shows the same analysis using claim-language instructions (the second language in each pair).

\begin{table*}[]
\resizebox{\textwidth}{!}{%
\small
\begin{tabular}{l|rrrr|r||rrrr|r||rrrr|r}
\toprule
 & \multicolumn{5}{c||}{\textbf{Monolingual}} & \multicolumn{5}{c||}{\textbf{Cross-lingual - post-language}} & \multicolumn{5}{c}{\textbf{Cross-lingual - claim-language}} \\ \midrule
\textbf{Model} & \multicolumn{1}{c}{\textbf{ZS}} & \multicolumn{1}{c}{\textbf{\begin{tabular}[c]{@{}c@{}}ZS\\ + Task\end{tabular}}} & \multicolumn{1}{c}{\textbf{\begin{tabular}[c]{@{}c@{}}FS\\ + Task\end{tabular}}} & \multicolumn{1}{c|}{\textbf{CoT}} & \multicolumn{1}{c||}{\textbf{Avg.}} & \multicolumn{1}{c}{\textbf{ZS}} & \multicolumn{1}{c}{\textbf{\begin{tabular}[c]{@{}c@{}}ZS\\ + Task\end{tabular}}} & \multicolumn{1}{c}{\textbf{\begin{tabular}[c]{@{}c@{}}FS\\ + Task\end{tabular}}} & \multicolumn{1}{c|}{\textbf{CoT}} & \multicolumn{1}{c||}{\textbf{Avg.}} & \multicolumn{1}{c}{\textbf{ZS}} & \multicolumn{1}{c}{\textbf{\begin{tabular}[c]{@{}c@{}}ZS\\ + Task\end{tabular}}} & \multicolumn{1}{c}{\textbf{\begin{tabular}[c]{@{}c@{}}FS\\ + Task\end{tabular}}} & \multicolumn{1}{c|}{\textbf{CoT}} & \multicolumn{1}{c}{\textbf{Avg.}} \\ \midrule
\multirow{2}{*}{\texttt{Qwen3 8B}} & 69.70 & 77.81 & 70.70 & 83.07 & 75.32 & 60.25 & 70.40 & 63.84 & 76.65 & 67.79 & 60.25 & 70.40 & 63.84 & 76.65 & 67.79 \\
& 7.06 & 3.82 & 6.15 & 6.31 & 5.84 & 6.17 & 3.97 & 6.84 & 8.85 & 6.46 & 3.06 & 3.99 & 3.22 & 5.08 & 3.84 \\
\midrule
\multirow{2}{*}{\texttt{Qwen3 14B}} & 73.40 & 77.08 & 74.72 & \textbf{85.15} & 77.59 & 64.12 & 70.99 & 69.29 & \textbf{78.24} & 70.66 & 64.12 & 70.99 & 69.29 & \textbf{78.24} & 70.66 \\
& 9.60 & 4.94 & 3.49 & 5.01 & 5.78 & 8.33 & 3.46 & 1.56 & 6.72 & 5.02 & 5.85 & 2.57 & 2.21 & 3.22 & 3.46 \\
\midrule
\multirow{2}{*}{\texttt{Gemma3 12B}} & 56.44 & 58.72 & 70.07 & 58.90 & 61.03 & 52.86 & 52.22 & 65.92 & 50.78 & 55.45 & 52.86 & 52.22 & 65.92 & 50.78 & 55.45 \\
& 9.47 & 7.21 & 3.67 & 6.32 & 6.67 & 8.30 & 4.46 & 3.50 & 5.04 & 5.33 & 4.39 & 2.95 & 1.67 & 2.79 & 2.95 \\
\midrule
\multirow{2}{*}{\texttt{Gemma3 27B}} & 59.35 & 55.54 & 72.72 & 52.41 & 60.00 & 53.69 & 49.54 & 64.94 & 46.37 & 53.64 & 53.69 & 49.54 & 64.94 & 46.37 & 53.64 \\
& 6.82 & 10.85 & 2.34 & 6.53 & 6.64 & 4.78 & 3.99 & 2.04 & 4.42 & 3.81 & 2.83 & 4.16 & 1.86 & 2.89 & 2.94 \\
\midrule
\multirow{2}{*}{\texttt{Llama3.1 8B}} & 50.46 & 71.13 & 51.12 & 67.18 & 59.97 & 45.53 & 66.48 & 50.56 & 58.82 & 55.35 & 45.53 & 66.48 & 50.56 & 58.82 & 55.35 \\
& 8.76 & 8.59 & 5.43 & 4.81 & 6.90 & 8.04 & 8.76 & 4.66 & 5.87 & 6.83 & 7.61 & 4.40 & 2.57 & 3.06 & 4.41  \\
\midrule
\multirow{2}{*}{\texttt{Llama3.1 70B}} & 73.97 & 74.38 & 54.04 & 75.48 & 69.47 & 65.57 & 65.77 & 54.34 & 67.66 & 63.33& 65.57 & 65.77 & 54.34 & 67.66 & 63.33 \\
& 6.82 & 7.69 & 3.48 & 7.10 & 6.27 & 7.09 & 9.22 & 4.52 & 4.73 & 6.39 & 3.59 & 4.12 & 2.70 & 3.35 & 3.44  \\ 
\midrule
\multirow{2}{*}{\textit{Average}} & 63.89 & 69.11 & 65.56 & 70.67 & 67.23 & 57.00 & 62.57 & 61.48 & 63.09 & 61.04 & 57.00 & 62.57 & 61.48 & 63.09 & 61.04 \\
& 8.09 &	7.18 &	4.09 &	6.01 &	6.34 &	7.12 &	5.64 &	3.85 &	5.94 &	5.64 &	4.56 &	3.70 &	2.37 &	3.40 &	3.51 \\
\midrule
\multirow{2}{*}{\makecell[l]{\texttt{Qwen3 8B}\\\texttt{(thinking)}}} & \textbf{77.19} & \textbf{82.74} & \textbf{81.27} & - & \textbf{80.40} & \textbf{66.29} & \textbf{74.43} & \textbf{73.22} & - & \textbf{71.31} & \textbf{66.29} & \textbf{74.43} & \textbf{73.22} & - & \textbf{71.31} \\
& 3.71 & 2.23 & 2.56 & - & 2.83 & 3.50 & 3.32 & 3.64 & - & 3.49 & 1.73 & 2.68 & 2.70 & - & 2.37  \\
\bottomrule
\end{tabular}}
\caption{Macro F1 performance across 20 languages and language combinations for multiple models in three settings: monolingual, cross-lingual with post-language instructions, and cross-lingual with claim-language instructions. The first row per model shows absolute performance with English instructions; the second row shows the absolute difference when using the target language for instruction. ZS = zero-shot; FS = few-shot; Task = using Task description. The highest absolute score for each column is in \textbf{bold}.}
\label{tab:overall-results-absolute}
\end{table*}

\begin{table*}[]
\centering
\resizebox{\textwidth}{!}{%
\begin{tabular}{llllllllllllllllllllll|l}
\toprule
\textbf{Model} & \textbf{Technique} & \multicolumn{1}{c}{\textbf{ara}} & \multicolumn{1}{c}{\textbf{bul}} & \multicolumn{1}{c}{\textbf{ces}} & \multicolumn{1}{c}{\textbf{deu}} & \multicolumn{1}{c}{\textbf{ell}} & \multicolumn{1}{c}{\textbf{eng}} & \multicolumn{1}{c}{\textbf{fra}} & \multicolumn{1}{c}{\textbf{hbs}} & \multicolumn{1}{c}{\textbf{hin}} & \multicolumn{1}{c}{\textbf{hun}} & \multicolumn{1}{c}{\textbf{kor}} & \multicolumn{1}{c}{\textbf{msa}} & \multicolumn{1}{c}{\textbf{mya}} & \multicolumn{1}{c}{\textbf{nld}} & \multicolumn{1}{c}{\textbf{pol}} & \multicolumn{1}{c}{\textbf{por}} & \multicolumn{1}{c}{\textbf{ron}} & \multicolumn{1}{c}{\textbf{slk}} & \multicolumn{1}{c}{\textbf{spa}} & \multicolumn{1}{c}{\textbf{tha}} & \multicolumn{1}{c}{\textbf{Avg.}} \\ 
\midrule
\multirow{4}{*}{\texttt{Qwen3 8B}}  & ZS & 77.18 & 72.26 & 63.19 & 64.14 & 77.41 & 75.52 & 73.27 & 59.14 & 74.88 & 69.9 & 74.51 & 69.12 & 65.65 & 67.8 & 57.96 & 70.75 & 69.73 & 64.97 & 71.05 & 75.48 & 69.7 \\
& ZS + Task & 88.31 & 88.95 & 68.97 & 78.9 & 82.31 & 78.26 & 80.39 & 63.24 & 79.79 & 78.22 & 76.54 & 76.02 & 82.47 & 75.56 & 67.48 & 78.13 & 75.13 & 76.22 & 80.63 & 80.61 & 77.81 \\
 & FS + Task  & 81.22 & 79.9 & 65.33 & 66.93 & 74.05 & 67.22 & 79.05 & 64.24 & 77.18 & 67.09 & 73.75 & 64.57 & 72.1 & 68.27 & 61.49 & 71.12 & 68.67 & 64.83 & 76.11 & 70.86 & 70.7 \\
& CoT & 89.44 & 91.26 & 81.94 & 80.42 & 82.51 & 82.45 & 88.51 & 73.6 & 87.3 & 80.04 & 86.07 & 86.06 & 78.01 & 81.03 & 74.16 & 80.63 & 84.16 & 80.48 & 84.81 & 88.57 & 83.07 \\
\midrule
\multicolumn{2}{l}{\textit{Average}} & 84.04 & 83.09 & 69.86 & 72.60 & 79.07 & 75.86 & 80.31 & 65.06 & 79.79 & 73.81 & 77.72 & 73.94 & 74.56 & 73.17 & 65.27 & 75.16 & 74.42 & 71.63 & 78.15 & 78.88 & 75.32 \\
\midrule
\multirow{4}{*}{\texttt{Qwen3 14B}}  & ZS & 77.43 & 80.9 & 61.97 & 73.9 & 80.22 & 75.34 & 77.01 & 60.32 & 77.31 & 68.47 & 79.34 & 77.35 & 73.32 & 72.49 & 62.45 & 69.07 & 73.32 & 70.79 & 73.96 & 83.01 & 73.4 \\
 & ZS + Task  & 86.1 & 83.79 & 66.34 & 79.65 & 84.62 & 79.66 & 82.35 & 65.99 & 80.12 & 70.32 & 74.79 & 80.5 & 79.88 & 75.57 & 67.69 & 77.32 & 73.73 & 70.7 & 79.7 & 82.68 & 77.08 \\
& FS + Task & 84.12 & 82.35 & 63.83 & 75.89 & 81.21 & 78.62 & 82.23 & 64.96 & 79.84 & 70.82 & 70.61 & 75.95 & 76.61 & 76.25 & 66.54 & 75.3 & 72.08 & 63.64 & 78.62 & 74.88 & 74.72 \\
 & CoT & 89.69 & 90.69 & 79.49 & 85.24 & 85.32 & 84.52 & 88.83 & 76.71 & 88.89 & 81.51 & 85.01 & 90.01 & 86.15 & 83.57 & 78.07 & 86.36 & 83.57 & 84.31 & 86.89 & 88.15 & 85.15 \\
 \midrule
 \multicolumn{2}{l}{\textit{Average}} & 84.34 & 84.43 & 67.91 & 78.67 & 82.84 & 79.54 & 82.61 & 67.00 & 81.54 & 72.78 & 77.44 & 80.95 & 78.99 & 76.97 & 68.69 & 77.01 & 75.68 & 72.36 & 79.79 & 82.18 & 77.59 \\
\midrule
\multirow{4}{*}{\texttt{Gemma3 12B}} & ZS  & 68.92 & 56.06 & 39.47 & 48.88 & 68.25 & 63.74 & 67.68 & 39.99 & 64.89 & 51.37 & 61.35 & 69.5 & 57.61 & 49.61 & 42.93 & 55.05 & 46.45 & 48.93 & 57.67 & 70.37 & 56.44 \\
 & ZS + Task & 70.04 & 62.21 & 49.27 & 52.8 & 67.33 & 60.96 & 65.17 & 44.73 & 67.2 & 53.49 & 59.92 & 65.43 & 65.53 & 51.46 & 48.63 & 59.48 & 52.52 & 51.06 & 59.71 & 67.55 & 58.72 \\
 & FS + Task & 76.97 & 80.46 & 67.32 & 68.27 & 77.74 & 69.57 & 75.04 & 65.11 & 71.31 & 62.52 & 69.74 & 63.63 & 69.37 & 71.63 & 61.67 & 75.34 & 66.23 & 63.72 & 72.21 & 73.6 & 70.07 \\
 & CoT & 71.85 & 61.02 & 49.78 & 54.34 & 69.17 & 65.45 & 62.25 & 45.42 & 69.43 & 55.11 & 60.94 & 64.93 & 63.62 & 52.57 & 48.58 & 59.48 & 51.57 & 49.13 & 60.9 & 62.36 & 58.9 \\
 \midrule
 \multicolumn{2}{l}{\textit{Average}} & 71.95 & 64.94 & 51.46 & 56.07 & 70.62 & 64.93 & 67.54 & 48.81 & 68.21 & 55.62 & 62.99 & 65.87 & 64.03 & 56.32 & 50.45 & 62.34 & 54.19 & 53.21 & 62.62 & 68.47 & 61.03 \\
\midrule
\multirow{4}{*}{\texttt{Gemma3 27B}} & ZS & 68.36 & 60.17 & 48.51 & 54.86 & 71.4 & 65.78 & 65.25 & 40.22 & 69.65 & 55.11 & 64.05 & 69.75 & 62.22 & 50.12 & 44.63 & 59.74 & 48.54 & 53.49 & 60.61 & 74.46 & 59.35 \\
 & ZS + Task & 66.05 & 59.81 & 48.3 & 48.32 & 66.19 & 57.92 & 62.79 & 38.0 & 64.85 & 50.92 & 56.93 & 60.98 & 64.15 & 48.08 & 44.17 & 53.12 & 47.59 & 51.2 & 59.08 & 62.36 & 55.54 \\
 & FS + Task & 84.04 & 81.41 & 65.28 & 64.65 & 83.13 & 74.48 & 76.19 & 58.5 & 75.96 & 64.58 & 74.54 & 71.26 & 83.35 & 67.64 & 63.95 & 76.85 & 67.97 & 64.05 & 74.98 & 81.59 & 72.72 \\
& CoT & 67.69 & 51.94 & 45.07 & 42.74 & 63.29 & 57.63 & 57.63 & 41.46 & 63.3 & 47.53 & 54.16 & 58.27 & 59.17 & 42.93 & 42.46 & 59.19 & 39.83 & 43.25 & 55.64 & 55.01 & 52.41 \\
\midrule
\multicolumn{2}{l}{\textit{Average}} & 71.54 & 63.33 & 51.79 & 52.64 & 71.00 & 63.95 & 65.47 & 44.55 & 68.44 & 54.54 & 62.42 & 65.07 & 67.22 & 52.19 & 48.80 & 62.23 & 50.98 & 53.00 & 62.58 & 68.36 & 60.01 \\
\midrule
\multirow{4}{*}{\texttt{Llama3.1 8B}} & ZS & 57.11 & 48.74 & 39.2 & 41.81 & 57.49 & 64.0 & 59.79 & 37.99 & 66.48 & 43.87 & 53.4 & 55.46 & 43.76 & 42.17 & 42.36 & 54.57 & 48.55 & 40.03 & 51.33 & 61.11 & 50.46 \\
 & ZS + Task & 79.84 & 72.29 & 62.5 & 68.36 & 78.36 & 77.26 & 80.4 & 61.12 & 79.96 & 64.18 & 68.44 & 75.32 & 74.43 & 70.63 & 56.81 & 70.93 & 73.2 & 63.63 & 73.34 & 71.55 & 71.13 \\
 & FS + Task & 56.97 & 52.69 & 49.3 & 49.74 & 48.31 & 47.3 & 50.71 & 48.19 & 54.87 & 51.47 & 53.66 & 41.85 & 52.0 & 45.13 & 52.28 & 51.3 & 54.93 & 58.04 & 50.0 & 53.66 & 51.12 \\
 & CoT & 73.39 & 74.6 & 60.73 & 65.96 & 69.8 & 76.23 & 70.71 & 56.94 & 73.52 & 64.66 & 67.09 & 74.33 & 62.22 & 64.88 & 55.72 & 66.87 & 65.81 & 64.97 & 64.3 & 70.89 & 67.18 \\
 \midrule
 \multicolumn{2}{l}{\textit{Average}} & 66.83 & 62.08 & 52.93 & 56.47 & 63.49 & 66.20 & 65.40 & 51.06 & 68.71 & 56.05 & 60.65 & 61.74 & 58.10 & 55.70 & 51.79 & 60.92 & 60.62 & 56.67 & 59.74 & 64.30 & 59.97 \\
\midrule
\multirow{4}{*}{\texttt{Llama3.1 70B}} & ZS & 79.21 & 81.31 & 69.19 & 74.95 & 81.04 & 78.21 & 79.34 & 64.38 & 85.59 & 69.41 & 73.46 & 81.89 & 56.59 & 70.78 & 61.25 & 73.26 & 72.4 & 73.26 & 77.11 & 76.81 & 73.97 \\
& ZS + Task & 80.45 & 88.23 & 66.76 & 71.4 & 80.72 & 77.37 & 77.31 & 62.76 & 85.52 & 67.11 & 74.76 & 78.28 & 70.41 & 68.76 & 62.64 & 74.58 & 73.45 & 68.23 & 77.84 & 81.11 & 74.38 \\
 & FS + Task & 56.98 & 47.65 & 54.01 & 53.99 & 47.65 & 69.42 & 64.7 & 52.97 & 51.92 & 46.87 & 58.97 & 41.41 & 47.02 & 52.48 & 54.11 & 62.64 & 57.54 & 51.33 & 60.44 & 48.8 & 54.04 \\
 & CoT & 82.62 & 46.51 & 77.75 & 78.92 & 77.68 & 81.34 & 80.54 & 68.15 & 76.67 & 72.64 & 83.19 & 81.21 & 56.48 & 78.56 & 70.6 & 77.83 & 81.97 & 76.79 & 75.43 & 84.78 & 75.48 \\
 \midrule
 \multicolumn{2}{l}{\textit{Average}} & 74.82 & 65.93 & 66.93 & 69.82 & 71.77 & 76.59 & 75.47 & 62.07 & 74.93 & 64.01 & 72.60 & 70.70 & 57.63 & 67.65 & 62.15 & 72.08 & 71.34 & 67.40 & 72.71 & 72.88 & 69.47 \\
\bottomrule
\end{tabular}}
\setlength{\fboxsep}{1pt}
\caption{Macro F1 performance using English instruction in monolingual settings across six LLMs and four prompting techniques. The average is calculated across all languages for each model and technique.}
\label{tab:monolingual-macrof1}
\end{table*}

\begin{table*}[]
\centering
\resizebox{\textwidth}{!}{%
\begin{tabular}{llllllllllllllllllllll|l}
\toprule
\textbf{Model} & \textbf{Technique} & \multicolumn{1}{c}{\textbf{ara}} & \multicolumn{1}{c}{\textbf{bul}} & \multicolumn{1}{c}{\textbf{ces}} & \multicolumn{1}{c}{\textbf{deu}} & \multicolumn{1}{c}{\textbf{ell}} & \multicolumn{1}{c}{\textbf{eng}} & \multicolumn{1}{c}{\textbf{fra}} & \multicolumn{1}{c}{\textbf{hbs}} & \multicolumn{1}{c}{\textbf{hin}} & \multicolumn{1}{c}{\textbf{hun}} & \multicolumn{1}{c}{\textbf{kor}} & \multicolumn{1}{c}{\textbf{msa}} & \multicolumn{1}{c}{\textbf{mya}} & \multicolumn{1}{c}{\textbf{nld}} & \multicolumn{1}{c}{\textbf{pol}} & \multicolumn{1}{c}{\textbf{por}} & \multicolumn{1}{c}{\textbf{ron}} & \multicolumn{1}{c}{\textbf{slk}} & \multicolumn{1}{c}{\textbf{spa}} & \multicolumn{1}{c}{\textbf{tha}} & \multicolumn{1}{c}{\textbf{Avg.}} \\ 
\midrule
\multirow{4}{*}{\texttt{Qwen3 8B}}  & ZS & \cellcolor{lightcoral}{-2.79} & \cellcolor{lightcoral}{-10.89} & \cellcolor{lightgreen}{0.43} & \cellcolor{lightgreen}{8.16} & \cellcolor{lightcoral}{-4.90} & 0.00 & \cellcolor{lightcoral}{-2.93} & \cellcolor{lightcoral}{-0.98} & \cellcolor{lightcoral}{-0.40} & \cellcolor{lightcoral}{-2.30} & \cellcolor{lightcoral}{-17.20} & \cellcolor{lightcoral}{-9.05} & \cellcolor{lightcoral}{-21.36} & \cellcolor{lightcoral}{-6.55} & \cellcolor{lightcoral}{-5.36} & \cellcolor{lightcoral}{-6.29} & \cellcolor{lightcoral}{-6.70} & \cellcolor{lightgreen}{1.89} & \cellcolor{lightcoral}{-8.33} & \cellcolor{lightcoral}{-24.67} & \cellcolor{lightcoral}{-6.01} \\
& ZS + Task & \cellcolor{lightcoral}{-1.13} & \cellcolor{lightcoral}{-0.18} & \cellcolor{lightgreen}{2.45} & \cellcolor{lightgreen}{2.79} & \cellcolor{lightcoral}{-1.68} & 0.00 & \cellcolor{lightgreen}{3.79} & \cellcolor{lightgreen}{2.50} & \cellcolor{lightcoral}{-3.14} & \cellcolor{lightcoral}{-3.19} & \cellcolor{lightcoral}{-6.89} & \cellcolor{lightcoral}{-3.44} & \cellcolor{lightcoral}{-19.68} & \cellcolor{lightgreen}{4.85} & \cellcolor{lightgreen}{3.68} & \cellcolor{lightcoral}{-2.71} & \cellcolor{lightcoral}{-1.11} & \cellcolor{lightcoral}{-3.58} & \cellcolor{lightgreen}{3.13} & \cellcolor{lightcoral}{-6.42} & \cellcolor{lightcoral}{-1.50} \\
 & FS + Task & \cellcolor{lightgreen}{4.36} & \cellcolor{lightgreen}{5.84} & \cellcolor{lightcoral}{-3.90} & \cellcolor{lightgreen}{6.86} & \cellcolor{lightcoral}{-10.00} & 0.00 & \cellcolor{lightcoral}{-4.76} & \cellcolor{lightgreen}{2.42} & \cellcolor{lightgreen}{7.17} & \cellcolor{lightgreen}{3.11} & \cellcolor{lightcoral}{-9.59} & \cellcolor{lightgreen}{1.04} & \cellcolor{lightcoral}{-22.66} & \cellcolor{lightgreen}{9.88} & \cellcolor{lightcoral}{-2.11} & \cellcolor{lightgreen}{4.62} & \cellcolor{lightgreen}{10.37} & \cellcolor{lightcoral}{-0.83} & \cellcolor{lightgreen}{2.83} & \cellcolor{lightgreen}{10.65} & \cellcolor{lightgreen}{0.76} \\
& CoT & \cellcolor{lightcoral}{-7.42} & \cellcolor{lightcoral}{-3.54} & \cellcolor{lightcoral}{-6.76} & \cellcolor{lightcoral}{-4.07} & \cellcolor{lightcoral}{-5.27} & 0.00 & \cellcolor{lightcoral}{-4.64} & \cellcolor{lightcoral}{-4.17} & \cellcolor{lightcoral}{-7.06} & \cellcolor{lightcoral}{-3.59} & \cellcolor{lightcoral}{-15.47} & \cellcolor{lightcoral}{-10.17} & \cellcolor{lightcoral}{-28.37} & \cellcolor{lightcoral}{-0.04} & \cellcolor{lightcoral}{-0.68} & \cellcolor{lightcoral}{-4.11} & \cellcolor{lightcoral}{-12.79} & \cellcolor{lightcoral}{-1.60} & \cellcolor{lightcoral}{-2.15} & \cellcolor{lightcoral}{-4.38} & \cellcolor{lightcoral}{-6.31} \\
\midrule
\multicolumn{2}{l}{\textit{Average}} & \cellcolor{lightcoral}{-1.74} & \cellcolor{lightcoral}{-2.20} & \cellcolor{lightcoral}{-1.94} & \cellcolor{lightgreen}{3.44} & \cellcolor{lightcoral}{-5.46} & 0.00 & \cellcolor{lightcoral}{-2.13} & \cellcolor{lightcoral}{-0.06} & \cellcolor{lightcoral}{-0.86} & \cellcolor{lightcoral}{-1.49} & \cellcolor{lightcoral}{-12.29} & \cellcolor{lightcoral}{-5.40} & \cellcolor{lightcoral}{-23.02} & \cellcolor{lightgreen}{2.04} & \cellcolor{lightcoral}{-1.12} & \cellcolor{lightcoral}{-2.12} & \cellcolor{lightcoral}{-2.56} & \cellcolor{lightcoral}{-1.03} & \cellcolor{lightcoral}{-1.13} & \cellcolor{lightcoral}{-6.21} & \cellcolor{lightcoral}{-3.26} \\
\midrule
\multirow{4}{*}{\texttt{Qwen3 14B}}  & ZS & \cellcolor{lightgreen}{2.17} & \cellcolor{lightcoral}{-0.33} & \cellcolor{lightgreen}{1.54} & \cellcolor{lightcoral}{-0.64} & \cellcolor{lightcoral}{-2.96} & 0.00 & \cellcolor{lightcoral}{-7.87} & \cellcolor{lightcoral}{-0.22} & \cellcolor{lightcoral}{-18.43} & \cellcolor{lightcoral}{-4.02} & \cellcolor{lightcoral}{-22.21} & \cellcolor{lightcoral}{-4.21} & \cellcolor{lightcoral}{-43.63} & \cellcolor{lightcoral}{-1.91} & \cellcolor{lightcoral}{-19.65} & \cellcolor{lightcoral}{-10.05} & \cellcolor{lightcoral}{-7.61} & \cellcolor{lightcoral}{-2.17} & \cellcolor{lightcoral}{-13.31} & \cellcolor{lightcoral}{-29.16} & \cellcolor{lightcoral}{-9.23} \\
 & ZS + Task & \cellcolor{lightgreen}{6.89} & \cellcolor{lightgreen}{7.62} & \cellcolor{lightgreen}{11.53} & \cellcolor{lightgreen}{2.45} & \cellcolor{lightgreen}{2.76} & 0.00 & \cellcolor{lightgreen}{2.50} & \cellcolor{lightgreen}{4.98} & \cellcolor{lightgreen}{6.37} & \cellcolor{lightgreen}{5.14} & \cellcolor{lightcoral}{-3.94} & \cellcolor{lightgreen}{1.20} & \cellcolor{lightcoral}{-10.46} & \cellcolor{lightgreen}{3.12} & \cellcolor{lightgreen}{6.94} & \cellcolor{lightgreen}{1.64} & \cellcolor{lightgreen}{6.64} & \cellcolor{lightgreen}{7.42} & \cellcolor{lightgreen}{4.09} & \cellcolor{lightcoral}{-3.15} & \cellcolor{lightgreen}{3.19} \\
& FS + Task & \cellcolor{lightgreen}{1.88} & \cellcolor{lightgreen}{7.68} & \cellcolor{lightgreen}{1.74} & \cellcolor{lightcoral}{-0.97} & \cellcolor{lightgreen}{1.46} & 0.00 & \cellcolor{lightcoral}{-2.27} & \cellcolor{lightgreen}{4.01} & \cellcolor{lightgreen}{1.54} & \cellcolor{lightcoral}{-2.80} & \cellcolor{lightgreen}{2.82} & \cellcolor{lightgreen}{1.03} & \cellcolor{lightcoral}{-19.01} & \cellcolor{lightcoral}{-0.95} & \cellcolor{lightgreen}{3.92} & \cellcolor{lightcoral}{-4.94} & \cellcolor{lightgreen}{0.30} & \cellcolor{lightgreen}{5.93} & \cellcolor{lightgreen}{3.26} & \cellcolor{lightgreen}{3.23} & \cellcolor{lightgreen}{0.39} \\
 & CoT & \cellcolor{lightgreen}{0.99} & \cellcolor{lightcoral}{-5.23} & \cellcolor{lightcoral}{-1.75} & \cellcolor{lightgreen}{0.94} & \cellcolor{lightcoral}{-0.58} & 0.00 & \cellcolor{lightcoral}{-3.40} & \cellcolor{lightcoral}{-3.13} & \cellcolor{lightcoral}{-2.92} & \cellcolor{lightcoral}{-0.67} & \cellcolor{lightcoral}{-9.55} & \cellcolor{lightcoral}{-25.56} & \cellcolor{lightcoral}{-26.19} & \cellcolor{lightcoral}{-0.32} & \cellcolor{lightcoral}{-1.25} & \cellcolor{lightcoral}{-6.45} & \cellcolor{lightcoral}{-4.18} & \cellcolor{lightcoral}{-1.59} & \cellcolor{lightcoral}{-0.99} & \cellcolor{lightcoral}{-4.52} & \cellcolor{lightcoral}{-4.82} \\
 \midrule
 \multicolumn{2}{l}{\textit{Average}} & \cellcolor{lightgreen}{2.98} & \cellcolor{lightgreen}{2.43} & \cellcolor{lightgreen}{3.27} & \cellcolor{lightgreen}{0.44} & \cellcolor{lightgreen}{0.17 }& 0.00 & \cellcolor{lightcoral}{-2.76} & \cellcolor{lightgreen}{1.41} & \cellcolor{lightcoral}{-3.36} & \cellcolor{lightcoral}{-0.59} & \cellcolor{lightcoral}{-8.22} & \cellcolor{lightcoral}{-6.88} & \cellcolor{lightcoral}{-24.82} & \cellcolor{lightcoral}{-0.01} & \cellcolor{lightcoral}{-2.51} & \cellcolor{lightcoral}{-4.95} & \cellcolor{lightcoral}{-1.21} & \cellcolor{lightgreen}{2.40} & \cellcolor{lightcoral}{-1.74} & \cellcolor{lightcoral}{-8.40} & \cellcolor{lightcoral}{-2.62} \\
\midrule
\multirow{4}{*}{\texttt{Gemma3 12B}} & ZS & \cellcolor{lightgreen}{1.44} & \cellcolor{lightgreen}{9.80} & \cellcolor{lightgreen}{17.40} & \cellcolor{lightcoral}{-0.40} & \cellcolor{lightcoral}{-6.25} & 0.00 & \cellcolor{lightgreen}{0.03} & \cellcolor{lightgreen}{6.65} & \cellcolor{lightcoral}{-18.48} & \cellcolor{lightcoral}{-0.22} & \cellcolor{lightcoral}{-23.97} & \cellcolor{lightcoral}{-15.68} & \cellcolor{lightcoral}{-32.20} & \cellcolor{lightcoral}{-1.81} & \cellcolor{lightgreen}{2.68} & \cellcolor{lightgreen}{1.95} & \cellcolor{lightgreen}{4.74} & \cellcolor{lightgreen}{10.59} & \cellcolor{lightgreen}{2.76} & \cellcolor{lightcoral}{-32.36} & \cellcolor{lightcoral}{-3.67} \\
 & ZS + Task & \cellcolor{lightgreen}{18.01} & \cellcolor{lightgreen}{7.85} & \cellcolor{lightgreen}{2.80} & \cellcolor{lightgreen}{3.15} & \cellcolor{lightgreen}{9.74} & 0.00 & \cellcolor{lightgreen}{4.39} & \cellcolor{lightgreen}{3.81} & \cellcolor{lightgreen}{8.47} & \cellcolor{lightgreen}{5.17} & \cellcolor{lightgreen}{4.77} & \cellcolor{lightgreen}{7.54} & \cellcolor{lightcoral}{-8.58} & \cellcolor{lightgreen}{17.67} & \cellcolor{lightgreen}{14.70} & \cellcolor{lightgreen}{6.43} & \cellcolor{lightgreen}{0.38} & \cellcolor{lightgreen}{12.93} & \cellcolor{lightgreen}{3.31} & \cellcolor{lightgreen}{4.58} & \cellcolor{lightgreen}{6.36} \\
 & FS + Task & \cellcolor{lightgreen}{9.61} & \cellcolor{lightcoral}{-0.80} & \cellcolor{lightcoral}{-6.85} & \cellcolor{lightcoral}{-4.71} & \cellcolor{lightgreen}{0.21} & 0.00 & \cellcolor{lightcoral}{-2.85} & \cellcolor{lightgreen}{0.59} & \cellcolor{lightgreen}{12.75} & \cellcolor{lightcoral}{-1.54} & \cellcolor{lightgreen}{2.38} & \cellcolor{lightgreen}{3.35} & \cellcolor{lightcoral}{-5.92} & \cellcolor{lightgreen}{0.97} & \cellcolor{lightgreen}{2.71} & \cellcolor{lightcoral}{-6.53} & \cellcolor{lightcoral}{-1.20} & \cellcolor{lightcoral}{-3.79} & \cellcolor{lightgreen}{2.90} & \cellcolor{lightgreen}{3.82} & \cellcolor{lightgreen}{0.26} \\
 & CoT & \cellcolor{lightgreen}{8.86} & \cellcolor{lightcoral}{-1.25} & \cellcolor{lightgreen}{3.94} & \cellcolor{lightgreen}{14.21} & \cellcolor{lightcoral}{-5.92} & 0.00 & \cellcolor{lightcoral}{-0.75} & \cellcolor{lightgreen}{8.72} & \cellcolor{lightgreen}{4.23} & \cellcolor{lightgreen}{7.68} & \cellcolor{lightcoral}{-8.68} & \cellcolor{lightgreen}{2.57} & \cellcolor{lightcoral}{-22.79} & \cellcolor{lightgreen}{15.44} & \cellcolor{lightgreen}{5.45} & \cellcolor{lightcoral}{-2.57} & \cellcolor{lightcoral}{-1.33} & \cellcolor{lightgreen}{5.19} & \cellcolor{lightgreen}{1.60} & \cellcolor{lightgreen}{5.20} & \cellcolor{lightgreen}{1.99} \\
 \midrule
 \multicolumn{2}{l}{\textit{Average}} & \cellcolor{lightgreen}{9.48} & \cellcolor{lightgreen}{3.90} & \cellcolor{lightgreen}{4.32} & \cellcolor{lightgreen}{3.06} & \cellcolor{lightcoral}{-0.56} & 0.00 & \cellcolor{lightgreen}{0.20} & \cellcolor{lightgreen}{4.94} & \cellcolor{lightgreen}{1.74} & \cellcolor{lightgreen}{2.78} & \cellcolor{lightcoral}{-6.37} & \cellcolor{lightcoral}{-0.56} & \cellcolor{lightcoral}{-17.37} & \cellcolor{lightgreen}{8.07} & \cellcolor{lightgreen}{6.38} & \cellcolor{lightcoral}{-0.18} & \cellcolor{lightgreen}{0.65} & \cellcolor{lightgreen}{6.23} & \cellcolor{lightgreen}{2.64} & \cellcolor{lightcoral}{-4.69} & \cellcolor{lightgreen}{1.23} \\
\midrule
\multirow{4}{*}{\texttt{Gemma3 27B}} & ZS & \cellcolor{lightgreen}{2.75} & \cellcolor{lightgreen}{10.45} & \cellcolor{lightgreen}{11.71} & \cellcolor{lightgreen}{4.38} & \cellcolor{lightgreen}{2.33} & 0.00 & \cellcolor{lightgreen}{1.58} & \cellcolor{lightgreen}{19.24} & \cellcolor{lightcoral}{-9.06} & \cellcolor{lightgreen}{0.42} & \cellcolor{lightcoral}{-5.79} & \cellcolor{lightgreen}{0.11} & \cellcolor{lightcoral}{-16.92} & \cellcolor{lightgreen}{3.55} & \cellcolor{lightcoral}{-0.09} & \cellcolor{lightcoral}{-5.43} & \cellcolor{lightgreen}{16.05} & \cellcolor{lightgreen}{5.82} & \cellcolor{lightgreen}{6.16} & \cellcolor{lightcoral}{-14.46} & \cellcolor{lightgreen}{1.64} \\
 & ZS + Task & \cellcolor{lightgreen}{12.74} & \cellcolor{lightgreen}{14.50} & \cellcolor{lightgreen}{6.90} & \cellcolor{lightgreen}{19.16} & \cellcolor{lightgreen}{7.35} & 0.00 & \cellcolor{lightgreen}{7.34} & \cellcolor{lightgreen}{11.11} & \cellcolor{lightgreen}{14.26} & \cellcolor{lightgreen}{6.77} & \cellcolor{lightgreen}{3.81} & \cellcolor{lightgreen}{16.11} & \cellcolor{lightgreen}{14.68} & \cellcolor{lightgreen}{13.62} & \cellcolor{lightgreen}{13.89} & \cellcolor{lightgreen}{10.86} & \cellcolor{lightgreen}{13.70} & \cellcolor{lightgreen}{3.99} & \cellcolor{lightgreen}{13.10} & \cellcolor{lightgreen}{13.19} & \cellcolor{lightgreen}{10.85} \\
 & FS + Task & \cellcolor{lightgreen}{2.43} & \cellcolor{lightcoral}{-3.08} & \cellcolor{lightgreen}{0.30} & \cellcolor{lightgreen}{6.53} & \cellcolor{lightgreen}{0.40} & 0.00 & \cellcolor{lightcoral}{-1.20} & \cellcolor{lightgreen}{0.01} & \cellcolor{lightgreen}{5.12} & \cellcolor{lightcoral}{-0.53} & \cellcolor{lightcoral}{-3.93} & \cellcolor{lightgreen}{3.55} & \cellcolor{lightcoral}{-2.75} & \cellcolor{lightgreen}{4.44} & \cellcolor{lightgreen}{4.80} & \cellcolor{lightgreen}{0.42} & \cellcolor{lightgreen}{1.41} & \cellcolor{lightcoral}{-1.52} & \cellcolor{lightcoral}{-2.09} & \cellcolor{lightgreen}{2.35} & \cellcolor{lightgreen}{0.83} \\
& CoT & \cellcolor{lightgreen}{10.61} & \cellcolor{lightgreen}{2.75} & \cellcolor{lightgreen}{3.98} & \cellcolor{lightgreen}{24.90} & \cellcolor{lightcoral}{-3.07} & 0.00 & \cellcolor{lightcoral}{-1.94} & \cellcolor{lightcoral}{-3.21} & \cellcolor{lightgreen}{10.23} & \cellcolor{lightgreen}{9.18} & \cellcolor{lightcoral}{-0.98} & \cellcolor{lightgreen}{10.48} & 0.00 & \cellcolor{lightgreen}{4.97} & \cellcolor{lightgreen}{5.57} & \cellcolor{lightgreen}{4.77} & \cellcolor{lightgreen}{7.25} & \cellcolor{lightgreen}{5.48} & \cellcolor{lightgreen}{5.52} & \cellcolor{lightgreen}{15.61} & \cellcolor{lightgreen}{5.60} \\
\midrule
\multicolumn{2}{l}{\textit{Average}} & \cellcolor{lightgreen}{7.13} & \cellcolor{lightgreen}{6.16} & \cellcolor{lightgreen}{5.72} & \cellcolor{lightgreen}{13.74} & \cellcolor{lightgreen}{1.75} & 0.00 & \cellcolor{lightgreen}{1.45} & \cellcolor{lightgreen}{6.79} & \cellcolor{lightgreen}{5.14} & \cellcolor{lightgreen}{3.96} & \cellcolor{lightcoral}{-1.72} & \cellcolor{lightgreen}{7.56} & \cellcolor{lightcoral}{-1.25} & \cellcolor{lightgreen}{6.64} & \cellcolor{lightgreen}{6.04} & \cellcolor{lightgreen}{2.65} & \cellcolor{lightgreen}{9.60} & \cellcolor{lightgreen}{3.44} & \cellcolor{lightgreen}{5.67} & \cellcolor{lightgreen}{4.17} & \cellcolor{lightgreen}{4.73} \\
\midrule
\multirow{4}{*}{\texttt{Llama3.1 8B}} & ZS & \cellcolor{lightgreen}{13.13} & \cellcolor{lightcoral}{-10.19} & \cellcolor{lightgreen}{5.67} & \cellcolor{lightgreen}{10.69} & \cellcolor{lightcoral}{-0.76} & 0.00 & \cellcolor{lightcoral}{-5.28} & \cellcolor{lightgreen}{15.08} & \cellcolor{lightcoral}{-17.93} & \cellcolor{lightcoral}{-6.59} & \cellcolor{lightgreen}{2.43} & \cellcolor{lightgreen}{3.03} & \cellcolor{lightcoral}{-8.50} & \cellcolor{lightcoral}{-4.45} & \cellcolor{lightgreen}{4.06} & \cellcolor{lightcoral}{-8.73} & \cellcolor{lightgreen}{7.36} & \cellcolor{lightgreen}{10.67} & \cellcolor{lightcoral}{-2.31} & \cellcolor{lightcoral}{-38.37} & \cellcolor{lightcoral}{-1.55} \\
 & ZS + Task & \cellcolor{lightgreen}{0.66} & \cellcolor{lightcoral}{-5.34} & \cellcolor{lightcoral}{-1.77} & \cellcolor{lightgreen}{0.34} & \cellcolor{lightcoral}{-15.35} & 0.00 & \cellcolor{lightcoral}{-7.80} & \cellcolor{lightcoral}{-2.08} & \cellcolor{lightcoral}{-17.14} & \cellcolor{lightgreen}{3.67} & \cellcolor{lightcoral}{-54.58} & \cellcolor{lightcoral}{-6.47} & \cellcolor{lightcoral}{-26.67} & \cellcolor{lightcoral}{-1.54} & \cellcolor{lightgreen}{6.61} & \cellcolor{lightcoral}{-4.94} & \cellcolor{lightcoral}{-5.73} & \cellcolor{lightcoral}{-1.70} & \cellcolor{lightgreen}{2.58} & \cellcolor{lightcoral}{-6.83} & \cellcolor{lightcoral}{-7.20} \\
 & FS + Task & \cellcolor{lightgreen}{5.07} & \cellcolor{lightgreen}{5.53} & \cellcolor{lightgreen}{1.69} & \cellcolor{lightgreen}{8.05} & \cellcolor{lightgreen}{10.59} & 0.00 & \cellcolor{lightgreen}{9.30} & \cellcolor{lightgreen}{3.67} & \cellcolor{lightcoral}{-3.20} & \cellcolor{lightgreen}{2.19} & \cellcolor{lightgreen}{5.20} & \cellcolor{lightgreen}{8.39} & \cellcolor{lightcoral}{-1.37} & \cellcolor{lightgreen}{15.76} & \cellcolor{lightgreen}{8.29} & \cellcolor{lightgreen}{0.06} & \cellcolor{lightgreen}{0.06} & \cellcolor{lightcoral}{-5.62} & \cellcolor{lightgreen}{11.69} & \cellcolor{lightgreen}{2.85} & \cellcolor{lightgreen}{4.41} \\
 & CoT & \cellcolor{lightcoral}{-6.69} & \cellcolor{lightcoral}{-2.83} & \cellcolor{lightgreen}{5.71} & \cellcolor{lightcoral}{-0.21} & \cellcolor{lightcoral}{-0.92} & 0.00 & \cellcolor{lightcoral}{-1.47} & \cellcolor{lightgreen}{4.26} & \cellcolor{lightcoral}{-20.68} & \cellcolor{lightcoral}{-5.69} & \cellcolor{lightcoral}{-2.94} & \cellcolor{lightcoral}{-8.12} & \cellcolor{lightcoral}{-11.36} & \cellcolor{lightcoral}{-0.17} & \cellcolor{lightgreen}{7.04} & \cellcolor{lightcoral}{-0.66} & \cellcolor{lightcoral}{-9.71} & \cellcolor{lightgreen}{1.36} & \cellcolor{lightgreen}{0.39} & \cellcolor{lightgreen}{5.92} & \cellcolor{lightcoral}{-2.34} \\
 \midrule
 \multicolumn{2}{l}{\textit{Average}} & \cellcolor{lightgreen}{3.04} & \cellcolor{lightcoral}{-3.21} & \cellcolor{lightgreen}{2.83} & \cellcolor{lightgreen}{4.72} & \cellcolor{lightcoral}{-1.61} & 0.00 & \cellcolor{lightcoral}{-1.31} & \cellcolor{lightgreen}{5.23} & \cellcolor{lightcoral}{-14.74} & \cellcolor{lightcoral}{-1.61} & \cellcolor{lightcoral}{-12.47} & \cellcolor{lightcoral}{-0.79} & \cellcolor{lightcoral}{-11.98} & \cellcolor{lightgreen}{2.40} & \cellcolor{lightgreen}{6.50} & \cellcolor{lightcoral}{-3.57} & \cellcolor{lightcoral}{-2.01} & \cellcolor{lightgreen}{1.18} & \cellcolor{lightgreen}{3.09} & \cellcolor{lightcoral}{-9.11} & \cellcolor{lightcoral}{-1.67} \\
\midrule
\multirow{4}{*}{\texttt{Llama3.1 70B}} & ZS & \cellcolor{lightcoral}{-10.83} & \cellcolor{lightcoral}{-6.35} & \cellcolor{lightgreen}{3.42} & \cellcolor{lightgreen}{1.22} & \cellcolor{lightcoral}{-7.71} & 0.00 & \cellcolor{lightcoral}{-2.81} & \cellcolor{lightcoral}{-11.48} & \cellcolor{lightcoral}{-20.65} & \cellcolor{lightcoral}{-2.19} & \cellcolor{lightcoral}{-6.28} & \cellcolor{lightcoral}{-6.48} & \cellcolor{lightcoral}{-10.16} & \cellcolor{lightgreen}{2.70} & \cellcolor{lightgreen}{9.05} & \cellcolor{lightcoral}{-4.32} & \cellcolor{lightcoral}{-3.65} & \cellcolor{lightcoral}{-4.80} & \cellcolor{lightcoral}{-4.92} & \cellcolor{lightcoral}{-17.44} & \cellcolor{lightcoral}{-5.18} \\
& ZS + Task & \cellcolor{lightcoral}{-23.07} & \cellcolor{lightgreen}{1.36} & \cellcolor{lightgreen}{8.24} & \cellcolor{lightgreen}{9.17} & \cellcolor{lightgreen}{0.79} & 0.00 & \cellcolor{lightgreen}{4.62} & \cellcolor{lightgreen}{11.38} & \cellcolor{lightgreen}{0.77} & \cellcolor{lightcoral}{-2.44} & \cellcolor{lightgreen}{0.70} & \cellcolor{lightgreen}{2.31} & \cellcolor{lightcoral}{-23.53} & \cellcolor{lightgreen}{11.37} & \cellcolor{lightgreen}{9.86} & \cellcolor{lightgreen}{1.80} & \cellcolor{lightgreen}{1.06} & \cellcolor{lightgreen}{13.40} & \cellcolor{lightgreen}{4.80} & \cellcolor{lightcoral}{-23.07} & \cellcolor{lightgreen}{0.48} \\
 & FS + Task & \cellcolor{lightcoral}{-6.86} & \cellcolor{lightgreen}{5.74} & \cellcolor{lightcoral}{-2.74} & \cellcolor{lightgreen}{0.43} & \cellcolor{lightcoral}{-8.18} & 0.00 & \cellcolor{lightcoral}{-8.22} & \cellcolor{lightcoral}{-0.26} & \cellcolor{lightgreen}{4.17} & \cellcolor{lightgreen}{3.28} & \cellcolor{lightgreen}{1.34} & \cellcolor{lightgreen}{6.28} & \cellcolor{lightgreen}{2.15} & \cellcolor{lightgreen}{6.28} & \cellcolor{lightgreen}{3.52} & \cellcolor{lightcoral}{-1.69} & \cellcolor{lightgreen}{2.75} & \cellcolor{lightgreen}{0.70} & \cellcolor{lightcoral}{-0.11} & \cellcolor{lightgreen}{4.82} & \cellcolor{lightgreen}{0.67} \\
 & CoT & \cellcolor{lightcoral}{-11.84} & \cellcolor{lightgreen}{33.18} & \cellcolor{lightcoral}{-3.79} & \cellcolor{lightcoral}{-1.29} & \cellcolor{lightgreen}{4.50} & 0.00 & \cellcolor{lightcoral}{-0.13} & \cellcolor{lightgreen}{0.69} & \cellcolor{lightgreen}{5.67} & \cellcolor{lightcoral}{-3.75} & \cellcolor{lightcoral}{-18.97} & \cellcolor{lightcoral}{-6.11} & \cellcolor{lightcoral}{-9.18} & \cellcolor{lightgreen}{2.80} & \cellcolor{lightgreen}{0.49} & \cellcolor{lightcoral}{-9.80} & \cellcolor{lightcoral}{-11.29} & \cellcolor{lightgreen}{3.86} & \cellcolor{lightcoral}{-5.08} & \cellcolor{lightcoral}{-9.50} & \cellcolor{lightcoral}{-1.98} \\
 \midrule
 \multicolumn{2}{l}{\textit{Average}} & \cellcolor{lightcoral}{-13.15} & \cellcolor{lightgreen}{8.48} & \cellcolor{lightgreen}{1.28} & \cellcolor{lightgreen}{2.38} & \cellcolor{lightcoral}{-2.65} & 0.00 & \cellcolor{lightcoral}{-1.64} & \cellcolor{lightgreen}{0.08} & \cellcolor{lightcoral}{-2.51} & \cellcolor{lightcoral}{-1.27} & \cellcolor{lightcoral}{-5.80} & \cellcolor{lightcoral}{-1.00} & \cellcolor{lightcoral}{-10.18} & \cellcolor{lightgreen}{5.79} & \cellcolor{lightgreen}{5.73} & \cellcolor{lightcoral}{-3.50} & \cellcolor{lightcoral}{-2.78} & \cellcolor{lightgreen}{3.29} & \cellcolor{lightcoral}{-1.33} & \cellcolor{lightcoral}{-11.30} & \cellcolor{lightcoral}{-1.50} \\
\bottomrule
\end{tabular}}
\setlength{\fboxsep}{1pt}
\caption{Performance difference in Macro F1 when using target language instruction versus English instruction in monolingual settings across six LLMs and four prompting techniques. Positive values (\colorbox{lightgreen}{green}) indicate improved performance with target language instructions; negative values (\colorbox{lightcoral}{red}) indicate better performance with English instructions. The average is calculated across all languages for each model and technique.}
\label{tab:monolingual-all}
\end{table*}

\begin{table*}[]
\centering
\resizebox{\textwidth}{!}{%
\begin{tabular}{llllllllllllllllllllll|l}
\toprule
\textbf{Model} & \textbf{Technique} & \textbf{spa-eng} & \textbf{hin-eng} & \textbf{eng-ara} & \textbf{fra-eng} & \textbf{deu-eng} & \textbf{eng-por} & \textbf{spa-por} & \textbf{deu-fra} & \textbf{slk-ces} & \textbf{slk-eng} & \textbf{pol-hbs} & \textbf{ces-eng} & \textbf{ces-pol} & \textbf{nld-deu} & \textbf{msa-ara} & \textbf{kor-eng} & \textbf{mya-msa} & \textbf{ara-fra} & \textbf{hun-pol} & \textbf{tha-por} & \textbf{Avg.} \\
\midrule
\multirow{4}{*}{\texttt{Qwen3 8B}}  & ZS & 68.93 & 70.63 & 63.07 & 61.95 & 62.53 & 62.79 & 56.81 & 63.81 & 55.33 & 65.66 & 57.51 & 63.18 & 50.41 & 59.14 & 50.12 & 62.21 & 45.24 & 65.4 & 66.88 & 53.31 & 60.25 \\
& ZS + Task & 80.93 & 84.45 & 78.01 & 71.11 & 72.93 & 69.69 & 64.19 & 73.3 & 57.43 & 75.06 & 60.87 & 69.56 & 64.84 & 68.57 & 65.03 & 71.24 & 60.67 & 74.09 & 74.31 & 71.73 & 70.4 \\
 & FS + Task & 68.88 & 71.49 & 80.8 & 62.24 & 59.65 & 57.25 & 58.79 & 67.79 & 57.58 & 67.1 & 56.85 & 60.57 & 66.09 & 61.95 & 58.3 & 68.14 & 49.81 & 77.29 & 67.25 & 58.95 & 63.84 \\
& CoT & 86.1 & 91.4 & 84.03 & 80.35 & 80.95 & 77.63 & 64.96 & 67.85 & 72.82 & 80.03 & 74.1 & 78.98 & 74.16 & 74.65 & 69.97 & 75.25 & 69.62 & 78.52 & 75.8 & 75.85 & 76.65 \\
\midrule
\multicolumn{2}{l}{\textit{Average}} & 76.21 & 79.49 & 76.48 & 68.91 & 69.02 & 66.84 & 61.19 & 68.19 & 60.79 & 71.96 & 62.33 & 68.07 & 63.88 & 66.08 & 60.86 & 69.21 & 56.34 & 73.83 & 71.06 & 64.96 & 67.79 \\
\midrule
\multirow{4}{*}{\texttt{Qwen3 14B}} & 80.05 & 74.69 & 69.86 & 70.83 & 68.21 & 64.95 & 57.89 & 73.88 & 52.89 & 62.72 & 54.05 & 68.44 & 56.26 & 64.8 & 51.47 & 64.7 & 49.14 & 70.7 & 69.28 & 57.59 & 64.12 \\
 & ZS + Task & 82.08 & 80.74 & 78.38 & 71.37 & 66.41 & 70.86 & 71.96 & 75.58 & 53.94 & 71.49 & 60.26 & 68.76 & 60.68 & 73.74 & 71.05 & 79.98 & 64.81 & 76.74 & 73.78 & 67.22 & 70.99 \\
& FS + Task & 75.17 & 82.93 & 83.29 & 66.64 & 69.54 & 66.2 & 68.21 & 73.89 & 55.0 & 67.79 & 60.0 & 67.0 & 62.02 & 69.35 & 67.33 & 74.49 & 58.97 & 77.27 & 74.49 & 66.18 & 69.29 \\
 & CoT & 88.62 & 91.71 & 84.03 & 85.66 & 82.76 & 78.16 & 74.62 & 77.36 & 67.75 & 81.91 & 71.1 & 83.28 & 76.04 & 75.73 & 75.76 & 80.82 & 63.97 & 74.49 & 78.33 & 72.64 & 78.24 \\
 \midrule
 \multicolumn{2}{l}{\textit{Average}} & 81.48 & 82.52 & 78.89 & 73.63 & 71.73 & 70.04 & 68.17 & 75.18 & 57.40 & 70.98 & 61.35 & 71.87 & 63.75 & 70.91 & 66.40 & 75.00 & 59.22 & 74.80 & 73.97 & 65.91 & 70.66 \\
\midrule
\multirow{4}{*}{\texttt{Gemma3 12B}} & ZS & 67.13 & 42.17 & 64.37 & 55.36 & 50.9 & 54.04 & 48.16 & 49.24 & 36.0 & 56.87 & 38.67 & 52.0 & 49.64 & 45.74 & 49.59 & 62.93 & 46.06 & 63.73 & 61.31 & 63.3 & 52.86 \\
 & ZS + Task & 61.75 & 45.92 & 63.36 & 49.68 & 50.9 & 52.69 & 48.44 & 50.83 & 38.14 & 57.96 & 40.76 & 48.68 & 48.93 & 44.76 & 50.44 & 61.34 & 49.17 & 60.66 & 60.9 & 59.07 & 52.22 \\
 & FS + Task & 71.86 & 77.38 & 87.05 & 66.97 & 62.04 & 57.54 & 57.28 & 63.36 & 58.45 & 61.47 & 62.77 & 69.87 & 67.08 & 66.97 & 60.22 & 68.39 & 49.87 & 70.92 & 70.52 & 68.4 & 65.92 \\
 & CoT & 66.84 & 49.4 & 62.46 & 54.02 & 53.21 & 53.55 & 47.42 & 47.74 & 35.45 & 53.66 & 38.67 & 50.82 & 46.26 & 36.67 & 45.71 & 64.41 & 45.24 & 59.24 & 54.68 & 50.12 & 50.78 \\
 \midrule
 \multicolumn{2}{l}{\textit{Average}} & 66.90 & 53.72 & 69.31 & 56.51 & 54.26 & 54.46 & 50.33 & 52.79 & 42.01 & 57.49 & 45.22 & 55.34 & 52.98 & 48.54 & 51.49 & 64.27 & 47.59 & 63.64 & 61.85 & 60.22 & 55.45 \\
\midrule
\multirow{4}{*}{\texttt{Gemma3 27B}} & ZS  & 70.94 & 48.76 & 62.39 & 54.7 & 57.0 & 54.21 & 48.8 & 50.83 & 41.65 & 57.5 & 36.76 & 55.18 & 49.11 & 42.07 & 56.61 & 64.84 & 49.55 & 61.28 & 58.27 & 53.46 & 53.69 \\
 & ZS + Task & 64.53 & 45.61 & 59.38 & 49.31 & 47.19 & 46.4 & 49.64 & 42.65 & 36.72 & 54.18 & 34.42 & 46.49 & 44.3 & 40.66 & 52.64 & 57.72 & 51.56 & 58.97 & 59.07 & 49.43 & 49.54 \\
 & FS + Task & 75.74 & 64.68 & 85.36 & 60.06 & 60.67 & 60.55 & 61.74 & 66.66 & 53.21 & 65.75 & 57.7 & 65.46 & 60.26 & 63.28 & 64.9 & 72.48 & 54.53 & 64.1 & 71.37 & 70.33 & 64.94 \\
& CoT & 63.6 & 44.36 & 57.97 & 48.37 & 45.94 & 46.24 & 46.44 & 39.05 & 31.72 & 49.06 & 31.75 & 42.46 & 43.22 & 27.17 & 51.99 & 55.99 & 46.37 & 56.82 & 53.49 & 45.48 & 46.37 \\
\midrule
\multicolumn{2}{l}{\textit{Average}} & 68.70 & 50.85 & 66.28 & 53.11 & 52.70 & 51.85 & 51.66 & 49.80 & 40.83 & 56.62 & 40.16 & 52.40 & 49.22 & 43.30 & 56.54 & 62.76 & 50.50 & 60.29 & 60.55 & 54.68 & 53.64 \\
\midrule
\multirow{4}{*}{\texttt{Llama3.1 8B}} & ZS & 62.12 & 52.19 & 47.31 & 57.28 & 49.67 & 46.89 & 47.05 & 45.94 & 29.98 & 53.72 & 31.53 & 45.24 & 27.06 & 31.28 & 35.86 & 60.57 & 38.11 & 52.82 & 45.03 & 51.01 & 45.53 \\
 & ZS + Task & 77.38 & 76.04 & 66.75 & 72.6 & 67.34 & 65.43 & 56.44 & 69.46 & 50.93 & 71.69 & 57.07 & 67.58 & 57.75 & 61.65 & 57.84 & 78.03 & 69.62 & 67.33 & 71.05 & 67.66 & 66.48 \\
 & FS + Task & 50.65 & 51.2 & 48.69 & 51.02 & 51.32 & 52.79 & 48.37 & 53.07 & 53.1 & 49.97 & 47.53 & 48.24 & 54.4 & 53.31 & 47.71 & 47.37 & 48.65 & 52.97 & 51.62 & 49.3 & 50.56 \\
 & CoT & 71.89 & 56.59 & 57.3 & 64.54 & 65.99 & 58.45 & 48.83 & 62.55 & 49.83 & 71.05 & 52.02 & 63.36 & 48.91 & 53.2 & 49.3 & 70.1 & 51.9 & 54.73 & 60.77 & 64.96 & 58.82 \\
 \midrule
 \multicolumn{2}{l}{\textit{Average}} & 65.51 & 59.01 & 55.01 & 61.36 & 58.58 & 55.89 & 50.17 & 57.76 & 45.96 & 61.61 & 47.04 & 56.11 & 47.03 & 49.86 & 47.68 & 64.02 & 52.07 & 56.96 & 57.12 & 58.23 & 55.35 \\
\midrule
\multirow{4}{*}{\texttt{Llama3.1 70B}} & ZS & 77.95 & 72.07 & 69.07 & 67.35 & 72.78 & 65.42 & 52.79 & 64.58 & 57.61 & 78.0 & 58.58 & 66.58 & 56.56 & 61.38 & 55.0 & 74.51 & 52.52 & 66.33 & 71.61 & 70.66 & 65.57 \\
& ZS + Task & 75.06 & 73.61 & 71.07 & 67.41 & 65.71 & 65.74 & 56.99 & 61.08 & 55.74 & 73.45 & 59.93 & 64.23 & 58.58 & 63.25 & 58.3 & 75.75 & 49.56 & 70.91 & 73.63 & 75.44 & 65.77 \\
 & FS + Task & 65.84 & 65.64 & 48.25 & 64.97 & 62.04 & 51.34 & 56.38 & 49.23 & 56.76 & 55.98 & 46.67 & 50.98 & 51.67 & 49.81 & 49.04 & 56.89 & 49.43 & 48.78 & 53.21 & 53.92 & 54.34 \\
 & CoT & 77.33 & 74.25 & 71.21 & 73.46 & 69.07 & 65.53 & 56.56 & 67.4 & 65.46 & 73.04 & 60.77 & 72.89 & 67.34 & 66.48 & 53.3 & 77.55 & 49.37 & 62.32 & 74.03 & 75.89 & 67.66 \\
 \midrule
 \multicolumn{2}{l}{\textit{Average}} & 74.05 & 71.39 & 64.90 & 68.30 & 67.40 & 62.01 & 55.68 & 60.57 & 58.89 & 70.12 & 56.49 & 63.67 & 58.54 & 60.23 & 53.91 & 71.18 & 50.22 & 62.09 & 68.12 & 68.98 & 63.34 \\
\bottomrule
\end{tabular}}
\setlength{\fboxsep}{1pt}
\caption{Macro F1 performance with English instruction in cross-lingual settings across six LLMs and four prompting techniques. The average is calculated across all languages for each model and technique.}
\label{tab:crosslingual-macrof1}
\end{table*}

\begin{table*}
\resizebox{\textwidth}{!}{%
\begin{tabular}{lllllllllllllllllllllll}
\toprule
\textbf{Model} & \textbf{Technique} & \textbf{spa-eng} & \textbf{hin-eng} & \textbf{eng-ara} & \textbf{fra-eng} & \textbf{deu-eng} & \textbf{eng-por} & \textbf{spa-por} & \textbf{deu-fra} & \textbf{slk-ces} & \textbf{slk-eng} & \textbf{pol-hbs} & \textbf{ces-eng} & \textbf{ces-pol} & \textbf{nld-deu} & \textbf{msa-ara} & \textbf{kor-eng} & \textbf{mya-msa} & \textbf{ara-fra} & \textbf{hun-pol} & \textbf{tha-por} & \textbf{Avg.} \\
\midrule
\multirow{4}{*}{\texttt{Qwen3 8B}} & ZS & \cellcolor{lightcoral}{-3.16} & \cellcolor{lightcoral}{-6.06} & 0.00 & \cellcolor{lightcoral}{-2.71} & \cellcolor{lightgreen}{2.73} & 0.00 & \cellcolor{lightcoral}{-2.47} & \cellcolor{lightgreen}{5.14} & \cellcolor{lightgreen}{0.95} & \cellcolor{lightcoral}{-6.50} & \cellcolor{lightcoral}{-5.14} & \cellcolor{lightcoral}{-5.66} & \cellcolor{lightcoral}{-0.91} & \cellcolor{lightcoral}{-5.25} & \cellcolor{lightcoral}{-20.79} & \cellcolor{lightcoral}{-8.90} & \cellcolor{lightcoral}{-17.21} & \cellcolor{lightgreen}{3.23} & \cellcolor{lightcoral}{-8.52} & \cellcolor{lightcoral}{-18.15} & \cellcolor{lightcoral}{-4.97} \\
 & ZS + Task & \cellcolor{lightgreen}{2.02} & \cellcolor{lightcoral}{-0.66} & 0.00 & \cellcolor{lightgreen}{0.49} & \cellcolor{lightgreen}{2.26} & 0.00 & \cellcolor{lightgreen}{4.02} & \cellcolor{lightcoral}{-2.26} & \cellcolor{lightcoral}{-0.43} & \cellcolor{lightcoral}{-2.81} & \cellcolor{lightgreen}{4.90} & \cellcolor{lightcoral}{-0.08} & \cellcolor{lightgreen}{3.48} & \cellcolor{lightgreen}{3.34} & \cellcolor{lightcoral}{-16.01} & \cellcolor{lightgreen}{1.78} & \cellcolor{lightcoral}{-11.74} & \cellcolor{lightcoral}{-4.12} & \cellcolor{lightcoral}{-8.10} & \cellcolor{lightcoral}{-10.99} & \cellcolor{lightcoral}{-1.75} \\
 & FS + Task & \cellcolor{lightgreen}{7.74} & \cellcolor{lightgreen}{14.28} & 0.00 & \cellcolor{lightgreen}{6.78} & \cellcolor{lightgreen}{4.75} & 0.00 & \cellcolor{lightgreen}{10.24} & \cellcolor{lightgreen}{7.09} & \cellcolor{lightcoral}{-7.30} & \cellcolor{lightcoral}{-7.15} & \cellcolor{lightgreen}{5.79} & \cellcolor{lightcoral}{-5.31} & \cellcolor{lightcoral}{-11.24} & \cellcolor{lightgreen}{8.09} & \cellcolor{lightcoral}{-6.00} & \cellcolor{lightcoral}{-9.00} & \cellcolor{lightcoral}{-11.70} & \cellcolor{lightcoral}{-5.68} & \cellcolor{lightgreen}{0.99} & \cellcolor{lightgreen}{7.72} & \cellcolor{lightgreen}{0.51} \\
 & CoT & \cellcolor{lightcoral}{-9.03} & \cellcolor{lightcoral}{-17.98} & 0.00 & \cellcolor{lightcoral}{-5.45} & \cellcolor{lightcoral}{-5.44} & 0.00 & \cellcolor{lightcoral}{-7.91} & \cellcolor{lightgreen}{3.32} & \cellcolor{lightcoral}{-8.34} & \cellcolor{lightcoral}{-4.56} & \cellcolor{lightcoral}{-2.33} & \cellcolor{lightcoral}{-10.80} & \cellcolor{lightcoral}{-10.18} & \cellcolor{lightcoral}{-3.48} & \cellcolor{lightcoral}{-21.66} & \cellcolor{lightcoral}{-5.34} & \cellcolor{lightcoral}{-32.27} & \cellcolor{lightcoral}{-16.79} & \cellcolor{lightcoral}{-3.45} & \cellcolor{lightcoral}{-8.66} & \cellcolor{lightcoral}{-8.52} \\
\midrule
\multicolumn{2}{l}{\textit{Average}} & \cellcolor{lightcoral}{-0.61} & \cellcolor{lightcoral}{-2.61} & 0.00 & \cellcolor{lightcoral}{-0.22} & \cellcolor{lightgreen}{1.07} & 0.00 & \cellcolor{lightgreen}{0.97} & \cellcolor{lightgreen}{3.32} & \cellcolor{lightcoral}{-3.78} & \cellcolor{lightcoral}{-5.25} & \cellcolor{lightgreen}{0.80} & \cellcolor{lightcoral}{-5.46} & \cellcolor{lightcoral}{-4.71} & \cellcolor{lightgreen}{0.68} & \cellcolor{lightcoral}{-16.12} & \cellcolor{lightcoral}{-5.36} & \cellcolor{lightcoral}{-18.23} & \cellcolor{lightcoral}{-5.84} & \cellcolor{lightcoral}{-4.77} & \cellcolor{lightcoral}{-7.52} & \cellcolor{lightcoral}{-3.68} \\
\midrule
\multirow{4}{*}{\texttt{Qwen3 14B}} & ZS & \cellcolor{lightcoral}{-5.23} & \cellcolor{lightcoral}{-28.14} & 0.00 & \cellcolor{lightcoral}{-8.06} & \cellcolor{lightcoral}{-1.32} & 0.00 & \cellcolor{lightcoral}{-8.52} & \cellcolor{lightcoral}{-1.34} & \cellcolor{lightcoral}{-3.79} & \cellcolor{lightcoral}{-9.73} & \cellcolor{lightcoral}{-20.13} & \cellcolor{lightcoral}{-5.22} & \cellcolor{lightcoral}{-1.19} & \cellcolor{lightcoral}{-3.94} & \cellcolor{lightcoral}{-4.82} & \cellcolor{lightcoral}{-13.25} & \cellcolor{lightcoral}{-23.21} & \cellcolor{lightgreen}{6.59} & \cellcolor{lightcoral}{-2.23} & \cellcolor{lightcoral}{-19.79} & \cellcolor{lightcoral}{-7.67} \\
 & ZS + Task & \cellcolor{lightcoral}{-2.59} & \cellcolor{lightgreen}{7.23} & 0.00 & \cellcolor{lightgreen}{1.03} & \cellcolor{lightgreen}{1.58} & 0.00 & \cellcolor{lightcoral}{-2.92} & \cellcolor{lightgreen}{3.12} & \cellcolor{lightgreen}{6.89} & \cellcolor{lightgreen}{0.94} & \cellcolor{lightgreen}{3.39} & \cellcolor{lightgreen}{1.37} & \cellcolor{lightgreen}{4.74} & \cellcolor{lightcoral}{-0.03} & \cellcolor{lightcoral}{-6.02} & \cellcolor{lightcoral}{-1.58} & \cellcolor{lightcoral}{-16.76} & \cellcolor{lightcoral}{-5.84} & \cellcolor{lightgreen}{1.31} & \cellcolor{lightcoral}{-1.85} & \cellcolor{lightcoral}{-0.30} \\
& FS + Task & \cellcolor{lightgreen}{2.34} & \cellcolor{lightgreen}{1.91} & 0.00 & \cellcolor{lightgreen}{1.15} & \cellcolor{lightcoral}{-2.02} & 0.00 & \cellcolor{lightgreen}{1.02} & \cellcolor{lightcoral}{-0.16} & \cellcolor{lightcoral}{-0.89} & \cellcolor{lightgreen}{0.98} & \cellcolor{lightgreen}{2.51} & \cellcolor{lightgreen}{0.12} & \cellcolor{lightcoral}{-0.21} & \cellcolor{lightcoral}{-1.44} & \cellcolor{lightcoral}{-0.51} & \cellcolor{lightgreen}{3.47} & \cellcolor{lightcoral}{-4.55} & \cellcolor{lightgreen}{5.62} & \cellcolor{lightgreen}{0.70} & \cellcolor{lightgreen}{1.63} & \cellcolor{lightgreen}{0.58} \\
& CoT & \cellcolor{lightcoral}{-8.51} & \cellcolor{lightcoral}{-3.24} & 0.00 & \cellcolor{lightcoral}{-7.24} & \cellcolor{lightcoral}{-7.41} & 0.00 & \cellcolor{lightcoral}{-10.40} & \cellcolor{lightgreen}{3.60} & \cellcolor{lightgreen}{3.72} & \cellcolor{lightcoral}{-0.25} & \cellcolor{lightcoral}{-2.33} & \cellcolor{lightcoral}{-4.88} & \cellcolor{lightcoral}{-2.57} & \cellcolor{lightcoral}{-3.40} & \cellcolor{lightcoral}{-45.90} & \cellcolor{lightcoral}{-5.48} & \cellcolor{lightcoral}{-18.73} & \cellcolor{lightgreen}{2.80} & \cellcolor{lightgreen}{0.15} & \cellcolor{lightcoral}{-3.79} & \cellcolor{lightcoral}{-5.69} \\
\midrule
\multicolumn{2}{l}{\textit{Average}} & \cellcolor{lightcoral}{-3.50} & \cellcolor{lightcoral}{-5.56} & 0.00 & \cellcolor{lightcoral}{-3.28} & \cellcolor{lightcoral}{-2.29} & 0.00 & \cellcolor{lightcoral}{-5.21} & \cellcolor{lightgreen}{1.30} & \cellcolor{lightgreen}{1.48} & \cellcolor{lightcoral}{-2.02} & \cellcolor{lightcoral}{-4.14} & \cellcolor{lightcoral}{-2.15} & \cellcolor{lightgreen}{0.19} & \cellcolor{lightcoral}{-2.20} & \cellcolor{lightcoral}{-14.31} & \cellcolor{lightcoral}{-4.21} & \cellcolor{lightcoral}{-15.81} & \cellcolor{lightgreen}{2.29} & \cellcolor{lightcoral}{-0.02} & \cellcolor{lightcoral}{-5.95} & \cellcolor{lightcoral}{-3.27} \\
\midrule
\multirow{4}{*}{\texttt{Gemma3 12B}} & ZS & \cellcolor{lightcoral}{-1.82} & \cellcolor{lightcoral}{-11.50} & 0.00 & \cellcolor{lightgreen}{0.74} & \cellcolor{lightcoral}{-4.23} & 0.00 & \cellcolor{lightcoral}{-0.37} & \cellcolor{lightcoral}{-4.98} & \cellcolor{lightgreen}{11.45} & \cellcolor{lightgreen}{5.52} & \cellcolor{lightgreen}{0.32} & \cellcolor{lightgreen}{6.57} & \cellcolor{lightgreen}{4.79} & \cellcolor{lightgreen}{4.47} & \cellcolor{lightcoral}{-21.41} & \cellcolor{lightcoral}{-20.48} & \cellcolor{lightcoral}{-24.04} & \cellcolor{lightcoral}{-2.46} & \cellcolor{lightcoral}{-14.20} & \cellcolor{lightcoral}{-26.58} & \cellcolor{lightcoral}{-4.91} \\
& ZS + Task & \cellcolor{lightgreen}{2.42} & \cellcolor{lightgreen}{11.46} & 0.00 & \cellcolor{lightgreen}{2.03} & \cellcolor{lightgreen}{0.97} & 0.00 & \cellcolor{lightcoral}{-0.65} & \cellcolor{lightcoral}{-0.74} & \cellcolor{lightgreen}{9.22} & \cellcolor{lightgreen}{6.61} & \cellcolor{lightgreen}{11.72} & \cellcolor{lightgreen}{2.38} & \cellcolor{lightgreen}{0.18} & \cellcolor{lightgreen}{16.46} & \cellcolor{lightgreen}{4.87} & \cellcolor{lightcoral}{-1.22} & \cellcolor{lightcoral}{-6.46} & \cellcolor{lightgreen}{7.46} & \cellcolor{lightcoral}{-3.77} & \cellcolor{lightgreen}{0.60} & \cellcolor{lightgreen}{3.18} \\
& FS + Task & 0.00 & \cellcolor{lightgreen}{7.31} & 0.00 & \cellcolor{lightcoral}{-1.67} & \cellcolor{lightcoral}{-0.50} & 0.00 & \cellcolor{lightgreen}{2.17} & \cellcolor{lightgreen}{3.13} & \cellcolor{lightcoral}{-2.71} & \cellcolor{lightgreen}{2.64} & \cellcolor{lightgreen}{1.65} & \cellcolor{lightcoral}{-4.38} & \cellcolor{lightcoral}{-10.98} & \cellcolor{lightcoral}{-2.52} & \cellcolor{lightgreen}{7.07} & \cellcolor{lightgreen}{7.42} & \cellcolor{lightgreen}{7.82} & \cellcolor{lightcoral}{-0.99} & \cellcolor{lightcoral}{-2.20} & \cellcolor{lightgreen}{4.81} & \cellcolor{lightgreen}{0.90} \\
& CoT & \cellcolor{lightcoral}{-1.10} & \cellcolor{lightgreen}{3.98} & 0.00 & \cellcolor{lightcoral}{-3.40} & \cellcolor{lightgreen}{9.70} & 0.00 & \cellcolor{lightgreen}{0.50} & \cellcolor{lightgreen}{7.89} & \cellcolor{lightgreen}{5.29} & \cellcolor{lightgreen}{6.17} & \cellcolor{lightgreen}{5.56} & \cellcolor{lightcoral}{-3.84} & \cellcolor{lightcoral}{-2.50} & \cellcolor{lightgreen}{7.43} & 0.00 & \cellcolor{lightcoral}{-15.60} & \cellcolor{lightcoral}{-19.74} & \cellcolor{lightgreen}{0.44} & \cellcolor{lightgreen}{6.65} & \cellcolor{lightgreen}{1.04} & \cellcolor{lightgreen}{0.42} \\
\midrule
\multicolumn{2}{l}{\textit{Average}} & \cellcolor{lightcoral}{-0.12} & \cellcolor{lightgreen}{2.81} & 0.00 & \cellcolor{lightcoral}{-0.57} & \cellcolor{lightgreen}{1.49} & 0.00 & \cellcolor{lightgreen}{0.41} & \cellcolor{lightgreen}{1.32} & \cellcolor{lightgreen}{5.81} & \cellcolor{lightgreen}{5.23} & \cellcolor{lightgreen}{4.81} & \cellcolor{lightgreen}{0.18} & \cellcolor{lightcoral}{-2.13} & \cellcolor{lightgreen}{6.46} & \cellcolor{lightcoral}{-2.37} & \cellcolor{lightcoral}{-7.47} & \cellcolor{lightcoral}{-10.60} & \cellcolor{lightgreen}{1.11} & \cellcolor{lightcoral}{-3.38} & \cellcolor{lightcoral}{-5.03 }& \cellcolor{lightcoral}{-0.10} \\
\midrule
\multirow{4}{*}{\texttt{Gemma3 27B}} & ZS & \cellcolor{lightgreen}{0.92} & \cellcolor{lightcoral}{-5.19} & 0.00 & \cellcolor{lightcoral}{-3.14} & \cellcolor{lightcoral}{-3.58} & 0.00 & \cellcolor{lightcoral}{-1.25} & \cellcolor{lightcoral}{-4.17} & \cellcolor{lightgreen}{5.49} & \cellcolor{lightgreen}{1.82} & \cellcolor{lightcoral}{-3.51} & \cellcolor{lightgreen}{3.61} & \cellcolor{lightgreen}{1.90} & \cellcolor{lightcoral}{-0.06} & \cellcolor{lightcoral}{-11.90} & \cellcolor{lightcoral}{-5.17} & \cellcolor{lightcoral}{-21.93} & \cellcolor{lightcoral}{-3.78} & \cellcolor{lightcoral}{-10.53} & \cellcolor{lightcoral}{-7.77} & \cellcolor{lightcoral}{-3.41} \\
& ZS + Task & \cellcolor{lightgreen}{2.16} & \cellcolor{lightgreen}{13.16} & 0.00 & \cellcolor{lightgreen}{1.69} & \cellcolor{lightgreen}{8.45} & 0.00 & \cellcolor{lightgreen}{1.67} & \cellcolor{lightgreen}{9.12} & 0.00 & \cellcolor{lightcoral}{-1.20} & \cellcolor{lightgreen}{9.65} & \cellcolor{lightgreen}{1.69} & \cellcolor{lightgreen}{2.50} & \cellcolor{lightgreen}{2.06} & \cellcolor{lightgreen}{5.67} & \cellcolor{lightgreen}{2.42} & \cellcolor{lightgreen}{1.90} & \cellcolor{lightgreen}{5.99} & \cellcolor{lightcoral}{-2.57} & \cellcolor{lightgreen}{7.89} & \cellcolor{lightgreen}{3.61} \\
& FS + Task & \cellcolor{lightcoral}{-3.85} & \cellcolor{lightgreen}{3.05} & 0.00 & \cellcolor{lightcoral}{-0.69} & \cellcolor{lightcoral}{-0.38} & 0.00 & \cellcolor{lightcoral}{-2.30} & \cellcolor{lightcoral}{-3.35} & \cellcolor{lightcoral}{-3.38} & \cellcolor{lightgreen}{0.44} & \cellcolor{lightgreen}{1.76} & \cellcolor{lightcoral}{-5.25} & \cellcolor{lightcoral}{-1.28} & \cellcolor{lightgreen}{3.21} & \cellcolor{lightgreen}{0.73} & \cellcolor{lightcoral}{-0.92} & \cellcolor{lightcoral}{-4.66} & \cellcolor{lightcoral}{-0.96} & \cellcolor{lightcoral}{-2.89} & \cellcolor{lightcoral}{-1.73} & \cellcolor{lightcoral}{-1.12} \\
& CoT & \cellcolor{lightcoral}{-1.73} & \cellcolor{lightgreen}{10.25} & 0.00 & \cellcolor{lightcoral}{-3.79} & \cellcolor{lightgreen}{13.97} & 0.00 & \cellcolor{lightgreen}{1.74} & \cellcolor{lightgreen}{16.60} & \cellcolor{lightgreen}{3.00} & \cellcolor{lightgreen}{2.49} & \cellcolor{lightgreen}{5.23} & \cellcolor{lightgreen}{1.78} & \cellcolor{lightcoral}{-2.73} & \cellcolor{lightgreen}{3.03} & \cellcolor{lightgreen}{2.02} & \cellcolor{lightcoral}{-2.75} & \cellcolor{lightcoral}{-0.72} & \cellcolor{lightgreen}{2.86} & \cellcolor{lightgreen}{7.22} & \cellcolor{lightgreen}{6.39} & \cellcolor{lightgreen}{3.24} \\
\midrule
\multicolumn{2}{l}{\textit{Average}} & \cellcolor{lightcoral}{-0.62} & \cellcolor{lightgreen}{5.32} & 0.00 & \cellcolor{lightcoral}{-1.48} & \cellcolor{lightgreen}{4.61} & 0.00 & \cellcolor{lightcoral}{-0.03} & \cellcolor{lightgreen}{4.55} & \cellcolor{lightgreen}{1.28} & \cellcolor{lightgreen}{0.89} & \cellcolor{lightgreen}{3.28} & \cellcolor{lightgreen}{0.46} & \cellcolor{lightgreen}{0.10} & \cellcolor{lightgreen}{2.06} & \cellcolor{lightcoral}{-0.87} & \cellcolor{lightcoral}{-1.60} & \cellcolor{lightcoral}{-6.35} & \cellcolor{lightgreen}{1.03} & \cellcolor{lightcoral}{-2.19} & \cellcolor{lightgreen}{1.20} & \cellcolor{lightgreen}{0.58} \\
\midrule
\multirow{4}{*}{\texttt{Llama3.1 8B}} & ZS & \cellcolor{lightcoral}{-4.93} & \cellcolor{lightcoral}{-11.13} & 0.00 & \cellcolor{lightcoral}{-4.41} & \cellcolor{lightcoral}{-0.81} & 0.00 & \cellcolor{lightcoral}{-11.25} & \cellcolor{lightgreen}{7.94} & \cellcolor{lightgreen}{19.51} & \cellcolor{lightcoral}{-5.30} & \cellcolor{lightgreen}{8.06} & \cellcolor{lightgreen}{1.30} & \cellcolor{lightgreen}{6.03} & \cellcolor{lightcoral}{-0.34} & \cellcolor{lightcoral}{-3.07} & \cellcolor{lightcoral}{-5.74} & \cellcolor{lightcoral}{-11.87} & \cellcolor{lightgreen}{11.11} & \cellcolor{lightcoral}{-12.56} & \cellcolor{lightcoral}{-35.51} & \cellcolor{lightcoral}{-2.65} \\
 & ZS + Task & \cellcolor{lightcoral}{-3.83} & \cellcolor{lightcoral}{-26.70} & 0.00 & \cellcolor{lightcoral}{-2.63} & \cellcolor{lightcoral}{-0.52} & 0.00 & \cellcolor{lightgreen}{1.46} & \cellcolor{lightcoral}{-1.45} & \cellcolor{lightgreen}{14.13} & \cellcolor{lightcoral}{-4.80} & \cellcolor{lightgreen}{6.00} & \cellcolor{lightcoral}{-5.67} & \cellcolor{lightgreen}{0.61} & \cellcolor{lightgreen}{1.92} & \cellcolor{lightcoral}{-14.69} & \cellcolor{lightcoral}{-52.87} & \cellcolor{lightcoral}{-19.70} & \cellcolor{lightgreen}{3.02} & \cellcolor{lightgreen}{3.23} & \cellcolor{lightcoral}{-11.93} & \cellcolor{lightcoral}{-5.72} \\
 & FS + Task & \cellcolor{lightgreen}{7.16} & \cellcolor{lightgreen}{14.81} & 0.00 & \cellcolor{lightgreen}{9.86} & \cellcolor{lightgreen}{6.53} & 0.00 & \cellcolor{lightgreen}{4.07} & \cellcolor{lightgreen}{1.99} & \cellcolor{lightcoral}{-2.01} & \cellcolor{lightgreen}{2.39} & \cellcolor{lightgreen}{7.24} & \cellcolor{lightgreen}{4.84} & \cellcolor{lightcoral}{-1.85} & \cellcolor{lightgreen}{4.48} & \cellcolor{lightgreen}{2.56} & \cellcolor{lightgreen}{3.33} & \cellcolor{lightcoral}{-3.07} & \cellcolor{lightgreen}{12.93} & \cellcolor{lightgreen}{1.32} & \cellcolor{lightgreen}{2.72} & \cellcolor{lightgreen}{3.97} \\
& CoT & \cellcolor{lightcoral}{-11.19} & \cellcolor{lightcoral}{-21.48} & 0.00 & \cellcolor{lightcoral}{-8.18} & \cellcolor{lightcoral}{-1.09} & 0.00 & \cellcolor{lightcoral}{-7.09} & \cellcolor{lightgreen}{0.32} & \cellcolor{lightgreen}{11.17} & \cellcolor{lightcoral}{-2.21} & \cellcolor{lightgreen}{6.95} & \cellcolor{lightcoral}{-5.71} & \cellcolor{lightcoral}{-0.10} & \cellcolor{lightgreen}{0.18} & \cellcolor{lightcoral}{-14.46} & \cellcolor{lightcoral}{-10.59} & \cellcolor{lightcoral}{-9.79} & \cellcolor{lightgreen}{0.24} & \cellcolor{lightcoral}{-0.43} & \cellcolor{lightcoral}{-6.20} & \cellcolor{lightcoral}{-3.98} \\
\midrule
\multicolumn{2}{l}{\textit{Average}} & \cellcolor{lightcoral}{-3.20} & \cellcolor{lightcoral}{-11.12} & 0.00 & \cellcolor{lightcoral}{-1.34} & \cellcolor{lightgreen}{1.03} & 0.00 & \cellcolor{lightcoral}{-3.20} & \cellcolor{lightgreen}{2.20} & \cellcolor{lightgreen}{10.70} & \cellcolor{lightcoral}{-2.48} & \cellcolor{lightgreen}{7.06} & \cellcolor{lightcoral}{-1.31} & \cellcolor{lightgreen}{1.17} & \cellcolor{lightgreen}{1.56} & \cellcolor{lightcoral}{-7.42} & \cellcolor{lightcoral}{-16.47} & \cellcolor{lightcoral}{-11.11} & \cellcolor{lightgreen}{6.82} & \cellcolor{lightcoral}{-2.11} & \cellcolor{lightcoral}{-12.73} & \cellcolor{lightcoral}{-2.10} \\
\midrule
\multirow{4}{*}{\texttt{Llama3.1 70B}} & ZS & \cellcolor{lightcoral}{-1.68} & \cellcolor{lightgreen}{1.19} & 0.00 & \cellcolor{lightcoral}{-6.05} & \cellcolor{lightcoral}{-5.12} & 0.00 & \cellcolor{lightcoral}{-0.80} & \cellcolor{lightcoral}{-0.61} & \cellcolor{lightgreen}{4.11} & \cellcolor{lightcoral}{-12.28} & \cellcolor{lightcoral}{-1.74} & \cellcolor{lightgreen}{5.05} & \cellcolor{lightgreen}{17.41} & \cellcolor{lightgreen}{0.20} & \cellcolor{lightcoral}{-9.87} & \cellcolor{lightcoral}{-4.86} & \cellcolor{lightgreen}{47.48} & \cellcolor{lightcoral}{-5.40} & \cellcolor{lightcoral}{-1.25} & \cellcolor{lightcoral}{-16.73} & \cellcolor{lightgreen}{0.45} \\
& ZS + Task & \cellcolor{lightgreen}{2.01} & \cellcolor{lightgreen}{8.90} & 0.00 & \cellcolor{lightgreen}{0.44} & \cellcolor{lightgreen}{6.48} & 0.00 & \cellcolor{lightgreen}{3.08} & \cellcolor{lightgreen}{5.81} & \cellcolor{lightgreen}{13.35} & \cellcolor{lightgreen}{6.68} & \cellcolor{lightgreen}{8.74} & \cellcolor{lightgreen}{4.68} & \cellcolor{lightgreen}{7.19} & \cellcolor{lightgreen}{4.49} & \cellcolor{lightcoral}{-5.68} & \cellcolor{lightcoral}{-0.96} & \cellcolor{lightgreen}{50.44} & \cellcolor{lightcoral}{-21.55} & \cellcolor{lightcoral}{-6.49} & \cellcolor{lightcoral}{-27.52} & \cellcolor{lightgreen}{3.00} \\
& FS + Task & \cellcolor{lightcoral}{-7.17} & \cellcolor{lightcoral}{-10.28} & 0.00 & \cellcolor{lightcoral}{-2.34} & \cellcolor{lightcoral}{-3.99} & 0.00 & \cellcolor{lightcoral}{-10.93} & \cellcolor{lightgreen}{4.49} & \cellcolor{lightcoral}{-10.96} & \cellcolor{lightcoral}{-8.07} & \cellcolor{lightgreen}{6.31} & \cellcolor{lightgreen}{1.02} & \cellcolor{lightgreen}{0.64} & \cellcolor{lightcoral}{-0.70} & \cellcolor{lightcoral}{-4.68} & \cellcolor{lightcoral}{-2.97} & \cellcolor{lightcoral}{-0.84} & \cellcolor{lightcoral}{-10.55} & \cellcolor{lightgreen}{0.69} & \cellcolor{lightgreen}{3.83} & \cellcolor{lightcoral}{-2.82} \\
& CoT & \cellcolor{lightcoral}{-10.82} & \cellcolor{lightcoral}{-3.62} & 0.00 & \cellcolor{lightcoral}{-6.89} & \cellcolor{lightgreen}{4.52} & 0.00 & \cellcolor{lightcoral}{-5.77} & \cellcolor{lightgreen}{6.76} & \cellcolor{lightgreen}{3.98} & \cellcolor{lightgreen}{4.79} & \cellcolor{lightgreen}{1.86} & \cellcolor{lightcoral}{-5.59} & \cellcolor{lightgreen}{1.66} & \cellcolor{lightgreen}{3.74} & \cellcolor{lightcoral}{-4.41} & \cellcolor{lightcoral}{-15.21} & \cellcolor{lightcoral}{-0.14} & \cellcolor{lightcoral}{-0.80} & \cellcolor{lightgreen}{1.17} & \cellcolor{lightcoral}{-12.91} & \cellcolor{lightcoral}{-1.88} \\
\midrule
\multicolumn{2}{l}{\textit{Average}} & \cellcolor{lightcoral}{-2.28} & \cellcolor{lightcoral}{-0.06} & 0.00 & \cellcolor{lightcoral}{-2.65} & \cellcolor{lightcoral}{-0.88} & 0.00 & \cellcolor{lightcoral}{-2.88} & \cellcolor{lightgreen}{3.23} & \cellcolor{lightgreen}{2.16} & \cellcolor{lightcoral}{-4.56} & \cellcolor{lightgreen}{4.43} & \cellcolor{lightgreen}{3.59} & \cellcolor{lightgreen}{8.42} & \cellcolor{lightgreen}{1.33} & \cellcolor{lightcoral}{-6.74} & \cellcolor{lightcoral}{-2.93} & \cellcolor{lightgreen}{32.36} & \cellcolor{lightcoral}{-12.50} & \cellcolor{lightcoral}{-2.35} & \cellcolor{lightcoral}{-13.47} & \cellcolor{lightgreen}{0.21} \\
\bottomrule
\end{tabular}}
\setlength{\fboxsep}{1pt}
\caption{Difference in Macro F1 score when using post-language instructions versus English in cross-lingual settings. Each column represents a language pair (post–claim language), where the first language indicates the instruction language. Positive values (\colorbox{lightgreen}{green}) indicate improved performance when instructions are given in the post language. The table presents results for six models and four prompting techniques, with average values computed for each model–prompting combination as well as for each language pair.}
\label{tab:post-langauge}
\end{table*}

\begin{table*}[]
\centering
\resizebox{\textwidth}{!}{%
\begin{tabular}{lllllllllllllllllllllll}
\toprule
\textbf{Model} & \textbf{Technique} & \textbf{spa-eng} & \textbf{hin-eng} & \textbf{eng-ara} & \textbf{fra-eng} & \textbf{deu-eng} & \textbf{eng-por} & \textbf{spa-por} & \textbf{deu-fra} & \textbf{slk-ces} & \textbf{slk-eng} & \textbf{pol-hbs} & \textbf{ces-eng} & \textbf{ces-pol} & \textbf{nld-deu} & \textbf{msa-ara} & \textbf{kor-eng} & \textbf{mya-msa} & \textbf{ara-fra} & \textbf{hun-pol} & \textbf{tha-por} & \textbf{Avg.} \\
\midrule
\multirow{4}{*}{\texttt{Qwen3 8B}} & ZS & 0.00 & 0.00 & \cellcolor{lightgreen}{1.75} & 0.00 & 0.00 & \cellcolor{lightcoral}{-8.86} & \cellcolor{lightgreen}{3.72} & \cellcolor{lightgreen}{8.58} & \cellcolor{lightgreen}{0.62} & 0.00 & \cellcolor{lightcoral}{-4.45} & 0.00 & \cellcolor{lightcoral}{-1.11} & \cellcolor{lightgreen}{0.42} & \cellcolor{lightcoral}{-0.56} & 0.00 & \cellcolor{lightcoral}{-13.95} & \cellcolor{lightgreen}{1.29} & \cellcolor{lightcoral}{-8.97} & \cellcolor{lightcoral}{-6.88} & \cellcolor{lightcoral}{-1.42} \\
& ZS + Task & 0.00 & 0.00 & \cellcolor{lightcoral}{-8.15} & 0.00 & 0.00 & \cellcolor{lightcoral}{-8.57} & \cellcolor{lightcoral}{-5.97} & \cellcolor{lightcoral}{-2.00} & \cellcolor{lightgreen}{5.15} & 0.00 & \cellcolor{lightgreen}{2.10} & 0.00 & \cellcolor{lightgreen}{2.34} & \cellcolor{lightgreen}{1.76} & \cellcolor{lightcoral}{-11.01} & 0.00 & \cellcolor{lightcoral}{-11.74} & \cellcolor{lightcoral}{-6.68} & \cellcolor{lightgreen}{4.51} & \cellcolor{lightcoral}{-9.84} & \cellcolor{lightcoral}{-2.40} \\
& FS + Task & 0.00 & 0.00 & \cellcolor{lightgreen}{5.06} & 0.00 & 0.00 & \cellcolor{lightgreen}{4.77} & \cellcolor{lightgreen}{4.35} & \cellcolor{lightgreen}{3.01} & \cellcolor{lightcoral}{-11.66} & 0.00 & \cellcolor{lightgreen}{8.19} & 0.00 & \cellcolor{lightcoral}{-4.01} & \cellcolor{lightgreen}{2.38} & \cellcolor{lightcoral}{-6.75} & 0.00 & \cellcolor{lightgreen}{0.06} & \cellcolor{lightgreen}{2.57} & \cellcolor{lightgreen}{1.82} & \cellcolor{lightgreen}{9.74} & \cellcolor{lightgreen}{0.98} \\
& CoT & 0.00 & 0.00 & \cellcolor{lightcoral}{-10.48} & 0.00 & 0.00 & \cellcolor{lightcoral}{-0.50} & \cellcolor{lightcoral}{-6.10} & \cellcolor{lightgreen}{7.38} & \cellcolor{lightcoral}{-9.10} & 0.00 & \cellcolor{lightcoral}{-5.78} & 0.00 & \cellcolor{lightcoral}{-6.42} & \cellcolor{lightcoral}{-6.65} & \cellcolor{lightcoral}{-8.02} & 0.00 & \cellcolor{lightcoral}{-23.56} & \cellcolor{lightcoral}{-8.94} & \cellcolor{lightgreen}{0.67} & \cellcolor{lightcoral}{-8.09} & \cellcolor{lightcoral}{-4.28} \\
\midrule
\multicolumn{2}{l}{\textit{Average}} & 0.00 & 0.00 & \cellcolor{lightcoral}{-2.95} & 0.00 & 0.00 & \cellcolor{lightcoral}{-3.29} & \cellcolor{lightcoral}{-1.00} & \cellcolor{lightgreen}{4.24} & \cellcolor{lightcoral}{-3.75} & 0.00 & \cellcolor{lightgreen}{0.01} & 0.00 & \cellcolor{lightcoral}{-2.30} & \cellcolor{lightcoral}{-0.52} & \cellcolor{lightcoral}{-6.59} & 0.00 & \cellcolor{lightcoral}{-12.30} & \cellcolor{lightcoral}{-2.94} & \cellcolor{lightcoral}{-0.49} & \cellcolor{lightcoral}{-3.77} & \cellcolor{lightcoral}{-1.78} \\
\midrule
\multirow{4}{*}{\texttt{Qwen3 14B}}  & ZS & 0.00 & 0.00 & \cellcolor{lightcoral}{-8.33} & 0.00 & 0.00 & \cellcolor{lightcoral}{-17.89} & \cellcolor{lightcoral}{-7.83} & \cellcolor{lightcoral}{-10.59} & \cellcolor{lightgreen}{2.84} & 0.00 & \cellcolor{lightgreen}{3.03} & 0.00 & \cellcolor{lightcoral}{-19.43} & \cellcolor{lightgreen}{3.23} & \cellcolor{lightcoral}{-0.57} & 0.00 & \cellcolor{lightcoral}{-7.10} & \cellcolor{lightcoral}{-8.00} & \cellcolor{lightcoral}{-16.27} & \cellcolor{lightcoral}{-11.94} & \cellcolor{lightcoral}{-4.94} \\
& ZS + Task & 0.00 & 0.00 & \cellcolor{lightgreen}{5.32} & 0.00 & 0.00 & \cellcolor{lightcoral}{-1.89} & \cellcolor{lightcoral}{-4.54} & \cellcolor{lightgreen}{1.21} & \cellcolor{lightgreen}{8.40} & 0.00 & \cellcolor{lightgreen}{3.61} & 0.00 & \cellcolor{lightgreen}{4.62} & \cellcolor{lightgreen}{1.43} & \cellcolor{lightgreen}{2.03} & 0.00 & \cellcolor{lightcoral}{-10.64} & \cellcolor{lightcoral}{-0.60} & \cellcolor{lightgreen}{5.19} & \cellcolor{lightgreen}{1.96} & \cellcolor{lightgreen}{0.81} \\
 & FS + Task & 0.00 & 0.00 & \cellcolor{lightcoral}{-1.02} & 0.00 & 0.00 & \cellcolor{lightcoral}{-2.03} & \cellcolor{lightcoral}{-7.28} & \cellcolor{lightgreen}{1.93} & \cellcolor{lightcoral}{-0.88} & 0.00 & \cellcolor{lightgreen}{1.69} & 0.00 & \cellcolor{lightgreen}{4.71} & \cellcolor{lightcoral}{-7.13} & \cellcolor{lightgreen}{3.58} & 0.00 & \cellcolor{lightcoral}{-7.46} & \cellcolor{lightgreen}{3.86} & \cellcolor{lightgreen}{2.54} & \cellcolor{lightcoral}{-0.17} & \cellcolor{lightcoral}{-0.38} \\
 & CoT & 0.00 & 0.00 & \cellcolor{lightcoral}{-5.75} & 0.00 & 0.00 & \cellcolor{lightcoral}{-5.91} & \cellcolor{lightcoral}{-15.33} & \cellcolor{lightcoral}{-1.58} & \cellcolor{lightcoral}{-1.46} & 0.00 & \cellcolor{lightcoral}{-3.62} & 0.00 & \cellcolor{lightcoral}{-1.50} & \cellcolor{lightgreen}{0.40} & \cellcolor{lightcoral}{-8.35} & 0.00 & \cellcolor{lightcoral}{-16.72} & \cellcolor{lightcoral}{-2.91} & \cellcolor{lightgreen}{0.73} & \cellcolor{lightcoral}{-0.18} & \cellcolor{lightcoral}{-3.11} \\
\midrule
\multicolumn{2}{l}{\textit{Average}} & 0.00 & 0.00 & \cellcolor{lightcoral}{-2.44} & 0.00 & 0.00 & \cellcolor{lightcoral}{-6.93} & \cellcolor{lightcoral}{-8.74} & \cellcolor{lightcoral}{-2.26} & \cellcolor{lightgreen}{2.22} & 0.00 & \cellcolor{lightgreen}{1.17} & 0.00 & \cellcolor{lightcoral}{-2.90} &\cellcolor{lightcoral}{-0.52} & \cellcolor{lightcoral}{-0.83} & 0.00 & \cellcolor{lightcoral}{-10.48} & \cellcolor{lightcoral}{-1.91} & \cellcolor{lightcoral}{-1.95} & \cellcolor{lightcoral}{-2.58} & \cellcolor{lightcoral}{-1.91} \\
\midrule
\multirow{4}{*}{\texttt{Gemma3 12B}} & ZS & 0.00 & 0.00 & \cellcolor{lightcoral}{-2.78} & 0.00 & 0.00 & \cellcolor{lightcoral}{-1.46} & \cellcolor{lightcoral}{-1.57} & \cellcolor{lightgreen}{3.16} & \cellcolor{lightgreen}{11.62} & 0.00 & \cellcolor{lightgreen}{8.22} & 0.00 & \cellcolor{lightcoral}{-5.35} & \cellcolor{lightcoral}{-8.98} & \cellcolor{lightgreen}{0.13} & 0.00 & \cellcolor{lightcoral}{-26.98} & \cellcolor{lightcoral}{-4.63} & \cellcolor{lightcoral}{-7.85} & \cellcolor{lightcoral}{-5.03} & \cellcolor{lightcoral}{-2.08} \\
& ZS + Task & 0.00 & 0.00 & \cellcolor{lightgreen}{10.90} & 0.00 & 0.00 & \cellcolor{lightgreen}{2.10} & \cellcolor{lightgreen}{1.28} & \cellcolor{lightcoral}{-0.53} & \cellcolor{lightgreen}{2.12} & 0.00 & \cellcolor{lightgreen}{5.74} & 0.00 & \cellcolor{lightgreen}{11.85} & 0.00 & \cellcolor{lightgreen}{8.85} & 0.00 & \cellcolor{lightgreen}{1.18} & \cellcolor{lightcoral}{-1.68} & \cellcolor{lightgreen}{11.49} & \cellcolor{lightcoral}{-1.36} & \cellcolor{lightgreen}{2.60} \\
 & FS + Task & 0.00 & 0.00 & \cellcolor{lightcoral}{-3.35} & 0.00 & 0.00 & \cellcolor{lightcoral}{-2.03} & \cellcolor{lightgreen}{0.33} & \cellcolor{lightcoral}{-1.69} & \cellcolor{lightcoral}{-3.07} & 0.00 & \cellcolor{lightcoral}{-3.39} & 0.00 & \cellcolor{lightcoral}{-5.64} & \cellcolor{lightcoral}{-5.38} & \cellcolor{lightcoral}{-1.78} & 0.00 & \cellcolor{lightcoral}{-0.06} & \cellcolor{lightgreen}{1.25} & \cellcolor{lightgreen}{1.98} & \cellcolor{lightcoral}{-3.44} & \cellcolor{lightcoral}{-1.31} \\
 & CoT & 0.00 & 0.00 & \cellcolor{lightgreen}{6.24} & 0.00 & 0.00 & \cellcolor{lightcoral}{-6.82} & \cellcolor{lightcoral}{-3.47} & \cellcolor{lightgreen}{0.43} & \cellcolor{lightgreen}{3.56} & 0.00 & \cellcolor{lightgreen}{11.17} & 0.00 & \cellcolor{lightgreen}{2.14} & \cellcolor{lightgreen}{4.69} & \cellcolor{lightgreen}{5.97} & 0.00 & \cellcolor{lightcoral}{-2.91} & \cellcolor{lightcoral}{-2.86} & \cellcolor{lightgreen}{2.21} & \cellcolor{lightcoral}{-3.35} & \cellcolor{lightgreen}{0.85} \\
\midrule
\multicolumn{2}{l}{\textit{Average}} & 0.00 & 0.00 & \cellcolor{lightgreen}{2.75} & 0.00 & 0.00 & \cellcolor{lightcoral}{-2.05} & \cellcolor{lightcoral}{-0.86} & \cellcolor{lightgreen}{0.34} & \cellcolor{lightgreen}{3.56} & 0.00 & \cellcolor{lightgreen}{5.43} & 0.00 & \cellcolor{lightgreen}{0.75} & \cellcolor{lightcoral}{-2.42} & \cellcolor{lightgreen}{3.29} & 0.00 & \cellcolor{lightcoral}{-7.19} & \cellcolor{lightcoral}{-1.98} & \cellcolor{lightgreen}{1.96} & \cellcolor{lightcoral}{-3.30} & \cellcolor{lightgreen}{0.01} \\
\midrule
\multirow{4}{*}{\texttt{Gemma3 27B}} & ZS & 0.00 & 0.00 & \cellcolor{lightcoral}{-1.26} & 0.00 & 0.00 & \cellcolor{lightcoral}{-3.21} & \cellcolor{lightgreen}{0.79} & \cellcolor{lightgreen}{10.19} & \cellcolor{lightgreen}{10.15} & 0.00 & \cellcolor{lightgreen}{11.40} & 0.00 & \cellcolor{lightcoral}{-5.53} & \cellcolor{lightgreen}{0.72} & \cellcolor{lightcoral}{-0.76} & 0.00 & \cellcolor{lightcoral}{-4.61} & \cellcolor{lightcoral}{-0.92} & \cellcolor{lightcoral}{-2.56} & \cellcolor{lightcoral}{-4.44} & \cellcolor{lightgreen}{0.50} \\
& ZS + Task & 0.00 & 0.00 & \cellcolor{lightgreen}{8.97} & 0.00 & 0.00 & \cellcolor{lightgreen}{4.39} & \cellcolor{lightgreen}{2.22} & \cellcolor{lightgreen}{10.78} & \cellcolor{lightgreen}{6.24} & 0.00 & \cellcolor{lightgreen}{6.98} & 0.00 & \cellcolor{lightgreen}{10.78} & \cellcolor{lightgreen}{9.73} & \cellcolor{lightgreen}{6.15} & 0.00 & \cellcolor{lightgreen}{3.13} & \cellcolor{lightgreen}{4.38} & \cellcolor{lightgreen}{8.87} & \cellcolor{lightgreen}{0.52} & \cellcolor{lightgreen}{4.16} \\
& FS + Task & 0.00 & 0.00 & \cellcolor{lightcoral}{-0.91} & 0.00 & 0.00 & \cellcolor{lightcoral}{-0.15} & \cellcolor{lightcoral}{-3.30} & \cellcolor{lightcoral}{-3.11} & \cellcolor{lightgreen}{0.55} & 0.00 & \cellcolor{lightgreen}{2.26} & 0.00 & \cellcolor{lightgreen}{3.04} & \cellcolor{lightgreen}{5.28} & \cellcolor{lightcoral}{-1.91} & 0.00 & \cellcolor{lightgreen}{13.21} & \cellcolor{lightgreen}{1.91} & \cellcolor{lightgreen}{0.72} & \cellcolor{lightcoral}{-0.90} & \cellcolor{lightgreen}{0.83} \\
& CoT & 0.00 & 0.00 & \cellcolor{lightgreen}{10.69} & 0.00 & 0.00 & \cellcolor{lightgreen}{2.91} & \cellcolor{lightgreen}{1.23} & \cellcolor{lightgreen}{0.89} & \cellcolor{lightgreen}{5.18} & 0.00 & \cellcolor{lightcoral}{-0.72} & 0.00 & \cellcolor{lightgreen}{3.40} & \cellcolor{lightgreen}{16.60} & \cellcolor{lightgreen}{4.53} & 0.00 & \cellcolor{lightgreen}{1.07} & \cellcolor{lightcoral}{-7.23} & \cellcolor{lightgreen}{2.97} & \cellcolor{lightgreen}{0.34} & \cellcolor{lightgreen}{2.09} \\
\midrule
\multicolumn{2}{l}{\textit{Average}} & 0.00 & 0.00 & \cellcolor{lightgreen}{4.37} & 0.00 & 0.00 & \cellcolor{lightgreen}{0.99} & \cellcolor{lightgreen}{0.23} & \cellcolor{lightgreen}{4.69} & \cellcolor{lightgreen}{5.53} & 0.00 & \cellcolor{lightgreen}{4.98} & 0.00 & \cellcolor{lightgreen}{2.92} & \cellcolor{lightgreen}{8.08} & \cellcolor{lightgreen}{2.00} & 0.00 & \cellcolor{lightgreen}{3.20} & \cellcolor{lightcoral}{-0.47} & \cellcolor{lightgreen}{2.50} & \cellcolor{lightcoral}{-1.12} & \cellcolor{lightgreen}{1.90} \\
\midrule
\multirow{4}{*}{\texttt{Llama3.1 8B}}  & ZS & 0.00 & 0.00 & \cellcolor{lightgreen}{5.85} & 0.00 & 0.00 & \cellcolor{lightcoral}{-17.65} & \cellcolor{lightcoral}{-25.71} & \cellcolor{lightgreen}{7.68} & \cellcolor{lightgreen}{9.09} & 0.00 & \cellcolor{lightgreen}{18.61} & 0.00 & \cellcolor{lightgreen}{8.46} & \cellcolor{lightgreen}{10.36} & \cellcolor{lightgreen}{14.57} & 0.00 & \cellcolor{lightcoral}{-16.72} & \cellcolor{lightgreen}{0.68} & \cellcolor{lightgreen}{2.05} & \cellcolor{lightcoral}{-14.85} & \cellcolor{lightgreen}{0.12} \\
& ZS + Task & 0.00 & 0.00 & \cellcolor{lightgreen}{10.21} & 0.00 & 0.00 & \cellcolor{lightcoral}{-4.07} & \cellcolor{lightcoral}{-1.69} & \cellcolor{lightgreen}{1.52} & \cellcolor{lightgreen}{7.12} & 0.00 & \cellcolor{lightcoral}{-9.61} & 0.00 & \cellcolor{lightgreen}{1.63} & \cellcolor{lightcoral}{-0.21} & \cellcolor{lightcoral}{-1.47} & 0.00 & \cellcolor{lightcoral}{-28.40} & \cellcolor{lightgreen}{10.46} & \cellcolor{lightcoral}{-1.90} & \cellcolor{lightcoral}{-9.75} & \cellcolor{lightcoral}{-1.31} \\
& FS + Task & 0.00 & 0.00 & \cellcolor{lightgreen}{3.01} & 0.00 & 0.00 & \cellcolor{lightcoral}{-2.98} & \cellcolor{lightgreen}{2.33} & \cellcolor{lightgreen}{2.89} & \cellcolor{lightcoral}{-2.17} & 0.00 & \cellcolor{lightgreen}{4.71} & 0.00 & \cellcolor{lightgreen}{5.35} & \cellcolor{lightgreen}{3.07} & \cellcolor{lightgreen}{9.33} & 0.00 & \cellcolor{lightcoral}{-1.35} & \cellcolor{lightgreen}{1.33} & \cellcolor{lightgreen}{10.20} & \cellcolor{lightgreen}{2.70} & \cellcolor{lightgreen}{1.92} \\
& CoT & 0.00 & 0.00 & \cellcolor{lightgreen}{2.25} & 0.00 & 0.00 & \cellcolor{lightgreen}{0.79} & \cellcolor{lightgreen}{3.42} & \cellcolor{lightcoral}{-2.67} & \cellcolor{lightgreen}{3.23} & 0.00 & \cellcolor{lightcoral}{-0.95} & 0.00 & \cellcolor{lightgreen}{16.06} & \cellcolor{lightcoral}{-1.73} & \cellcolor{lightgreen}{3.04} & 0.00 & \cellcolor{lightcoral}{-14.75} & \cellcolor{lightgreen}{1.43} & \cellcolor{lightgreen}{10.46} & \cellcolor{lightgreen}{0.43} & \cellcolor{lightgreen}{1.05} \\
\midrule
\multicolumn{2}{l}{\textit{Average}} & 0.00 & 0.00 & \cellcolor{lightgreen}{5.33} & 0.00 & 0.00 & \cellcolor{lightcoral}{-5.98} & \cellcolor{lightcoral}{-5.41} & \cellcolor{lightgreen}{2.35} & \cellcolor{lightgreen}{4.32} & 0.00 & \cellcolor{lightgreen}{3.19} & 0.00 & \cellcolor{lightgreen}{7.88} & \cellcolor{lightgreen}{2.87} & \cellcolor{lightgreen}{6.37} & 0.00 & \cellcolor{lightcoral}{-15.31} & \cellcolor{lightgreen}{3.48} & \cellcolor{lightgreen}{5.20} & \cellcolor{lightcoral}{-5.37} & \cellcolor{lightgreen}{0.45} \\
\midrule
\multirow{4}{*}{\texttt{Llama3.1 70B}}  & ZS & 0.00 & 0.00 & \cellcolor{lightcoral}{-4.71} & 0.00 & 0.00 & \cellcolor{lightcoral}{-4.67} & \cellcolor{lightcoral}{-1.86} & \cellcolor{lightcoral}{-7.44} & \cellcolor{lightgreen}{9.53} & 0.00 & \cellcolor{lightcoral}{-1.34} & 0.00 & \cellcolor{lightgreen}{11.41} & \cellcolor{lightcoral}{-3.10} & \cellcolor{lightcoral}{-1.07} & 0.00 & \cellcolor{lightcoral}{-10.27} & \cellcolor{lightcoral}{-5.97} & \cellcolor{lightgreen}{0.79} & \cellcolor{lightcoral}{-9.69} & \cellcolor{lightcoral}{-1.42} \\
& ZS + Task & 0.00 & 0.00 & \cellcolor{lightcoral}{-6.85} & 0.00 & 0.00 & \cellcolor{lightgreen}{2.92} & \cellcolor{lightcoral}{-0.28} & \cellcolor{lightgreen}{9.52} & \cellcolor{lightgreen}{7.48} & 0.00 & \cellcolor{lightgreen}{2.99} & 0.00 & \cellcolor{lightgreen}{11.56} & \cellcolor{lightgreen}{7.31} & \cellcolor{lightcoral}{-9.06} & 0.00 & \cellcolor{lightcoral}{-6.78} & \cellcolor{lightgreen}{6.88} & \cellcolor{lightgreen}{7.18} & \cellcolor{lightcoral}{-3.62} & \cellcolor{lightgreen}{1.46} \\
& FS + Task & 0.00 & 0.00 & \cellcolor{lightcoral}{-3.34} & 0.00 & 0.00 & \cellcolor{lightgreen}{4.09} & \cellcolor{lightcoral}{-6.30} & \cellcolor{lightgreen}{6.28} & \cellcolor{lightcoral}{-8.85} & 0.00 & \cellcolor{lightgreen}{2.07} & 0.00 & \cellcolor{lightcoral}{-0.66} & \cellcolor{lightgreen}{2.60} & \cellcolor{lightcoral}{-6.74} & 0.00 & \cellcolor{lightcoral}{-5.06} & \cellcolor{lightgreen}{0.02} & \cellcolor{lightgreen}{4.58} & \cellcolor{lightcoral}{-3.47} & \cellcolor{lightcoral}{-0.74} \\
& CoT & 0.00 & 0.00 & \cellcolor{lightcoral}{-11.97} & 0.00 & 0.00 & \cellcolor{lightcoral}{-9.77} & \cellcolor{lightcoral}{-4.04} & \cellcolor{lightgreen}{1.54} & \cellcolor{lightcoral}{-2.48} & 0.00 & \cellcolor{lightgreen}{1.37} & 0.00 & \cellcolor{lightcoral}{-1.11} & \cellcolor{lightgreen}{4.76} & \cellcolor{lightcoral}{-8.32} & 0.00 & \cellcolor{lightcoral}{-13.07} & \cellcolor{lightcoral}{-1.66} & \cellcolor{lightgreen}{2.08} & \cellcolor{lightcoral}{-4.82} & \cellcolor{lightcoral}{-2.37} \\
\midrule
\multicolumn{2}{l}{\textit{Average}} & 0.00 & 0.00 & \cellcolor{lightcoral}{-6.72} & 0.00 & 0.00 & \cellcolor{lightcoral}{-1.86} & \cellcolor{lightcoral}{-3.12} & \cellcolor{lightgreen}{2.47} & \cellcolor{lightgreen}{1.42} & 0.00 & \cellcolor{lightgreen}{1.27} & 0.00 & \cellcolor{lightgreen}{5.30} & \cellcolor{lightgreen}{2.89} & \cellcolor{lightcoral}{-6.30} & 0.00 & \cellcolor{lightcoral}{-8.80} & \cellcolor{lightcoral}{-0.18} & \cellcolor{lightgreen}{3.66} & \cellcolor{lightcoral}{-5.40} & \cellcolor{lightcoral}{-0.77} \\
\bottomrule
\end{tabular}
}
\setlength{\fboxsep}{1pt}
\caption{Difference in Macro F1 score when using claim-language instructions versus English in cross-lingual settings. Each column represents a language pair (post–claim language), with the second language indicating the instruction language. Positive values (\colorbox{lightgreen}{green}) reflect improved performance when using claim-language instructions. The table presents results for six models and four prompting techniques, with average values computed for each model–prompting combination as well as for each language pair.}
\label{tab:claim-language}
\end{table*}

\begin{table*}
\resizebox{\textwidth}{!}{%
\begin{tabular}{lcrrrrrrrrrrrrrrrrrrrrr}
\toprule
\multicolumn{1}{c}{\textbf{Technique}} & \multicolumn{1}{c}{\textbf{Thinking}} & \multicolumn{1}{c}{\textbf{ara}} & \multicolumn{1}{c}{\textbf{bul}} & \multicolumn{1}{c}{\textbf{ces}} & \multicolumn{1}{c}{\textbf{deu}} & \multicolumn{1}{c}{\textbf{ell}} & \multicolumn{1}{c}{\textbf{eng}} & \multicolumn{1}{c}{\textbf{fra}} & \multicolumn{1}{c}{\textbf{hbs}} & \multicolumn{1}{c}{\textbf{hin}} & \multicolumn{1}{c}{\textbf{hun}} & \multicolumn{1}{c}{\textbf{kor}} & \multicolumn{1}{c}{\textbf{msa}} & \multicolumn{1}{c}{\textbf{mya}} & \multicolumn{1}{c}{\textbf{nld}} & \multicolumn{1}{c}{\textbf{pol}} & \multicolumn{1}{c}{\textbf{por}} & \multicolumn{1}{c}{\textbf{ron}} & \multicolumn{1}{c}{\textbf{slk}} & \multicolumn{1}{c}{\textbf{spa}} & \multicolumn{1}{c}{\textbf{tha}} & \multicolumn{1}{c}{\textbf{Avg.}} \\
\midrule
\multirow{2}{*}{Zero-Shot} & \xmark & 77.18 & 72.26 & 63.19 & 64.14 & 77.41 & 75.52 & 73.27 & 59.14 & 74.88 & 69.9 & 74.51 & 69.12 & 65.65 & 67.8 & 57.96 & 70.75 & 69.73 & 64.97 & 71.05 & 75.48 & 69.7 \\
& \checkmark & 81.73 & 90.87 & 72.1 & 73.08 & 80.47 & 80.19 & 83.39 & 60.37 & 80.94 & 75.1 & 82.23 & 84.9 & 68.63 & 74.24 & 67.82 & 72.1 & 76.93 & 74.76 & 76.65 & 87.31 & 77.19 \\
\midrule
\multirow{2}{*}{\makecell[l]{Zero-Shot +\\Task Description}} & \xmark & 88.31 & 88.95 & 68.97 & 78.9 & 82.31 & 78.26 & 80.39 & 63.24 & 79.79 & 78.22 & 76.54 & 76.02 & 82.47 & 75.56 & 67.48 & 78.13 & 75.13 & 76.22 & 80.63 & 80.61 & 77.81 \\
& \checkmark & 86.98 & 92.18 & 79.69 & 79.97 & 84.13 & 84.83 & 87.3 & 73.12 & 84.88 & 79.54 & 83.28 & 87.7 & 79.6 & 80.77 & 75.91 & 80.92 & 84.0 & 80.31 & 83.96 & 85.66 & 82.74 \\
\midrule
\multirow{2}{*}{\makecell[l]{Few-Shot +\\Task Description}} & \xmark & 81.22 & 79.9 & 65.33 & 66.93 & 74.05 & 67.22 & 79.05 & 64.24 & 77.18 & 67.09 & 73.75 & 64.57 & 72.1 & 68.27 & 61.49 & 71.12 & 68.67 & 64.83 & 76.11 & 70.86 & 70.7 \\
& \checkmark & 89.55 & 90.49 & 75.49 & 80.47 & 76.77 & 82.45 & 84.07 & 71.74 & 85.48 & 80.39 & 84.78 & 82.29 & 72.78 & 83.43 & 74.56 & 80.14 & 76.66 & 81.87 & 85.48 & 86.51 & 81.27 \\
\bottomrule
\end{tabular}}
\caption{Comparison of Macro F1 performance using English for the instruction with (\checkmark) and without (\xmark) thinking mode for the \texttt{Qwen3 8B} in a monolingual setting.}
\end{table*}

\begin{table*}[]
\resizebox{\textwidth}{!}{%
\begin{tabular}{lcrrrrrrrrrrrrrrrrrrrrr}
\toprule
\multicolumn{1}{c}{\textbf{Technique}} & \multicolumn{1}{c}{\textbf{Thinking}} & \multicolumn{1}{c}{\textbf{ara}} & \multicolumn{1}{c}{\textbf{bul}} & \multicolumn{1}{c}{\textbf{ces}} & \multicolumn{1}{c}{\textbf{deu}} & \multicolumn{1}{c}{\textbf{ell}} & \multicolumn{1}{c}{\textbf{eng}} & \multicolumn{1}{c}{\textbf{fra}} & \multicolumn{1}{c}{\textbf{hbs}} & \multicolumn{1}{c}{\textbf{hin}} & \multicolumn{1}{c}{\textbf{hun}} & \multicolumn{1}{c}{\textbf{kor}} & \multicolumn{1}{c}{\textbf{msa}} & \multicolumn{1}{c}{\textbf{mya}} & \multicolumn{1}{c}{\textbf{nld}} & \multicolumn{1}{c}{\textbf{pol}} & \multicolumn{1}{c}{\textbf{por}} & \multicolumn{1}{c}{\textbf{ron}} & \multicolumn{1}{c}{\textbf{slk}} & \multicolumn{1}{c}{\textbf{spa}} & \multicolumn{1}{c}{\textbf{tha}} & \multicolumn{1}{c}{\textbf{Avg.}} \\
\midrule
\multirow{2}{*}{Zero-Shot} & \xmark & \cellcolor{lightcoral}{-2.79} & \cellcolor{lightcoral}{-10.89} & \cellcolor{lightgreen}{0.43} & \cellcolor{lightgreen}{8.16} & \cellcolor{lightcoral}{-4.90} & 0.00 & \cellcolor{lightcoral}{-2.93} & \cellcolor{lightcoral}{-0.98} & \cellcolor{lightcoral}{-0.40} & \cellcolor{lightcoral}{-2.30} & \cellcolor{lightcoral}{-17.20} & \cellcolor{lightcoral}{-9.05} & \cellcolor{lightcoral}{-21.36} & \cellcolor{lightcoral}{-6.55} & \cellcolor{lightcoral}{-5.36} & \cellcolor{lightcoral}{-6.29} & \cellcolor{lightcoral}{-6.70} & \cellcolor{lightgreen}{1.89} & \cellcolor{lightcoral}{-8.33} & \cellcolor{lightcoral}{-24.67} & \cellcolor{lightcoral}{-6.01} \\
  & \checkmark & \cellcolor{lightcoral}{-0.08} & \cellcolor{lightcoral}{-8.17} & \cellcolor{lightgreen}{0.41} & \cellcolor{lightgreen}{2.97} & \cellcolor{lightcoral}{-5.84} & 0.00 & \cellcolor{lightcoral}{-4.79} & \cellcolor{lightgreen}{0.20} & \cellcolor{lightcoral}{-0.42} & \cellcolor{lightgreen}{3.73} & \cellcolor{lightcoral}{-0.38} & \cellcolor{lightcoral}{-3.29} & \cellcolor{lightcoral}{-17.27} & \cellcolor{lightgreen}{0.61} & \cellcolor{lightcoral}{-3.92} & \cellcolor{lightcoral}{-4.23} & \cellcolor{lightcoral}{-3.48} & \cellcolor{lightgreen}{4.78} & \cellcolor{lightcoral}{-4.11} & \cellcolor{lightcoral}{-6.69} & \cellcolor{lightcoral}{-2.50} \\
\midrule
\multirow{2}{*}{\makecell[l]{Zero-Shot +\\Task Description}} & \xmark & \cellcolor{lightcoral}{-1.13} & \cellcolor{lightcoral}{-0.18} & \cellcolor{lightgreen}{2.45} & \cellcolor{lightgreen}{2.79} & \cellcolor{lightcoral}{-1.68} & 0.00 & \cellcolor{lightgreen}{3.79} & \cellcolor{lightgreen}{2.50} & \cellcolor{lightcoral}{-3.14} & \cellcolor{lightcoral}{-3.19} & \cellcolor{lightcoral}{-6.89} & \cellcolor{lightcoral}{-3.44} & \cellcolor{lightcoral}{-19.68} & \cellcolor{lightgreen}{4.85} & \cellcolor{lightgreen}{3.68} & \cellcolor{lightcoral}{-2.71} & \cellcolor{lightcoral}{-1.11} & \cellcolor{lightcoral}{-3.58} & \cellcolor{lightgreen}{3.13} & \cellcolor{lightcoral}{-6.42} & \cellcolor{lightcoral}{-1.50} \\
 & \checkmark & \cellcolor{lightgreen}{3.81} & \cellcolor{lightgreen}{0.58} & \cellcolor{lightgreen}{0.89} & \cellcolor{lightgreen}{0.66} & \cellcolor{lightcoral}{-1.83} & 0.00 & \cellcolor{lightcoral}{-1.93} & \cellcolor{lightcoral}{-0.48} & \cellcolor{lightgreen}{0.48} & \cellcolor{lightgreen}{0.34} & \cellcolor{lightcoral}{-2.93} & \cellcolor{lightcoral}{-4.35} & \cellcolor{lightcoral}{-14.52} & \cellcolor{lightgreen}{1.90} & \cellcolor{lightgreen}{1.15} & \cellcolor{lightcoral}{-1.73} & \cellcolor{lightcoral}{-3.73} & \cellcolor{lightcoral}{-2.52} & \cellcolor{lightcoral}{-0.13} & \cellcolor{lightcoral}{-0.70} & \cellcolor{lightcoral}{-1.25} \\
\midrule
\multirow{2}{*}{\makecell[l]{Few-Shot +\\Task Description}} &  \xmark & \cellcolor{lightgreen}{4.36} & \cellcolor{lightgreen}{5.84} & \cellcolor{lightcoral}{-3.90} & \cellcolor{lightgreen}{6.86} & \cellcolor{lightcoral}{-10.00} & 0.00 & \cellcolor{lightcoral}{-4.76} & \cellcolor{lightgreen}{2.42} & \cellcolor{lightgreen}{7.17} & \cellcolor{lightgreen}{3.11} & \cellcolor{lightcoral}{-9.59} & \cellcolor{lightgreen}{1.04} & \cellcolor{lightcoral}{-22.66} & \cellcolor{lightgreen}{9.88} & \cellcolor{lightcoral}{-2.11} & \cellcolor{lightgreen}{4.62} & \cellcolor{lightgreen}{10.37} & \cellcolor{lightcoral}{-0.83} & \cellcolor{lightgreen}{2.83} & \cellcolor{lightgreen}{10.65} & \cellcolor{lightgreen}{0.76} \\
 & \checkmark & \cellcolor{lightcoral}{-3.10} & \cellcolor{lightgreen}{1.79} & \cellcolor{lightgreen}{0.28} & \cellcolor{lightcoral}{-1.78} & \cellcolor{lightgreen}{4.31} & 0.00 & \cellcolor{lightcoral}{-0.17} & \cellcolor{lightcoral}{-1.37} & \cellcolor{lightgreen}{0.65} & \cellcolor{lightgreen}{5.47} & \cellcolor{lightcoral}{-4.73} & \cellcolor{lightcoral}{-3.15} & \cellcolor{lightcoral}{-11.55} & \cellcolor{lightcoral}{-3.35} & \cellcolor{lightcoral}{-1.36} & \cellcolor{lightcoral}{-0.62} & \cellcolor{lightgreen}{3.51} & \cellcolor{lightcoral}{-1.12} & \cellcolor{lightcoral}{-2.35} & \cellcolor{lightcoral}{-0.64} & \cellcolor{lightcoral}{-0.96} \\
\bottomrule
\end{tabular}}
\setlength{\fboxsep}{1pt}
\caption{Impact of the thinking mode on performance in monolingual settings across three prompting techniques for the \texttt{Qwen3 8B}. The table shows the difference in Macro F1 score when using target language instructions versus English. Each row compares performance with (\checkmark) and without (\xmark) "thinking mode", across 20 languages. Positive values (\colorbox{lightgreen}{green}) indicate improved performance with target language instructions. Average scores are reported in the final column.}
\label{tab:thinking-monolingual}
\end{table*}

\begin{table*}
\resizebox{\textwidth}{!}{%
\begin{tabular}{lcrrrrrrrrrrrrrrrrrrrrr}
\toprule
 \textbf{Technique} & \textbf{Thinking} & \textbf{spa-eng} & \textbf{hin-eng} & \textbf{eng-ara} & \textbf{fra-eng} & \textbf{deu-eng} & \textbf{eng-por} & \textbf{spa-por} & \textbf{deu-fra} & \textbf{slk-ces} & \textbf{slk-eng} & \textbf{pol-hbs} & \textbf{ces-eng} & \textbf{ces-pol} & \textbf{nld-deu} & \textbf{msa-ara} & \textbf{kor-eng} & \textbf{mya-msa} & \textbf{ara-fra} & \textbf{hun-pol} & \textbf{tha-por} & \textbf{Avg.} \\
\midrule
\multirow{2}{*}{Zero-Shot} & \xmark & 68.93 & 70.63 & 63.07 & 61.95 & 62.53 & 62.79 & 56.81 & 63.81 & 55.33 & 65.66 & 57.51 & 63.18 & 50.41 & 59.14 & 50.12 & 62.21 & 45.24 & 65.4 & 66.88 & 53.31 & 60.25 \\
& \checkmark & 78.39 & 69.36 & 65.59 & 70.61 & 68.47 & 67.75 & 55.17 & 66.7 & 63.5 & 75.65 & 60.61 & 71.96 & 57.94 & 61.89 & 55.74 & 75.2 & 57.93 & 69.0 & 69.37 & 64.9 & 66.29 \\
\midrule
\multirow{2}{*}{\makecell[l]{Zero-Shot +\\Task Description}} & \xmark & 80.93 & 84.45 & 78.01 & 71.11 & 72.93 & 69.69 & 64.19 & 73.3 & 57.43 & 75.06 & 60.87 & 69.56 & 64.84 & 68.57 & 65.03 & 71.24 & 60.67 & 74.09 & 74.31 & 71.73 & 70.4 \\
& \checkmark & 85.16 & 81.52 & 76.53 & 80.37 & 76.11 & 76.04 & 62.43 & 75.78 & 70.29 & 82.69 & 67.03 & 76.15 & 69.05 & 71.93 & 66.85 & 78.39 & 66.16 & 72.17 & 79.14 & 74.73 & 74.43 \\
\midrule
\multirow{2}{*}{\makecell[l]{Few-Shot +\\Task Description}} & \xmark & 68.88 & 71.49 & 80.8 & 62.24 & 59.65 & 57.25 & 58.79 & 67.79 & 57.58 & 67.1 & 56.85 & 60.57 & 66.09 & 61.95 & 58.3 & 68.14 & 49.81 & 77.29 & 67.25 & 58.95 & 63.84 \\
& \checkmark & 80.62 & 84.92 & 77.84 & 76.14 & 75.41 & 70.72 & 66.43 & 75.18 & 69.1 & 76.2 & 70.8 & 74.5 & 70.95 & 74.65 & 67.45 & 78.43 & 58.97 & 68.21 & 76.14 & 71.75 & 73.22 \\
\bottomrule
\end{tabular}}
\caption{Macro F1 performance using English for the instruction with (\checkmark) and without (\xmark) thinking mode for the \texttt{Qwen3 8B} in a cross-lingual setting.}
\end{table*}

\begin{table*}
\resizebox{\textwidth}{!}{%
\begin{tabular}{lcrrrrrrrrrrrrrrrrrrrrr}
\toprule
 \textbf{Technique} & \textbf{Thinking} & \textbf{spa-eng} & \textbf{hin-eng} & \textbf{eng-ara} & \textbf{fra-eng} & \textbf{deu-eng} & \textbf{eng-por} & \textbf{spa-por} & \textbf{deu-fra} & \textbf{slk-ces} & \textbf{slk-eng} & \textbf{pol-hbs} & \textbf{ces-eng} & \textbf{ces-pol} & \textbf{nld-deu} & \textbf{msa-ara} & \textbf{kor-eng} & \textbf{mya-msa} & \textbf{ara-fra} & \textbf{hun-pol} & \textbf{tha-por} & \textbf{Avg.} \\
\midrule
\multirow{2}{*}{Zero-Shot} & \xmark & \cellcolor{lightcoral}{-3.16} & \cellcolor{lightcoral}{-6.06} & 0.00 & \cellcolor{lightcoral}{-2.71} & \cellcolor{lightgreen}{2.73} & 0.00 & \cellcolor{lightcoral}{-2.47} & \cellcolor{lightgreen}{5.14} & \cellcolor{lightgreen}{0.95} & \cellcolor{lightcoral}{-6.50} & \cellcolor{lightcoral}{-5.14} & \cellcolor{lightcoral}{-5.66} & \cellcolor{lightcoral}{-0.91} & \cellcolor{lightcoral}{-5.25} & \cellcolor{lightcoral}{-20.79} & \cellcolor{lightcoral}{-8.90} & \cellcolor{lightcoral}{-17.21} & \cellcolor{lightgreen}{3.23} & \cellcolor{lightcoral}{-8.52} & \cellcolor{lightcoral}{-18.15} & \cellcolor{lightcoral}{-4.97} \\
& \checkmark & \cellcolor{lightcoral}{-1.67} & \cellcolor{lightgreen}{4.03} & 0.00 & \cellcolor{lightcoral}{-3.89} & \cellcolor{lightgreen}{4.50} & 0.00 & \cellcolor{lightcoral}{-4.16} & \cellcolor{lightgreen}{0.61} & \cellcolor{lightgreen}{1.65} & \cellcolor{lightcoral}{-0.36} & \cellcolor{lightcoral}{-0.36} & \cellcolor{lightcoral}{-4.53} & \cellcolor{lightgreen}{1.23} & \cellcolor{lightcoral}{-2.01} & \cellcolor{lightcoral}{-9.06} & \cellcolor{lightcoral}{-0.58} & \cellcolor{lightcoral}{-22.69} & \cellcolor{lightcoral}{-0.37} & \cellcolor{lightcoral}{-2.84} & \cellcolor{lightcoral}{-5.15} & \cellcolor{lightcoral}{-2.28} \\
\midrule
\multirow{2}{*}{\makecell[l]{Zero-Shot +\\Task Description}} & \xmark & \cellcolor{lightgreen}{2.02} & \cellcolor{lightcoral}{-0.66} & 0.00 & \cellcolor{lightgreen}{0.49} & \cellcolor{lightgreen}{2.26} & 0.00 & \cellcolor{lightgreen}{4.02} & \cellcolor{lightcoral}{-2.26} & \cellcolor{lightcoral}{-0.43} & \cellcolor{lightcoral}{-2.81} & \cellcolor{lightgreen}{4.90} & \cellcolor{lightcoral}{-0.08} & \cellcolor{lightgreen}{3.48} & \cellcolor{lightgreen}{3.34} & \cellcolor{lightcoral}{-16.01} & \cellcolor{lightgreen}{1.78} & \cellcolor{lightcoral}{-11.74} & \cellcolor{lightcoral}{-4.12} & \cellcolor{lightcoral}{-8.10} & \cellcolor{lightcoral}{-10.99} & \cellcolor{lightcoral}{-1.75} \\
 & \checkmark & \cellcolor{lightcoral}{-3.26} & \cellcolor{lightgreen}{6.52} & 0.00 & \cellcolor{lightcoral}{-6.35} & \cellcolor{lightgreen}{0.38} & 0.00 & \cellcolor{lightcoral}{-0.53} & \cellcolor{lightcoral}{-1.77} & \cellcolor{lightgreen}{3.42} & \cellcolor{lightcoral}{-1.76} & \cellcolor{lightgreen}{4.94} & \cellcolor{lightgreen}{0.53} & \cellcolor{lightcoral}{-0.54} & \cellcolor{lightgreen}{2.86} & \cellcolor{lightcoral}{-10.78} & \cellcolor{lightcoral}{-2.14} & \cellcolor{lightcoral}{-13.33} & \cellcolor{lightcoral}{-0.22} & \cellcolor{lightcoral}{-5.95} & \cellcolor{lightcoral}{-1.12} & \cellcolor{lightcoral}{-1.45} \\
\midrule
\multirow{2}{*}{\makecell[l]{Few-Shot +\\Task Description}} & \xmark & \cellcolor{lightgreen}{7.74} & \cellcolor{lightgreen}{14.28} & 0.00 & \cellcolor{lightgreen}{6.78} & \cellcolor{lightgreen}{4.75} & 0.00 & \cellcolor{lightgreen}{10.24} & \cellcolor{lightgreen}{7.09} & \cellcolor{lightcoral}{-7.30} & \cellcolor{lightcoral}{-7.15} & \cellcolor{lightgreen}{5.79} & \cellcolor{lightcoral}{-5.31} & \cellcolor{lightcoral}{-11.24} & \cellcolor{lightgreen}{8.09} & \cellcolor{lightcoral}{-6.00} & \cellcolor{lightcoral}{-9.00} & \cellcolor{lightcoral}{-11.70} & \cellcolor{lightcoral}{-5.68} & \cellcolor{lightgreen}{0.99} & \cellcolor{lightgreen}{7.72} & \cellcolor{lightgreen}{0.51} \\
& \checkmark & \cellcolor{lightgreen}{0.75} & \cellcolor{lightgreen}{2.95} & 0.00 & \cellcolor{lightgreen}{2.02} & \cellcolor{lightgreen}{1.72} & 0.00 & \cellcolor{lightgreen}{8.06} & \cellcolor{lightgreen}{3.36} & \cellcolor{lightgreen}{1.06} & \cellcolor{lightgreen}{2.23} & \cellcolor{lightgreen}{3.02} & \cellcolor{lightcoral}{-1.16} & \cellcolor{lightgreen}{3.37} & \cellcolor{lightgreen}{3.65} & \cellcolor{lightcoral}{-18.42} & \cellcolor{lightcoral}{-0.78} & \cellcolor{lightcoral}{-11.33} & \cellcolor{lightgreen}{2.71} & \cellcolor{lightcoral}{-0.69} & \cellcolor{lightcoral}{-5.46} & \cellcolor{lightcoral}{-0.15} \\
\bottomrule
\end{tabular}}
\setlength{\fboxsep}{1pt}
\caption{Impact of the thinking mode on performance in cross-lingual settings across three prompting techniques for the \texttt{Qwen3 8B}. The table shows the difference in the Macro F1 score when using post-language instructions (the first language in the column name) versus English. Each row compares performance with (\checkmark) and without (\xmark) "thinking mode", across 20 language pairs. Positive values (\colorbox{lightgreen}{green}) indicate improved performance with post-language instructions. Average scores are reported in the final column.}
\label{tab:thinking-post-language}
\end{table*}

\begin{table*}
\resizebox{\textwidth}{!}{%
\begin{tabular}{lcrrrrrrrrrrrrrrrrrrrrr}
\toprule
 \textbf{Technique} & \textbf{Thinking} & \textbf{spa-eng} & \textbf{hin-eng} & \textbf{eng-ara} & \textbf{fra-eng} & \textbf{deu-eng} & \textbf{eng-por} & \textbf{spa-por} & \textbf{deu-fra} & \textbf{slk-ces} & \textbf{slk-eng} & \textbf{pol-hbs} & \textbf{ces-eng} & \textbf{ces-pol} & \textbf{nld-deu} & \textbf{msa-ara} & \textbf{kor-eng} & \textbf{mya-msa} & \textbf{ara-fra} & \textbf{hun-pol} & \textbf{tha-por} & \textbf{Avg.} \\
\midrule
\multirow{2}{*}{Zero-Shot} & \xmark & 0.00 & 0.00 & \cellcolor{lightgreen}{1.75} & 0.00 & 0.00 & \cellcolor{lightcoral}{-8.86} & \cellcolor{lightgreen}{3.72} & \cellcolor{lightgreen}{8.58} & \cellcolor{lightgreen}{0.62} & 0.00 & \cellcolor{lightcoral}{-4.45} & 0.00 & \cellcolor{lightcoral}{-1.11} & \cellcolor{lightgreen}{0.42} & \cellcolor{lightcoral}{-0.56} & 0.00 & \cellcolor{lightcoral}{-13.95} & \cellcolor{lightgreen}{1.29} & \cellcolor{lightcoral}{-8.97} & \cellcolor{lightcoral}{-6.88} & \cellcolor{lightcoral}{-1.42} \\
& \checkmark & 0.00 & 0.00 & \cellcolor{lightcoral}{-0.48} & 0.00 & 0.00 & \cellcolor{lightcoral}{-3.91} & \cellcolor{lightcoral}{-0.83} & \cellcolor{lightcoral}{-2.00} & \cellcolor{lightgreen}{0.48} & 0.00 & \cellcolor{lightcoral}{-0.15} & 0.00 & \cellcolor{lightcoral}{-2.04} & \cellcolor{lightgreen}{2.92} & \cellcolor{lightcoral}{-2.71} & 0.00 & \cellcolor{lightcoral}{-7.44} & \cellcolor{lightcoral}{-4.18} & \cellcolor{lightgreen}{2.94} & \cellcolor{lightcoral}{-4.44} & \cellcolor{lightcoral}{-1.09} \\
\midrule
\multirow{2}{*}{\makecell[l]{Zero-Shot +\\Task Description}} & \xmark & 0.00 & 0.00 & \cellcolor{lightcoral}{-8.15} & 0.00 & 0.00 & \cellcolor{lightcoral}{-8.57} & \cellcolor{lightcoral}{-5.97} & \cellcolor{lightcoral}{-2.00} & \cellcolor{lightgreen}{5.15} & 0.00 & \cellcolor{lightgreen}{2.10} & 0.00 & \cellcolor{lightgreen}{2.34} & \cellcolor{lightgreen}{1.76} & \cellcolor{lightcoral}{-11.01} & 0.00 & \cellcolor{lightcoral}{-11.74} & \cellcolor{lightcoral}{-6.68} & \cellcolor{lightgreen}{4.51} & \cellcolor{lightcoral}{-9.84} & \cellcolor{lightcoral}{-2.40} \\
& \checkmark & 0.00 & 0.00 & \cellcolor{lightcoral}{-2.89} & 0.00 & 0.00 & \cellcolor{lightcoral}{-6.57} & \cellcolor{lightcoral}{-4.21} & \cellcolor{lightcoral}{-1.96} & \cellcolor{lightcoral}{-4.00} & 0.00 & \cellcolor{lightgreen}{0.88} & 0.00 & \cellcolor{lightgreen}{4.20} & \cellcolor{lightcoral}{-1.75} & \cellcolor{lightcoral}{-15.15} & 0.00 & \cellcolor{lightcoral}{-7.18} & \cellcolor{lightcoral}{-0.92} & \cellcolor{lightgreen}{0.90} & \cellcolor{lightcoral}{-3.03} & \cellcolor{lightcoral}{-2.08} \\
\midrule
\multirow{2}{*}{\makecell[l]{Few-Shot +\\Task Description}} & \xmark & 0.00 & 0.00 & \cellcolor{lightgreen}{5.06} & 0.00 & 0.00 & \cellcolor{lightgreen}{4.77} & \cellcolor{lightgreen}{4.35} & \cellcolor{lightgreen}{3.01} & \cellcolor{lightcoral}{-11.66} & 0.00 & \cellcolor{lightgreen}{8.19} & 0.00 & \cellcolor{lightcoral}{-4.01} & \cellcolor{lightgreen}{2.38} & \cellcolor{lightcoral}{-6.75} & 0.00 & \cellcolor{lightgreen}{0.06} & \cellcolor{lightgreen}{2.57} & \cellcolor{lightgreen}{1.82} & \cellcolor{lightgreen}{9.74} & \cellcolor{lightgreen}{0.98} \\
& \checkmark & 0.00 & 0.00 & \cellcolor{lightgreen}{0.91} & 0.00 & 0.00 & \cellcolor{lightgreen}{7.12} & \cellcolor{lightcoral}{-2.78} & \cellcolor{lightgreen}{2.17} & \cellcolor{lightgreen}{0.44} & 0.00 & \cellcolor{lightcoral}{-1.77} & 0.00 & \cellcolor{lightgreen}{4.19} & \cellcolor{lightgreen}{2.14} & \cellcolor{lightcoral}{-15.56} & 0.00 & \cellcolor{lightcoral}{-1.04} & \cellcolor{lightgreen}{10.75} & \cellcolor{lightgreen}{1.38} & \cellcolor{lightcoral}{-3.70} & \cellcolor{lightgreen}{0.21} \\
\bottomrule
\end{tabular}}
\setlength{\fboxsep}{1pt}
\caption{Impact of the thinking mode on performance in cross-lingual settings across three prompting techniques for the \texttt{Qwen3 8B}. The table shows the difference in the Macro F1 score when using claim-language instructions (the second language in the column name) versus English. Each row compares performance with (\checkmark) and without (\xmark) "thinking mode", across 20 language pairs. Positive values (\colorbox{lightgreen}{green}) indicate improved performance with claim-language instructions. Average scores are reported in the final column.}
\label{tab:thinking-claim-language}
\end{table*}

\begin{table*}
\resizebox{\textwidth}{!}{%
\begin{tabular}{lllllllllllllllllllllll}
\toprule
\textbf{Model} & \textbf{Technique} & \textbf{ara} & \textbf{bul} & \textbf{ces} & \textbf{deu} & \textbf{ell} & \textbf{eng} & \textbf{fra} & \textbf{hbs} & \textbf{hin} & \textbf{hun} & \textbf{kor} & \textbf{msa} & \textbf{mya} & \textbf{nld} & \textbf{pol} & \textbf{por} & \textbf{ron} & \textbf{slk} & \textbf{spa} & \textbf{tha} & \textbf{Avg.} \\
\midrule
\multirow{4}{*}{\texttt{Qwen3 8B}} & ZS & 0.79 & 0.85 & 0.68 & 0.58 & 0.84 & 0.76 & 0.80 & 0.61 & 0.76 & 0.84 & 0.90 & 0.66 & 0.78 & 0.74 & 0.61 & 0.64 & 0.90 & 0.78 & 0.71 & 0.86 & 0.75 \\
 & ZS + Task & 0.91 & 0.95 & 0.73 & 0.77 & 0.86 & 0.72 & 0.83 & 0.65 & 0.82 & 0.87 & 0.86 & 0.67 & 0.94 & 0.81 & 0.68 & 0.77 & 0.84 & 0.84 & 0.80 & 0.89 & 0.81 \\
 & FS + Task & 0.86 & 0.91 & 0.71 & 0.62 & 0.81 & 0.57 & 0.87 & 0.68 & 0.81 & 0.75 & 0.85 & 0.59 & 0.92 & 0.68 & 0.62 & 0.67 & 0.82 & 0.73 & 0.75 & 0.79 & 0.75 \\
 & CoT & 0.97 & 0.99 & 0.93 & 0.92 & 0.97 & 0.88 & 0.95 & 0.87 & 0.96 & 0.95 & 0.97 & 0.91 & 0.98 & 0.90 & 0.89 & 0.88 & 0.95 & 0.94 & 0.92 & 0.97 & 0.94 \\
 \midrule
\multicolumn{2}{l}{\textit{Average}} & 0.88 & 0.93 & 0.76 & 0.72 & 0.87 & 0.73 & 0.86 & 0.70 & 0.84 & 0.85 & 0.90 & 0.71 & 0.91 & 0.78 & 0.70 & 0.74 & 0.88 & 0.82 & 0.80 & 0.88 & 0.81 \\
\midrule
\multirow{4}{*}{\texttt{Qwen3 14B}} & ZS & 0.78 & 0.94 & 0.63 & 0.68 & 0.89 & 0.78 & 0.80 & 0.62 & 0.81 & 0.77 & 0.91 & 0.79 & 0.85 & 0.73 & 0.65 & 0.67 & 0.87 & 0.83 & 0.80 & 0.95 & 0.79 \\
 & ZS + Task & 0.88 & 0.93 & 0.68 & 0.75 & 0.85 & 0.70 & 0.84 & 0.64 & 0.81 & 0.76 & 0.84 & 0.72 & 0.93 & 0.78 & 0.65 & 0.72 & 0.82 & 0.75 & 0.79 & 0.91 & 0.79 \\
 & FS + Task & 0.88 & 0.92 & 0.69 & 0.71 & 0.82 & 0.70 & 0.84 & 0.66 & 0.84 & 0.77 & 0.80 & 0.74 & 0.93 & 0.80 & 0.67 & 0.72 & 0.83 & 0.68 & 0.78 & 0.85 & 0.78 \\
 & CoT & 0.95 & 0.99 & 0.90 & 0.92 & 0.96 & 0.87 & 0.94 & 0.85 & 0.96 & 0.92 & 0.96 & 0.92 & 0.99 & 0.92 & 0.85 & 0.92 & 0.95 & 0.94 & 0.95 & 0.95 & 0.93 \\
 \midrule
\multicolumn{2}{l}{\textit{Average}} & 0.87 & 0.95 & 0.73 & 0.77 & 0.88 & 0.76 & 0.86 & 0.69 & 0.86 & 0.81 & 0.88 & 0.79 & 0.93 & 0.81 & 0.71 & 0.76 & 0.87 & 0.80 & 0.83 & 0.92 & 0.82 \\
\midrule
\multirow{4}{*}{\texttt{Gemma3 12B}} & ZS & 0.66 & 0.60 & 0.28 & 0.28 & 0.59 & 0.45 & 0.65 & 0.26 & 0.59 & 0.48 & 0.65 & 0.55 & 0.65 & 0.38 & 0.29 & 0.36 & 0.44 & 0.45 & 0.46 & 0.78 & 0.49 \\
 & ZS + Task & 0.67 & 0.69 & 0.43 & 0.33 & 0.58 & 0.39 & 0.61 & 0.32 & 0.62 & 0.51 & 0.61 & 0.48 & 0.77 & 0.41 & 0.36 & 0.42 & 0.53 & 0.47 & 0.49 & 0.74 & 0.52 \\
 & FS + Task & 0.96 & 0.96 & 0.78 & 0.70 & 0.80 & 0.72 & 0.90 & 0.73 & 0.93 & 0.82 & 0.88 & 0.93 & 0.98 & 0.75 & 0.67 & 0.80 & 0.82 & 0.73 & 0.77 & 0.94 & 0.83 \\
 & CoT & 0.70 & 0.66 & 0.42 & 0.35 & 0.61 & 0.46 & 0.57 & 0.33 & 0.65 & 0.54 & 0.63 & 0.47 & 0.74 & 0.43 & 0.36 & 0.42 & 0.52 & 0.45 & 0.50 & 0.67 & 0.52 \\
 \midrule
\multicolumn{2}{l}{\textit{Average}} & 0.75 & 0.73 & 0.48 & 0.42 & 0.65 & 0.51 & 0.68 & 0.41 & 0.70 & 0.59 & 0.69 & 0.61 & 0.79 & 0.49 & 0.42 & 0.50 & 0.58 & 0.53 & 0.56 & 0.78 & 0.59 \\
\midrule
\multirow{4}{*}{\texttt{Gemma3 27B}} & ZS & 0.65 & 0.66 & 0.40 & 0.35 & 0.65 & 0.47 & 0.62 & 0.26 & 0.65 & 0.54 & 0.69 & 0.55 & 0.71 & 0.39 & 0.31 & 0.43 & 0.47 & 0.51 & 0.50 & 0.81 & 0.53 \\
 & ZS + Task & 0.62 & 0.65 & 0.40 & 0.27 & 0.56 & 0.35 & 0.57 & 0.23 & 0.58 & 0.47 & 0.57 & 0.41 & 0.74 & 0.36 & 0.30 & 0.33 & 0.46 & 0.47 & 0.47 & 0.67 & 0.47 \\
 & FS + Task & 0.92 & 0.92 & 0.69 & 0.58 & 0.84 & 0.63 & 0.77 & 0.54 & 0.83 & 0.73 & 0.84 & 0.83 & 0.98 & 0.69 & 0.64 & 0.71 & 0.79 & 0.68 & 0.73 & 0.93 & 0.76 \\
 & CoT & 0.64 & 0.53 & 0.35 & 0.20 & 0.51 & 0.33 & 0.50 & 0.27 & 0.56 & 0.42 & 0.53 & 0.38 & 0.67 & 0.29 & 0.28 & 0.41 & 0.34 & 0.36 & 0.42 & 0.56 & 0.43 \\
 \midrule
\multicolumn{2}{l}{\textit{Average}} & 0.71 & 0.69 & 0.46 & 0.35 & 0.64 & 0.45 & 0.62 & 0.33 & 0.66 & 0.54 & 0.66 & 0.54 & 0.78 & 0.43 & 0.38 & 0.47 & 0.52 & 0.51 & 0.53 & 0.74 & 0.55 \\
\midrule
\multirow{4}{*}{\texttt{Llama3.1 8B}} & ZS & 0.50 & 0.50 & 0.28 & 0.20 & 0.47 & 0.51 & 0.54 & 0.24 & 0.63 & 0.38 & 0.52 & 0.35 & 0.44 & 0.29 & 0.30 & 0.36 & 0.50 & 0.32 & 0.40 & 0.66 & 0.42 \\
 & ZS + Task & 0.82 & 0.84 & 0.62 & 0.62 & 0.84 & 0.75 & 0.84 & 0.58 & 0.84 & 0.72 & 0.77 & 0.70 & 0.88 & 0.75 & 0.52 & 0.63 & 0.83 & 0.69 & 0.74 & 0.83 & 0.74 \\
 & FS + Task & 0.89 & 0.94 & 0.87 & 0.87 & 0.89 & 0.83 & 0.88 & 0.84 & 0.87 & 0.90 & 0.95 & 0.91 & 0.96 & 0.87 & 0.90 & 0.88 & 0.94 & 0.87 & 0.84 & 0.94 & 0.89 \\
 & CoT & 0.76 & 0.86 & 0.65 & 0.61 & 0.74 & 0.74 & 0.73 & 0.56 & 0.77 & 0.75 & 0.74 & 0.68 & 0.84 & 0.67 & 0.55 & 0.57 & 0.74 & 0.75 & 0.62 & 0.84 & 0.71 \\
 \midrule
\multicolumn{2}{l}{\textit{Average}} & 0.74 & 0.79 & 0.61 & 0.58 & 0.74 & 0.71 & 0.75 & 0.56 & 0.78 & 0.69 & 0.75 & 0.66 & 0.78 & 0.65 & 0.57 & 0.61 & 0.75 & 0.66 & 0.65 & 0.82 & 0.69 \\
\midrule
\multirow{4}{*}{\texttt{Llama3.1 70B}} & ZS & 0.82 & 0.92 & 0.69 & 0.69 & 0.84 & 0.77 & 0.82 & 0.64 & 0.92 & 0.78 & 0.86 & 0.82 & 0.78 & 0.71 & 0.63 & 0.68 & 0.81 & 0.80 & 0.78 & 0.91 & 0.78 \\
 & ZS + Task & 0.81 & 0.94 & 0.66 & 0.64 & 0.82 & 0.69 & 0.78 & 0.59 & 0.87 & 0.74 & 0.83 & 0.73 & 0.90 & 0.68 & 0.60 & 0.67 & 0.81 & 0.74 & 0.76 & 0.92 & 0.76 \\
 & FS + Task & 0.95 & 0.96 & 0.94 & 0.95 & 0.95 & 0.93 & 0.96 & 0.96 & 0.95 & 0.96 & 0.99 & 0.98 & 0.98 & 0.95 & 0.97 & 0.95 & 0.97 & 0.92 & 0.96 & 1.00 & 0.96 \\
 & CoT & 0.88 & 0.48 & 0.82 & 0.82 & 0.86 & 0.78 & 0.84 & 0.73 & 0.82 & 0.81 & 0.93 & 0.80 & 0.80 & 0.82 & 0.77 & 0.75 & 0.92 & 0.87 & 0.75 & 0.95 & 0.81 \\
 \midrule
\multicolumn{2}{l}{\textit{Average}} & 0.87 & 0.83 & 0.78 & 0.78 & 0.87 & 0.79 & 0.85 & 0.73 & 0.89 & 0.82 & 0.90 & 0.83 & 0.87 & 0.79 & 0.74 & 0.76 & 0.88 & 0.83 & 0.81 & 0.95 & 0.83 \\
\bottomrule
\end{tabular}}
\caption{The capabilities of LLMs in filtering irrelevant pairs based on TNR (higher is better) using English instruction in monolingual settings. The average TNR is calculated across all languages for each model and technique.}
\label{tab:tnr-monolingual}
\end{table*}

\begin{table*}
\resizebox{\textwidth}{!}{%
\begin{tabular}{lllllllllllllllllllllll}
\toprule
\textbf{Model} & \textbf{Technique} & \textbf{ara} & \textbf{bul} & \textbf{ces} & \textbf{deu} & \textbf{ell} & \textbf{eng} & \textbf{fra} & \textbf{hbs} & \textbf{hin} & \textbf{hun} & \textbf{kor} & \textbf{msa} & \textbf{mya} & \textbf{nld} & \textbf{pol} & \textbf{por} & \textbf{ron} & \textbf{slk} & \textbf{spa} & \textbf{tha} & \textbf{Avg.} \\
\midrule
\multirow{4}{*}{\texttt{Qwen3 8B}} & ZS & \cellcolor{lightcoral}{-0.01} & \cellcolor{lightcoral}{-0.10} & \cellcolor{lightgreen}{0.01} & \cellcolor{lightgreen}{0.17} & \cellcolor{lightcoral}{-0.10} & 0.00 & \cellcolor{lightgreen}{0.17} & \cellcolor{lightgreen}{0.11} & \cellcolor{lightgreen}{0.03} & \cellcolor{lightcoral}{-0.05} & \cellcolor{lightcoral}{-0.29} & \cellcolor{lightcoral}{-0.13} & \cellcolor{lightcoral}{-0.32} & \cellcolor{lightcoral}{-0.11} & \cellcolor{lightgreen}{0.02} & \cellcolor{lightgreen}{0.12} & \cellcolor{lightgreen}{0.04} & \cellcolor{lightgreen}{0.07} & \cellcolor{lightcoral}{-0.08} & \cellcolor{lightcoral}{-0.35} & \cellcolor{lightcoral}{-0.04} \\
& ZS + Task & \cellcolor{lightcoral}{-0.02} & 0.00 & \cellcolor{lightgreen}{0.05} & \cellcolor{lightgreen}{0.14} & \cellcolor{lightcoral}{-0.04} & 0.00 & \cellcolor{lightgreen}{0.06} & \cellcolor{lightgreen}{0.03} & \cellcolor{lightcoral}{-0.05} & \cellcolor{lightcoral}{-0.06} & \cellcolor{lightcoral}{-0.09} & \cellcolor{lightcoral}{-0.04} & \cellcolor{lightcoral}{-0.18} & \cellcolor{lightgreen}{0.10} & \cellcolor{lightgreen}{0.09} & \cellcolor{lightcoral}{-0.03} & \cellcolor{lightgreen}{0.02} & \cellcolor{lightcoral}{-0.01} & \cellcolor{lightgreen}{0.09} & \cellcolor{lightcoral}{-0.08} & 0.00 \\
& FS + Task & \cellcolor{lightgreen}{0.03} & \cellcolor{lightgreen}{0.05} & \cellcolor{lightcoral}{-0.08} & \cellcolor{lightgreen}{0.19} & \cellcolor{lightcoral}{-0.21} & 0.00 & \cellcolor{lightgreen}{0.08} & \cellcolor{lightgreen}{0.09} & \cellcolor{lightgreen}{0.11} & \cellcolor{lightgreen}{0.09} & \cellcolor{lightcoral}{-0.10} & \cellcolor{lightgreen}{0.30} & \cellcolor{lightcoral}{-0.36} & \cellcolor{lightgreen}{0.22} & \cellcolor{lightgreen}{0.01} & \cellcolor{lightgreen}{0.11} & \cellcolor{lightgreen}{0.11} & \cellcolor{lightcoral}{-0.02} & \cellcolor{lightgreen}{0.16} & \cellcolor{lightgreen}{0.15} & \cellcolor{lightgreen}{0.05} \\
& CoT & \cellcolor{lightcoral}{-0.01} & \cellcolor{lightcoral}{-0.02} & \cellcolor{lightcoral}{-0.08} & \cellcolor{lightgreen}{0.04} & \cellcolor{lightcoral}{-0.03} & 0.00 & \cellcolor{lightcoral}{-0.04} & \cellcolor{lightcoral}{-0.10} & \cellcolor{lightcoral}{-0.11} & \cellcolor{lightcoral}{-0.05} & \cellcolor{lightcoral}{-0.16} & \cellcolor{lightcoral}{-0.12} & \cellcolor{lightcoral}{-0.44} & \cellcolor{lightgreen}{0.01} & \cellcolor{lightgreen}{0.03} & \cellcolor{lightcoral}{-0.01} & \cellcolor{lightcoral}{-0.11} & 0.00 & \cellcolor{lightgreen}{0.02} & \cellcolor{lightcoral}{-0.03} & \cellcolor{lightcoral}{-0.06} \\
\midrule
\multicolumn{2}{l}{Average} & 0.00 & \cellcolor{lightcoral}{-0.02} & \cellcolor{lightcoral}{-0.03} & \cellcolor{lightgreen}{0.13} & \cellcolor{lightcoral}{-0.09} & 0.00 & \cellcolor{lightgreen}{0.07} & \cellcolor{lightgreen}{0.03} & \cellcolor{lightcoral}{-0.01} & \cellcolor{lightcoral}{-0.02} & \cellcolor{lightcoral}{-0.16} & 0.00 & \cellcolor{lightcoral}{-0.33} & \cellcolor{lightgreen}{0.05} & \cellcolor{lightgreen}{0.04} & \cellcolor{lightgreen}{0.04} & \cellcolor{lightgreen}{0.01} & \cellcolor{lightgreen}{0.01} & \cellcolor{lightgreen}{0.05} & \cellcolor{lightcoral}{-0.08} & \cellcolor{lightcoral}{-0.01} \\
\midrule
\multirow{4}{*}{\texttt{Qwen3 14B}} & ZS & \cellcolor{lightgreen}{0.05} & 0.00 & \cellcolor{lightgreen}{0.03} & 0.00 & 0.00 & 0.00 & \cellcolor{lightgreen}{0.04} & \cellcolor{lightgreen}{0.07} & \cellcolor{lightcoral}{-0.31} & \cellcolor{lightcoral}{-0.05} & \cellcolor{lightcoral}{-0.34} & \cellcolor{lightcoral}{-0.17} & \cellcolor{lightcoral}{-0.61} & \cellcolor{lightcoral}{-0.04} & \cellcolor{lightcoral}{-0.28} & \cellcolor{lightcoral}{-0.09} & \cellcolor{lightcoral}{-0.10} & \cellcolor{lightcoral}{-0.04} & \cellcolor{lightcoral}{-0.19} & \cellcolor{lightcoral}{-0.40} & \cellcolor{lightcoral}{-0.12} \\
& ZS + Task & \cellcolor{lightgreen}{0.07} & \cellcolor{lightgreen}{0.05} & \cellcolor{lightgreen}{0.18} & \cellcolor{lightgreen}{0.05} & \cellcolor{lightgreen}{0.10} & 0.00 & \cellcolor{lightgreen}{0.03} & \cellcolor{lightgreen}{0.07} & \cellcolor{lightgreen}{0.11} & \cellcolor{lightgreen}{0.07} & \cellcolor{lightcoral}{-0.03} & \cellcolor{lightgreen}{0.06} & \cellcolor{lightcoral}{-0.09} & \cellcolor{lightgreen}{0.04} & \cellcolor{lightgreen}{0.11} & \cellcolor{lightgreen}{0.05} & \cellcolor{lightgreen}{0.06} & \cellcolor{lightgreen}{0.10} & \cellcolor{lightgreen}{0.07} & \cellcolor{lightcoral}{-0.03} & \cellcolor{lightgreen}{0.05} \\
& FS + Task & 0.00 & \cellcolor{lightgreen}{0.05} & \cellcolor{lightcoral}{-0.01} & \cellcolor{lightcoral}{-0.03} & \cellcolor{lightgreen}{0.09} & 0.00 & \cellcolor{lightcoral}{-0.03} & \cellcolor{lightgreen}{0.05} & \cellcolor{lightgreen}{0.01} & \cellcolor{lightcoral}{-0.03} & \cellcolor{lightgreen}{0.03} & \cellcolor{lightcoral}{-0.02} & \cellcolor{lightcoral}{-0.19} & \cellcolor{lightcoral}{-0.02} & \cellcolor{lightgreen}{0.05} & \cellcolor{lightcoral}{-0.07} & \cellcolor{lightcoral}{-0.01} & \cellcolor{lightgreen}{0.07} & \cellcolor{lightgreen}{0.04} & \cellcolor{lightgreen}{0.02} & 0.00 \\
& CoT & \cellcolor{lightcoral}{-0.02} & \cellcolor{lightcoral}{-0.02} & \cellcolor{lightgreen}{0.01} & 0.00 & \cellcolor{lightcoral}{-0.01} & 0.00 & \cellcolor{lightcoral}{-0.04} & \cellcolor{lightcoral}{-0.02} & \cellcolor{lightcoral}{-0.03} & \cellcolor{lightgreen}{0.01} & \cellcolor{lightcoral}{-0.09} & \cellcolor{lightcoral}{-0.45} & \cellcolor{lightcoral}{-0.29} & \cellcolor{lightcoral}{-0.01} & \cellcolor{lightcoral}{-0.03} & \cellcolor{lightcoral}{-0.10} & \cellcolor{lightcoral}{-0.01} & 0.00 & \cellcolor{lightcoral}{-0.03} & \cellcolor{lightcoral}{-0.01} & \cellcolor{lightcoral}{-0.06} \\
\midrule
\multicolumn{2}{l}{Average} & \cellcolor{lightgreen}{0.03} & \cellcolor{lightgreen}{0.02} & \cellcolor{lightgreen}{0.05} & \cellcolor{lightgreen}{0.01} & \cellcolor{lightgreen}{0.04} & 0.00 & 0.00 & \cellcolor{lightgreen}{0.04} & \cellcolor{lightcoral}{-0.05} & 0.00 & \cellcolor{lightcoral}{-0.11} & \cellcolor{lightcoral}{-0.14} & \cellcolor{lightcoral}{-0.30} & \cellcolor{lightcoral}{-0.01} & \cellcolor{lightcoral}{-0.04} & \cellcolor{lightcoral}{-0.05} & \cellcolor{lightcoral}{-0.01} & \cellcolor{lightgreen}{0.03} & \cellcolor{lightcoral}{-0.03} & \cellcolor{lightcoral}{-0.11} & \cellcolor{lightcoral}{-0.03} \\

\midrule
\multirow{4}{*}{\texttt{Gemma3 12B}} & ZS & \cellcolor{lightgreen}{0.03} & \cellcolor{lightgreen}{0.16} & \cellcolor{lightgreen}{0.26} & 0.00 & \cellcolor{lightcoral}{-0.09} & 0.00 & \cellcolor{lightgreen}{0.06} & \cellcolor{lightgreen}{0.10} & \cellcolor{lightcoral}{-0.26} & \cellcolor{lightgreen}{0.02} & \cellcolor{lightcoral}{-0.36} & \cellcolor{lightcoral}{-0.24} & \cellcolor{lightcoral}{-0.47} & \cellcolor{lightcoral}{-0.03} & \cellcolor{lightgreen}{0.05} & \cellcolor{lightgreen}{0.07} & \cellcolor{lightgreen}{0.08} & \cellcolor{lightgreen}{0.17} & \cellcolor{lightgreen}{0.12} & \cellcolor{lightcoral}{-0.47} & \cellcolor{lightcoral}{-0.04} \\
& ZS + Task & \cellcolor{lightgreen}{0.22} & \cellcolor{lightgreen}{0.10} & \cellcolor{lightgreen}{0.04} & \cellcolor{lightgreen}{0.04} & \cellcolor{lightgreen}{0.16} & 0.00 & \cellcolor{lightgreen}{0.07} & \cellcolor{lightgreen}{0.06} & \cellcolor{lightgreen}{0.13} & \cellcolor{lightgreen}{0.08} & \cellcolor{lightgreen}{0.07} & \cellcolor{lightgreen}{0.12} & \cellcolor{lightcoral}{-0.12} & \cellcolor{lightgreen}{0.26} & \cellcolor{lightgreen}{0.22} & \cellcolor{lightgreen}{0.10} & \cellcolor{lightgreen}{0.01} & \cellcolor{lightgreen}{0.19} & \cellcolor{lightgreen}{0.05} & \cellcolor{lightgreen}{0.06} & \cellcolor{lightgreen}{0.09} \\
& FS + Task & \cellcolor{lightcoral}{-0.04} & \cellcolor{lightcoral}{-0.04} & \cellcolor{lightcoral}{-0.15} & \cellcolor{lightcoral}{-0.14} & \cellcolor{lightcoral}{-0.03} & 0.00 & \cellcolor{lightcoral}{-0.07} & \cellcolor{lightcoral}{-0.04} & \cellcolor{lightcoral}{-0.01} & \cellcolor{lightcoral}{-0.08} & \cellcolor{lightcoral}{-0.08} & \cellcolor{lightcoral}{-0.06} & \cellcolor{lightcoral}{-0.20} & \cellcolor{lightgreen}{0.05} & \cellcolor{lightcoral}{-0.03} & \cellcolor{lightcoral}{-0.14} & \cellcolor{lightcoral}{-0.06} & \cellcolor{lightcoral}{-0.08} & 0.00 & \cellcolor{lightcoral}{-0.04} & \cellcolor{lightcoral}{-0.06} \\
& CoT & \cellcolor{lightgreen}{0.12} & \cellcolor{lightcoral}{-0.01} & \cellcolor{lightgreen}{0.07} & \cellcolor{lightgreen}{0.23} & \cellcolor{lightcoral}{-0.09} & 0.00 & \cellcolor{lightcoral}{-0.01} & \cellcolor{lightgreen}{0.13} & \cellcolor{lightgreen}{0.06} & \cellcolor{lightgreen}{0.12} & \cellcolor{lightcoral}{-0.12} & \cellcolor{lightgreen}{0.05} & \cellcolor{lightcoral}{-0.35} & \cellcolor{lightgreen}{0.22} & \cellcolor{lightgreen}{0.07} & \cellcolor{lightcoral}{-0.04} & 0.00 & \cellcolor{lightgreen}{0.08} & \cellcolor{lightgreen}{0.04} & \cellcolor{lightgreen}{0.07} & \cellcolor{lightgreen}{0.03} \\
\midrule
\multicolumn{2}{l}{Average} & \cellcolor{lightgreen}{0.08} & \cellcolor{lightgreen}{0.05} & \cellcolor{lightgreen}{0.05} & \cellcolor{lightgreen}{0.04} & \cellcolor{lightcoral}{-0.01} & 0.00 & \cellcolor{lightgreen}{0.01} & \cellcolor{lightgreen}{0.06} & \cellcolor{lightcoral}{-0.02} & \cellcolor{lightgreen}{0.03} & \cellcolor{lightcoral}{-0.12} & \cellcolor{lightcoral}{-0.03} & \cellcolor{lightcoral}{-0.29} & \cellcolor{lightgreen}{0.13} & \cellcolor{lightgreen}{0.08} & 0.00 & \cellcolor{lightgreen}{0.01} & \cellcolor{lightgreen}{0.09} & \cellcolor{lightgreen}{0.05} & \cellcolor{lightcoral}{-0.09} & \cellcolor{lightgreen}{0.01} \\
\midrule
\multirow{4}{*}{\texttt{Gemma3 27B}} & ZS & \cellcolor{lightgreen}{0.06} & \cellcolor{lightgreen}{0.17} & \cellcolor{lightgreen}{0.19} & \cellcolor{lightgreen}{0.06} & \cellcolor{lightgreen}{0.06} & 0.00 & \cellcolor{lightgreen}{0.09} & \cellcolor{lightgreen}{0.54} & \cellcolor{lightcoral}{-0.11} & \cellcolor{lightgreen}{0.01} & \cellcolor{lightcoral}{-0.06} & \cellcolor{lightgreen}{0.05} & \cellcolor{lightcoral}{-0.23} & \cellcolor{lightgreen}{0.05} & \cellcolor{lightgreen}{0.01} & \cellcolor{lightgreen}{0.11} & \cellcolor{lightgreen}{0.27} & \cellcolor{lightgreen}{0.13} & \cellcolor{lightgreen}{0.18} & \cellcolor{lightcoral}{-0.13} & \cellcolor{lightgreen}{0.07} \\
& ZS + Task & \cellcolor{lightgreen}{0.17} & \cellcolor{lightgreen}{0.19} & \cellcolor{lightgreen}{0.10} & \cellcolor{lightgreen}{0.28} & \cellcolor{lightgreen}{0.11} & 0.00 & \cellcolor{lightgreen}{0.11} & \cellcolor{lightgreen}{0.15} & \cellcolor{lightgreen}{0.22} & \cellcolor{lightgreen}{0.10} & \cellcolor{lightgreen}{0.06} & \cellcolor{lightgreen}{0.25} & \cellcolor{lightgreen}{0.15} & \cellcolor{lightgreen}{0.19} & \cellcolor{lightgreen}{0.21} & \cellcolor{lightgreen}{0.16} & \cellcolor{lightgreen}{0.20} & \cellcolor{lightgreen}{0.06} & \cellcolor{lightgreen}{0.19} & \cellcolor{lightgreen}{0.16} & \cellcolor{lightgreen}{0.15} \\
 & FS + Task & \cellcolor{lightcoral}{-0.01} & \cellcolor{lightcoral}{-0.04} & \cellcolor{lightcoral}{-0.04} & \cellcolor{lightgreen}{0.08} & 0.00 & 0.00 & \cellcolor{lightcoral}{-0.02} & \cellcolor{lightcoral}{-0.01} & \cellcolor{lightgreen}{0.01} & \cellcolor{lightcoral}{-0.06} & \cellcolor{lightcoral}{-0.04} & 0.00 & \cellcolor{lightcoral}{-0.02} & \cellcolor{lightgreen}{0.06} & \cellcolor{lightgreen}{0.04} & \cellcolor{lightcoral}{-0.01} & \cellcolor{lightcoral}{-0.01} & \cellcolor{lightcoral}{-0.04} & \cellcolor{lightcoral}{-0.03} & \cellcolor{lightcoral}{-0.01} & \cellcolor{lightcoral}{-0.01} \\
& CoT & \cellcolor{lightgreen}{0.15} & \cellcolor{lightgreen}{0.06} & \cellcolor{lightgreen}{0.06} & \cellcolor{lightgreen}{0.39} & \cellcolor{lightcoral}{-0.04} & 0.00 & \cellcolor{lightcoral}{-0.03} & \cellcolor{lightcoral}{-0.04} & \cellcolor{lightgreen}{0.15} & \cellcolor{lightgreen}{0.14} & \cellcolor{lightcoral}{-0.01} & \cellcolor{lightgreen}{0.15} & 0.00 & \cellcolor{lightgreen}{0.06} & \cellcolor{lightgreen}{0.08} & \cellcolor{lightgreen}{0.08} & \cellcolor{lightgreen}{0.12} & \cellcolor{lightgreen}{0.09} & \cellcolor{lightgreen}{0.09} & \cellcolor{lightgreen}{0.21} & \cellcolor{lightgreen}{0.09} \\
\midrule
\multicolumn{2}{l}{Average} & \cellcolor{lightgreen}{0.09} & \cellcolor{lightgreen}{0.10} & \cellcolor{lightgreen}{0.08} & \cellcolor{lightgreen}{0.20} & \cellcolor{lightgreen}{0.03} & 0.00 & \cellcolor{lightgreen}{0.04} & \cellcolor{lightgreen}{0.16} & \cellcolor{lightgreen}{0.07} & \cellcolor{lightgreen}{0.05} & \cellcolor{lightcoral}{-0.01} & \cellcolor{lightgreen}{0.11} & \cellcolor{lightcoral}{-0.03} & \cellcolor{lightgreen}{0.09} & \cellcolor{lightgreen}{0.09} & \cellcolor{lightgreen}{0.09} & \cellcolor{lightgreen}{0.14} & \cellcolor{lightgreen}{0.06} & \cellcolor{lightgreen}{0.11} & \cellcolor{lightgreen}{0.06} & \cellcolor{lightgreen}{0.08} \\
\midrule
\multirow{4}{*}{\texttt{Llama3.1 8B}} & ZS & \cellcolor{lightgreen}{0.30} & \cellcolor{lightcoral}{-0.16} & \cellcolor{lightgreen}{0.16} & \cellcolor{lightgreen}{0.17} & \cellcolor{lightgreen}{0.25} & 0.00 & \cellcolor{lightgreen}{0.10} & \cellcolor{lightgreen}{0.40} & \cellcolor{lightcoral}{-0.28} & \cellcolor{lightcoral}{-0.08} & \cellcolor{lightgreen}{0.10} & \cellcolor{lightgreen}{0.09} & \cellcolor{lightcoral}{-0.07} & \cellcolor{lightcoral}{-0.04} & \cellcolor{lightgreen}{0.16} & \cellcolor{lightcoral}{-0.10} & \cellcolor{lightgreen}{0.32} & \cellcolor{lightgreen}{0.40} & \cellcolor{lightgreen}{0.04} & \cellcolor{lightcoral}{-0.53} & \cellcolor{lightgreen}{0.06} \\
 & ZS + Task & \cellcolor{lightgreen}{0.12} & \cellcolor{lightcoral}{-0.06} & \cellcolor{lightgreen}{0.11} & \cellcolor{lightgreen}{0.33} & \cellcolor{lightgreen}{0.11} & 0.00 & \cellcolor{lightgreen}{0.08} & \cellcolor{lightcoral}{-0.04} & \cellcolor{lightcoral}{-0.28} & \cellcolor{lightgreen}{0.13} & \cellcolor{lightcoral}{-0.75} & \cellcolor{lightgreen}{0.02} & \cellcolor{lightcoral}{-0.12} & \cellcolor{lightgreen}{0.17} & \cellcolor{lightgreen}{0.20} & \cellcolor{lightgreen}{0.01} & \cellcolor{lightcoral}{-0.01} & \cellcolor{lightgreen}{0.25} & \cellcolor{lightgreen}{0.14} & \cellcolor{lightcoral}{-0.03} & \cellcolor{lightgreen}{0.02} \\
& FS + Task & \cellcolor{lightgreen}{0.10} & \cellcolor{lightcoral}{-0.06} & \cellcolor{lightcoral}{-0.19} & \cellcolor{lightgreen}{0.05} & \cellcolor{lightcoral}{-0.15} & 0.00 & \cellcolor{lightgreen}{0.03} & \cellcolor{lightcoral}{-0.07} & \cellcolor{lightgreen}{0.10} & 0.00 & \cellcolor{lightcoral}{-0.05} & \cellcolor{lightcoral}{-0.07} & \cellcolor{lightcoral}{-0.24} & \cellcolor{lightcoral}{-0.03} & \cellcolor{lightcoral}{-0.05} & \cellcolor{lightgreen}{0.03} & \cellcolor{lightcoral}{-0.09} & \cellcolor{lightgreen}{0.06} & \cellcolor{lightgreen}{0.09} & \cellcolor{lightcoral}{-0.07} & \cellcolor{lightcoral}{-0.03} \\
& CoT & \cellcolor{lightgreen}{0.13} & \cellcolor{lightgreen}{0.03} & \cellcolor{lightgreen}{0.10} & \cellcolor{lightgreen}{0.16} & 0.00 & 0.00 & \cellcolor{lightgreen}{0.03} & \cellcolor{lightgreen}{0.06} & \cellcolor{lightcoral}{-0.37} & \cellcolor{lightgreen}{0.08} & \cellcolor{lightcoral}{-0.02} & 0.00 & \cellcolor{lightcoral}{-0.05} & \cellcolor{lightgreen}{0.08} & \cellcolor{lightgreen}{0.20} & \cellcolor{lightgreen}{0.32} & \cellcolor{lightcoral}{-0.09} & \cellcolor{lightgreen}{0.11} & \cellcolor{lightgreen}{0.04} & \cellcolor{lightgreen}{0.07} & \cellcolor{lightgreen}{0.04} \\
\midrule
\multicolumn{2}{l}{Average} & \cellcolor{lightgreen}{0.16} & \cellcolor{lightcoral}{-0.06} & \cellcolor{lightgreen}{0.05} & \cellcolor{lightgreen}{0.18} & \cellcolor{lightgreen}{0.05} & 0.00 & \cellcolor{lightgreen}{0.06} & \cellcolor{lightgreen}{0.09} & \cellcolor{lightcoral}{-0.21} & \cellcolor{lightgreen}{0.04} & \cellcolor{lightcoral}{-0.18} & \cellcolor{lightgreen}{0.01} & \cellcolor{lightcoral}{-0.12} & \cellcolor{lightgreen}{0.05} & \cellcolor{lightgreen}{0.13} & \cellcolor{lightgreen}{0.06} & \cellcolor{lightgreen}{0.03} & \cellcolor{lightgreen}{0.20} & \cellcolor{lightgreen}{0.08} & \cellcolor{lightcoral}{-0.14} & \cellcolor{lightgreen}{0.02} \\
\midrule
\multirow{4}{*}{\texttt{Llama3.1 70B}} & ZS & \cellcolor{lightgreen}{0.13} & \cellcolor{lightgreen}{0.03} & \cellcolor{lightgreen}{0.14} & \cellcolor{lightgreen}{0.05} & \cellcolor{lightgreen}{0.04} & 0.00 & \cellcolor{lightgreen}{0.03} & \cellcolor{lightgreen}{0.30} & \cellcolor{lightgreen}{0.05} & \cellcolor{lightgreen}{0.06} & \cellcolor{lightcoral}{-0.08} & \cellcolor{lightcoral}{-0.10} & \cellcolor{lightgreen}{0.15} & \cellcolor{lightgreen}{0.06} & \cellcolor{lightgreen}{0.15} & \cellcolor{lightgreen}{0.04} & \cellcolor{lightgreen}{0.07} & \cellcolor{lightgreen}{0.13} & \cellcolor{lightgreen}{0.06} & \cellcolor{lightcoral}{-0.26} & \cellcolor{lightgreen}{0.05} \\
 & ZS + Task & \cellcolor{lightgreen}{0.18} & \cellcolor{lightgreen}{0.01} & \cellcolor{lightgreen}{0.13} & \cellcolor{lightgreen}{0.18} & \cellcolor{lightgreen}{0.15} & 0.00 & \cellcolor{lightgreen}{0.09} & \cellcolor{lightgreen}{0.25} & \cellcolor{lightgreen}{0.05} & \cellcolor{lightgreen}{0.17} & \cellcolor{lightgreen}{0.03} & \cellcolor{lightgreen}{0.14} & \cellcolor{lightgreen}{0.06} & \cellcolor{lightgreen}{0.15} & \cellcolor{lightgreen}{0.18} & \cellcolor{lightgreen}{0.04} & \cellcolor{lightgreen}{0.07} & \cellcolor{lightgreen}{0.20} & \cellcolor{lightgreen}{0.08} & \cellcolor{lightgreen}{0.07} & \cellcolor{lightgreen}{0.11} \\
 & FS + Task & \cellcolor{lightcoral}{-0.43} & \cellcolor{lightcoral}{-0.23} & \cellcolor{lightcoral}{-0.11} & \cellcolor{lightcoral}{-0.13} & \cellcolor{lightcoral}{-0.70} & 0.00 & \cellcolor{lightcoral}{-0.05} & \cellcolor{lightcoral}{-0.22} & \cellcolor{lightcoral}{-0.22} & \cellcolor{lightcoral}{-0.23} & \cellcolor{lightcoral}{-0.01} & \cellcolor{lightcoral}{-0.16} & \cellcolor{lightcoral}{-0.04} & \cellcolor{lightcoral}{-0.03} & \cellcolor{lightcoral}{-0.15} & \cellcolor{lightcoral}{-0.14} & \cellcolor{lightcoral}{-0.15} & \cellcolor{lightcoral}{-0.24} & \cellcolor{lightcoral}{-0.24} & \cellcolor{lightcoral}{-0.08} & \cellcolor{lightcoral}{-0.18} \\
 & CoT & \cellcolor{lightcoral}{-0.15} & \cellcolor{lightgreen}{0.42} & \cellcolor{lightcoral}{-0.02} & \cellcolor{lightgreen}{0.05} & \cellcolor{lightgreen}{0.03} & 0.00 & 0.00 & \cellcolor{lightgreen}{0.04} & \cellcolor{lightgreen}{0.03} & \cellcolor{lightgreen}{0.01} & \cellcolor{lightcoral}{-0.21} & 0.00 & \cellcolor{lightgreen}{0.19} & \cellcolor{lightgreen}{0.07} & \cellcolor{lightcoral}{-0.03} & \cellcolor{lightgreen}{0.01} & \cellcolor{lightcoral}{-0.10} & \cellcolor{lightgreen}{0.08} & \cellcolor{lightcoral}{-0.07} & \cellcolor{lightcoral}{-0.02} & \cellcolor{lightgreen}{0.02} \\
 \midrule
 \multicolumn{2}{l}{Average} & \cellcolor{lightcoral}{-0.07} & \cellcolor{lightgreen}{0.05} & \cellcolor{lightgreen}{0.03} & \cellcolor{lightgreen}{0.04} & \cellcolor{lightcoral}{-0.12} & 0.00 & \cellcolor{lightgreen}{0.02} & \cellcolor{lightgreen}{0.09} & \cellcolor{lightcoral}{-0.02} & 0.00 & \cellcolor{lightcoral}{-0.07} & \cellcolor{lightcoral}{-0.03} & \cellcolor{lightgreen}{0.09} & \cellcolor{lightgreen}{0.06} & \cellcolor{lightgreen}{0.04} & \cellcolor{lightcoral}{-0.01} & \cellcolor{lightcoral}{-0.02} & \cellcolor{lightgreen}{0.04} & \cellcolor{lightcoral}{-0.04} & \cellcolor{lightcoral}{-0.07} & 0.00 \\
\bottomrule
\end{tabular}}
\setlength{\fboxsep}{1pt}
\caption{Difference in TNR when using target-language instructions versus English in monolingual settings. Positive values (\colorbox{lightgreen}{green}) reflect improved filtering when using target-language instructions. The table presents results for six models and four prompting techniques, with average values computed for each model–prompting combination as well as for each language pair.}
\label{tab:tnr-monolingual-diff}
\end{table*}

\begin{table*}
\resizebox{\textwidth}{!}{%
\begin{tabular}{lllllllllllllllllllllll}
\toprule
\textbf{Model} & \textbf{Technique} & \textbf{spa-eng} & \textbf{hin-eng} & \textbf{eng-ara} & \textbf{fra-eng} & \textbf{deu-eng} & \textbf{eng-por} & \textbf{spa-por} & \textbf{deu-fra} & \textbf{slk-ces} & \textbf{slk-eng} & \textbf{pol-hbs} & \textbf{ces-eng} & \textbf{ces-pol} & \textbf{nld-deu} & \textbf{msa-ara} & \textbf{kor-eng} & \textbf{mya-msa} & \textbf{ara-fra} & \textbf{hun-pol} & \textbf{tha-por} & \textbf{Avg.} \\
\midrule
\multirow{4}{*}{\texttt{Qwen3 8B}} & ZS & 0.78 & 0.88 & 0.89 & 0.70 & 0.71 & 0.85 & 0.90 & 0.68 & 0.70 & 0.60 & 0.70 & 0.69 & 0.59 & 0.68 & 0.77 & 0.84 & 0.76 & 0.93 & 0.81 & 0.84 & 0.76 \\
& ZS + Task & 0.86 & 0.96 & 0.96 & 0.80 & 0.79 & 0.89 & 0.95 & 0.83 & 0.71 & 0.66 & 0.73 & 0.66 & 0.76 & 0.81 & 0.91 & 0.81 & 0.98 & 0.97 & 0.87 & 0.87 & 0.84 \\
& FS + Task & 0.73 & 0.92 & 0.97 & 0.71 & 0.67 & 0.78 & 0.91 & 0.77 & 0.76 & 0.58 & 0.65 & 0.60 & 0.80 & 0.70 & 0.88 & 0.71 & 1.00 & 0.98 & 0.80 & 0.75 & 0.78 \\
& CoT & 0.94 & 0.99 & 0.98 & 0.93 & 0.92 & 0.98 & 0.97 & 0.90 & 0.92 & 0.83 & 0.89 & 0.89 & 0.90 & 0.92 & 0.96 & 0.90 & 0.98 & 0.98 & 0.92 & 0.93 & 0.93 \\
\midrule
\multicolumn{2}{l}{\textit{Average}} & 0.83 & 0.94 & 0.95 & 0.79 & 0.77 & 0.88 & 0.93 & 0.80 & 0.77 & 0.67 & 0.74 & 0.71 & 0.76 & 0.78 & 0.88 & 0.82 & 0.93 & 0.97 & 0.85 & 0.85 & 0.83 \\
\midrule
\multirow{4}{*}{\texttt{Qwen3 14B}} & ZS & 0.86 & 0.91 & 0.91 & 0.84 & 0.79 & 0.87 & 0.92 & 0.81 & 0.64 & 0.61 & 0.63 & 0.69 & 0.65 & 0.72 & 0.85 & 0.82 & 0.89 & 0.96 & 0.81 & 0.83 & 0.80 \\
& ZS + Task & 0.86 & 0.94 & 0.94 & 0.80 & 0.71 & 0.88 & 0.97 & 0.81 & 0.65 & 0.58 & 0.67 & 0.64 & 0.70 & 0.83 & 0.94 & 0.77 & 0.98 & 0.97 & 0.80 & 0.80 & 0.81 \\
& FS + Task & 0.80 & 0.97 & 0.98 & 0.79 & 0.75 & 0.87 & 0.97 & 0.85 & 0.69 & 0.60 & 0.67 & 0.63 & 0.72 & 0.84 & 0.94 & 0.73 & 0.96 & 0.99 & 0.83 & 0.80 & 0.82 \\
& CoT & 0.96 & 1.00 & 0.98 & 0.94 & 0.93 & 0.96 & 0.99 & 0.93 & 0.90 & 0.87 & 0.90 & 0.88 & 0.90 & 0.92 & 0.97 & 0.90 & 0.99 & 0.99 & 0.91 & 0.91 & 0.94 \\
\midrule
\multicolumn{2}{l}{\textit{Average}} & 0.87 & 0.96 & 0.95 & 0.84 & 0.80 & 0.90 & 0.96 & 0.85 & 0.72 & 0.67 & 0.72 & 0.71 & 0.74 & 0.83 & 0.93 & 0.81 & 0.96 & 0.98 & 0.84 & 0.84 & 0.84 \\
\midrule
\multirow{4}{*}{\texttt{Gemma3 12B}} & ZS & 0.67 & 0.50 & 0.83 & 0.58 & 0.45 & 0.68 & 0.79 & 0.41 & 0.35 & 0.34 & 0.34 & 0.38 & 0.53 & 0.42 & 0.75 & 0.50 & 0.78 & 0.92 & 0.62 & 0.76 & 0.58 \\
& ZS + Task & 0.58 & 0.56 & 0.82 & 0.49 & 0.45 & 0.65 & 0.77 & 0.44 & 0.39 & 0.35 & 0.37 & 0.34 & 0.52 & 0.40 & 0.75 & 0.47 & 0.85 & 0.89 & 0.62 & 0.70 & 0.57 \\
& FS + Task & 0.84 & 0.98 & 0.98 & 0.84 & 0.82 & 0.91 & 0.92 & 0.85 & 0.81 & 0.64 & 0.81 & 0.74 & 0.79 & 0.84 & 0.98 & 0.86 & 1.00 & 0.99 & 0.88 & 0.90 & 0.87 \\
& CoT & 0.66 & 0.62 & 0.80 & 0.56 & 0.49 & 0.66 & 0.75 & 0.39 & 0.34 & 0.30 & 0.34 & 0.37 & 0.48 & 0.30 & 0.66 & 0.51 & 0.76 & 0.88 & 0.53 & 0.57 & 0.55 \\
\midrule
\multicolumn{2}{l}{\textit{Average}} & 0.69 & 0.67 & 0.86 & 0.62 & 0.55 & 0.73 & 0.81 & 0.52 & 0.47 & 0.41 & 0.47 & 0.46 & 0.58 & 0.49 & 0.79 & 0.59 & 0.85 & 0.92 & 0.66 & 0.73 & 0.64 \\
\midrule
\multirow{4}{*}{\texttt{Gemma3 27B}} & ZS & 0.72 & 0.61 & 0.81 & 0.59 & 0.54 & 0.68 & 0.80 & 0.44 & 0.44 & 0.35 & 0.32 & 0.43 & 0.52 & 0.37 & 0.87 & 0.52 & 0.86 & 0.90 & 0.58 & 0.62 & 0.60 \\
& ZS + Task & 0.63 & 0.55 & 0.77 & 0.49 & 0.40 & 0.55 & 0.79 & 0.32 & 0.36 & 0.30 & 0.28 & 0.31 & 0.45 & 0.35 & 0.78 & 0.42 & 0.89 & 0.88 & 0.59 & 0.56 & 0.53 \\
 & FS + Task & 0.83 & 0.93 & 0.96 & 0.68 & 0.62 & 0.81 & 0.93 & 0.74 & 0.67 & 0.52 & 0.68 & 0.60 & 0.70 & 0.75 & 0.97 & 0.74 & 0.96 & 0.98 & 0.81 & 0.86 & 0.79 \\
 & CoT & 0.62 & 0.53 & 0.75 & 0.48 & 0.38 & 0.55 & 0.73 & 0.27 & 0.29 & 0.24 & 0.24 & 0.26 & 0.43 & 0.17 & 0.77 & 0.40 & 0.79 & 0.85 & 0.51 & 0.49 & 0.49 \\
\midrule
\multicolumn{2}{l}{\textit{Average}} & 0.70 & 0.66 & 0.82 & 0.56 & 0.49 & 0.65 & 0.81 & 0.44 & 0.44 & 0.35 & 0.38 & 0.40 & 0.53 & 0.41 & 0.85 & 0.52 & 0.88 & 0.90 & 0.62 & 0.63 & 0.60 \\
\midrule
\multirow{4}{*}{\texttt{Llama3.1 8B}} & ZS & 0.59 & 0.66 & 0.61 & 0.61 & 0.46 & 0.56 & 0.74 & 0.37 & 0.26 & 0.33 & 0.24 & 0.30 & 0.21 & 0.23 & 0.47 & 0.47 & 0.59 & 0.80 & 0.41 & 0.62 & 0.48 \\
 & ZS + Task & 0.82 & 0.92 & 0.88 & 0.84 & 0.75 & 0.90 & 0.91 & 0.75 & 0.59 & 0.65 & 0.63 & 0.65 & 0.70 & 0.76 & 0.85 & 0.79 & 0.98 & 0.94 & 0.83 & 0.86 & 0.80 \\
 & FS + Task & 0.83 & 0.90 & 0.92 & 0.81 & 0.83 & 0.91 & 0.88 & 0.86 & 0.87 & 0.86 & 0.88 & 0.83 & 0.88 & 0.88 & 0.93 & 0.87 & 0.95 & 0.88 & 0.90 & 0.92 & 0.88 \\
 & CoT & 0.77 & 0.75 & 0.82 & 0.72 & 0.71 & 0.80 & 0.78 & 0.63 & 0.61 & 0.62 & 0.65 & 0.64 & 0.55 & 0.60 & 0.74 & 0.71 & 0.90 & 0.84 & 0.73 & 0.83 & 0.72 \\
\midrule
\multicolumn{2}{l}{\textit{Average}} & 0.75 & 0.81 & 0.81 & 0.75 & 0.69 & 0.79 & 0.83 & 0.65 & 0.58 & 0.62 & 0.60 & 0.61 & 0.59 & 0.62 & 0.75 & 0.71 & 0.86 & 0.87 & 0.72 & 0.81 & 0.72 \\
\midrule
\multirow{4}{*}{\texttt{Llama3.1 70B}} & ZS & 0.84 & 0.89 & 0.91 & 0.80 & 0.78 & 0.88 & 0.87 & 0.68 & 0.70& 0.70 & 0.67 & 0.63 & 0.65 & 0.68 & 0.86 & 0.79 & 0.94 & 0.93 & 0.80 & 0.94 & 0.80 \\
 & ZS + Task & 0.78 & 0.90 & 0.92 & 0.77 & 0.69 & 0.84 & 0.92 & 0.64 & 0.67 & 0.61 & 0.68 & 0.58 & 0.67 & 0.71 & 0.90 & 0.73 & 0.99 & 0.95 & 0.81 & 0.91 & 0.78 \\
 & FS + Task & 0.91 & 0.98 & 0.98 & 0.95 & 0.93 & 0.98 & 0.98 & 0.93 & 0.94 & 0.88 & 0.98 & 0.94 & 0.95 & 0.98 & 0.98 & 0.94 & 0.98 & 0.97 & 0.97 & 1.00 & 0.96 \\
 & CoT & 0.81 & 0.91 & 0.90 & 0.82 & 0.74 & 0.88 & 0.89 & 0.71 & 0.83 & 0.64 & 0.76 & 0.73 & 0.82 & 0.78 & 0.84 & 0.82 & 0.89 & 0.92 & 0.87 & 0.90 & 0.82 \\
\midrule
\multicolumn{2}{l}{\textit{Average}} & 0.84 & 0.92 & 0.93 & 0.84 & 0.79 & 0.90 & 0.92 & 0.74 & 0.79 & 0.71 & 0.77 & 0.72 & 0.77 & 0.79 & 0.90 & 0.82 & 0.95 & 0.94 & 0.86 & 0.94 & 0.84 \\
\bottomrule
\end{tabular}}
\caption{The capabilities of LLMs in filtering irrelevant pairs based on TNR (higher is better) using English instruction in cross-lingual settings. The average TNR is calculated across all languages for each model and technique.}
\end{table*}

\begin{table*}
\resizebox{\textwidth}{!}{%
\begin{tabular}{llrrrrrrrrrrrrrrrrrrrrr}
\toprule
\textbf{Model} & \textbf{Technique} & \textbf{spa-eng} & \textbf{hin-eng} & \textbf{eng-ara} & \textbf{fra-eng} & \textbf{deu-eng} & \textbf{eng-por} & \textbf{spa-por} & \textbf{deu-fra} & \textbf{slk-ces} & \textbf{slk-eng} & \textbf{pol-hbs} & \textbf{ces-eng} & \textbf{ces-pol} & \textbf{nld-deu} & \textbf{msa-ara} & \textbf{kor-eng} & \textbf{mya-msa} & \textbf{ara-fra} & \textbf{hun-pol} & \textbf{tha-por} & \textbf{Avg.} \\
\midrule
\multirow{4}{*}{\texttt{Qwen3 8B}} & ZS & \cellcolor{lightcoral}{-0.05} & \cellcolor{lightcoral}{-0.04} & 0.00 & \cellcolor{lightgreen}{0.21} & \cellcolor{lightgreen}{0.08} & 0.00 & \cellcolor{lightcoral}{-0.03} & \cellcolor{lightgreen}{0.11} & \cellcolor{lightgreen}{0.09} & \cellcolor{lightgreen}{0.04} & \cellcolor{lightcoral}{-0.04} & 0.00 & \cellcolor{lightgreen}{0.02} & \cellcolor{lightcoral}{-0.05} & \cellcolor{lightcoral}{-0.41} & \cellcolor{lightcoral}{-0.40} & \cellcolor{lightcoral}{-0.39} & \cellcolor{lightgreen}{0.03} & \cellcolor{lightcoral}{-0.13} & \cellcolor{lightcoral}{-0.46} & \cellcolor{lightcoral}{-0.07} \\
& ZS + Task & \cellcolor{lightgreen}{0.01} & 0.00 & 0.00 & \cellcolor{lightgreen}{0.02} & \cellcolor{lightgreen}{0.13} & 0.00 & \cellcolor{lightgreen}{0.03} & \cellcolor{lightgreen}{0.07} & \cellcolor{lightgreen}{0.01} & \cellcolor{lightcoral}{-0.02} & \cellcolor{lightgreen}{0.07} & \cellcolor{lightgreen}{0.02} & \cellcolor{lightgreen}{0.05} & \cellcolor{lightgreen}{0.11} & \cellcolor{lightcoral}{-0.14} & \cellcolor{lightcoral}{-0.10} & \cellcolor{lightcoral}{-0.14} & \cellcolor{lightcoral}{-0.01} & \cellcolor{lightcoral}{-0.16} & \cellcolor{lightcoral}{-0.14} & \cellcolor{lightcoral}{-0.01} \\
& FS + Task & \cellcolor{lightgreen}{0.10} & \cellcolor{lightgreen}{0.08} & 0.00 & \cellcolor{lightgreen}{0.17} & \cellcolor{lightgreen}{0.14} & 0.00 & \cellcolor{lightgreen}{0.05} & \cellcolor{lightgreen}{0.13} & \cellcolor{lightcoral}{-0.08} & \cellcolor{lightcoral}{-0.08} & \cellcolor{lightgreen}{0.13} & \cellcolor{lightcoral}{-0.12} & \cellcolor{lightcoral}{-0.16} & \cellcolor{lightgreen}{0.19} & \cellcolor{lightgreen}{0.07} & \cellcolor{lightcoral}{-0.19} & \cellcolor{lightcoral}{-0.41} & \cellcolor{lightcoral}{-0.02} & \cellcolor{lightgreen}{0.03} & \cellcolor{lightgreen}{0.16} & \cellcolor{lightgreen}{0.01} \\
& CoT & 0.00 & \cellcolor{lightcoral}{-0.08} & 0.00 & \cellcolor{lightcoral}{-0.06} & \cellcolor{lightcoral}{-0.01} & 0.00 & \cellcolor{lightcoral}{-0.02} & \cellcolor{lightgreen}{0.04} & \cellcolor{lightcoral}{-0.01} & \cellcolor{lightcoral}{-0.03} & \cellcolor{lightcoral}{-0.01} & \cellcolor{lightcoral}{-0.20} & \cellcolor{lightcoral}{-0.12} & \cellcolor{lightcoral}{-0.05} & \cellcolor{lightcoral}{-0.18} & \cellcolor{lightcoral}{-0.25} & \cellcolor{lightcoral}{-0.42} & \cellcolor{lightgreen}{0.01} & \cellcolor{lightcoral}{-0.07} & \cellcolor{lightcoral}{-0.10} & \cellcolor{lightcoral}{-0.08} \\
\midrule
\multicolumn{2}{l}{\textit{Average}} & \cellcolor{lightgreen}{0.01} & \cellcolor{lightcoral}{-0.01} & 0.00 & \cellcolor{lightgreen}{0.09} & \cellcolor{lightgreen}{0.08} & 0.00 & \cellcolor{lightgreen}{0.01} & \cellcolor{lightgreen}{0.08} & 0.00 & \cellcolor{lightcoral}{-0.02} & \cellcolor{lightgreen}{0.04} & \cellcolor{lightcoral}{-0.07} & \cellcolor{lightcoral}{-0.05} & \cellcolor{lightgreen}{0.05} & \cellcolor{lightcoral}{-0.16} & \cellcolor{lightcoral}{-0.24} & \cellcolor{lightcoral}{-0.34} & 0.00 & \cellcolor{lightcoral}{-0.08} & \cellcolor{lightcoral}{-0.13} & \cellcolor{lightcoral}{-0.04} \\

\midrule
\multirow{4}{*}{\texttt{Qwen3 14B}} & ZS & \cellcolor{lightcoral}{-0.05} & \cellcolor{lightcoral}{-0.34} & 0.00 & \cellcolor{lightcoral}{-0.03} & \cellcolor{lightcoral}{-0.05} & 0.00 & \cellcolor{lightcoral}{-0.13} & \cellcolor{lightcoral}{-0.05} & \cellcolor{lightcoral}{-0.07} & \cellcolor{lightcoral}{-0.18} & \cellcolor{lightcoral}{-0.33} & \cellcolor{lightcoral}{-0.10} & \cellcolor{lightcoral}{-0.02} & \cellcolor{lightcoral}{-0.06} & \cellcolor{lightcoral}{-0.18} & \cellcolor{lightcoral}{-0.45} & \cellcolor{lightcoral}{-0.54} & \cellcolor{lightgreen}{0.02} & \cellcolor{lightcoral}{-0.05} & \cellcolor{lightcoral}{-0.46} & \cellcolor{lightcoral}{-0.15} \\
& ZS + Task & \cellcolor{lightcoral}{-0.01} & \cellcolor{lightgreen}{0.04} & 0.00 & \cellcolor{lightgreen}{0.01} & \cellcolor{lightgreen}{0.03} & 0.00 & \cellcolor{lightcoral}{-0.01} & \cellcolor{lightgreen}{0.05} & \cellcolor{lightgreen}{0.11} & \cellcolor{lightgreen}{0.03} & \cellcolor{lightgreen}{0.06} & \cellcolor{lightgreen}{0.03} & \cellcolor{lightgreen}{0.07} & \cellcolor{lightgreen}{0.03} & \cellcolor{lightcoral}{-0.03} & \cellcolor{lightcoral}{-0.01} & \cellcolor{lightcoral}{-0.11} & \cellcolor{lightgreen}{0.01} & \cellcolor{lightgreen}{0.01} & \cellcolor{lightcoral}{-0.01} & \cellcolor{lightgreen}{0.01} \\
& FS + Task & \cellcolor{lightgreen}{0.04} & 0.00 & 0.00 & \cellcolor{lightcoral}{-0.01} & \cellcolor{lightcoral}{-0.02} & 0.00 & 0.00 & \cellcolor{lightcoral}{-0.02} & \cellcolor{lightcoral}{-0.01} & \cellcolor{lightgreen}{0.01} & \cellcolor{lightgreen}{0.05} & \cellcolor{lightgreen}{0.01} & 0.00 & \cellcolor{lightcoral}{-0.02} & 0.00 & \cellcolor{lightgreen}{0.04} & \cellcolor{lightcoral}{-0.03} & 0.00 & \cellcolor{lightcoral}{-0.01} & \cellcolor{lightgreen}{0.02} & 0.00 \\
 & CoT & \cellcolor{lightcoral}{-0.05} & \cellcolor{lightcoral}{-0.02} & 0.00 & \cellcolor{lightcoral}{-0.07} & \cellcolor{lightcoral}{-0.06} & 0.00 & \cellcolor{lightcoral}{-0.03} & \cellcolor{lightcoral}{-0.01} & \cellcolor{lightgreen}{0.01} & \cellcolor{lightcoral}{-0.05} & \cellcolor{lightcoral}{-0.05} & \cellcolor{lightcoral}{-0.07} & \cellcolor{lightcoral}{-0.03} & \cellcolor{lightcoral}{-0.04} & \cellcolor{lightcoral}{-0.61} & \cellcolor{lightcoral}{-0.14} & \cellcolor{lightcoral}{-0.23} & \cellcolor{lightcoral}{-0.02} & \cellcolor{lightcoral}{-0.01} & \cellcolor{lightcoral}{-0.06} & \cellcolor{lightcoral}{-0.08} \\
\midrule
\multicolumn{2}{l}{\textit{Average}} & \cellcolor{lightcoral}{-0.02} & \cellcolor{lightcoral}{-0.08} & 0.00 & \cellcolor{lightcoral}{-0.02} & \cellcolor{lightcoral}{-0.03} & 0.00 & \cellcolor{lightcoral}{-0.04} & \cellcolor{lightcoral}{-0.01} & \cellcolor{lightgreen}{0.01} & \cellcolor{lightcoral}{-0.05} & \cellcolor{lightcoral}{-0.07} & \cellcolor{lightcoral}{-0.03} & 0.00 & \cellcolor{lightcoral}{-0.02} & \cellcolor{lightcoral}{-0.20} & \cellcolor{lightcoral}{-0.14} & \cellcolor{lightcoral}{-0.23} & 0.00 & \cellcolor{lightcoral}{-0.02} & \cellcolor{lightcoral}{-0.13} & \cellcolor{lightcoral}{-0.05} \\

\midrule
\multirow{4}{*}{\texttt{Gemma3 12B}} & ZS & \cellcolor{lightgreen}{0.01} & \cellcolor{lightcoral}{-0.18} & 0.00 & \cellcolor{lightgreen}{0.02} & \cellcolor{lightcoral}{-0.06} & 0.00 & \cellcolor{lightcoral}{-0.01} & \cellcolor{lightcoral}{-0.07} & \cellcolor{lightgreen}{0.18} & \cellcolor{lightgreen}{0.07} & 0.00 & \cellcolor{lightgreen}{0.10} & \cellcolor{lightgreen}{0.08} & \cellcolor{lightgreen}{0.07} & \cellcolor{lightcoral}{-0.41} & \cellcolor{lightcoral}{-0.28} & \cellcolor{lightcoral}{-0.51} & \cellcolor{lightcoral}{-0.02} & \cellcolor{lightcoral}{-0.18} & \cellcolor{lightcoral}{-0.40} & \cellcolor{lightcoral}{-0.08} \\
& ZS + Task & \cellcolor{lightgreen}{0.04} & \cellcolor{lightgreen}{0.18} & 0.00 & \cellcolor{lightgreen}{0.05} & \cellcolor{lightgreen}{0.02} & 0.00 & \cellcolor{lightgreen}{0.01} & \cellcolor{lightcoral}{-0.02} & \cellcolor{lightgreen}{0.15} & \cellcolor{lightgreen}{0.10} & \cellcolor{lightgreen}{0.18} & \cellcolor{lightgreen}{0.03} & 0.00 & \cellcolor{lightgreen}{0.27} & \cellcolor{lightgreen}{0.07} & \cellcolor{lightcoral}{-0.01} & \cellcolor{lightcoral}{-0.13} & \cellcolor{lightgreen}{0.07} & \cellcolor{lightcoral}{-0.04} & \cellcolor{lightgreen}{0.01} & \cellcolor{lightgreen}{0.05} \\
& FS + Task & 0.00 & 0.00 & 0.00 & \cellcolor{lightcoral}{-0.03} & \cellcolor{lightcoral}{-0.10} & 0.00 & 0.00 & \cellcolor{lightcoral}{-0.03} & \cellcolor{lightcoral}{-0.12} & \cellcolor{lightcoral}{-0.08} & \cellcolor{lightgreen}{0.03} & \cellcolor{lightcoral}{-0.10} & \cellcolor{lightcoral}{-0.15} & \cellcolor{lightgreen}{0.04} & \cellcolor{lightcoral}{-0.01} & \cellcolor{lightcoral}{-0.10} & \cellcolor{lightcoral}{-0.02} & \cellcolor{lightcoral}{-0.02} & \cellcolor{lightcoral}{-0.08} & \cellcolor{lightgreen}{0.02} & \cellcolor{lightcoral}{-0.04} \\
& CoT & 0.00 & \cellcolor{lightgreen}{0.06} & 0.00 & \cellcolor{lightcoral}{-0.05} & \cellcolor{lightgreen}{0.16} & 0.00 & \cellcolor{lightgreen}{0.01} & \cellcolor{lightgreen}{0.15} & \cellcolor{lightgreen}{0.09} & \cellcolor{lightgreen}{0.09} & \cellcolor{lightgreen}{0.08} & \cellcolor{lightcoral}{-0.05} & \cellcolor{lightcoral}{-0.04} & \cellcolor{lightgreen}{0.13} & 0.00 & \cellcolor{lightcoral}{-0.22} & \cellcolor{lightcoral}{-0.43} & \cellcolor{lightgreen}{0.02} & \cellcolor{lightgreen}{0.10} & \cellcolor{lightgreen}{0.02} & \cellcolor{lightgreen}{0.01} \\
\midrule
\multicolumn{2}{l}{\textit{Average}} & \cellcolor{lightgreen}{0.01} & \cellcolor{lightgreen}{0.02} & 0.00 & 0.00 & \cellcolor{lightgreen}{0.01} & 0.00 & 0.00 & \cellcolor{lightgreen}{0.01} & \cellcolor{lightgreen}{0.07} & \cellcolor{lightgreen}{0.05} & \cellcolor{lightgreen}{0.07} & \cellcolor{lightcoral}{-0.01} & \cellcolor{lightcoral}{-0.03} & \cellcolor{lightgreen}{0.13} & \cellcolor{lightcoral}{-0.09} & \cellcolor{lightcoral}{-0.15} & \cellcolor{lightcoral}{-0.27} & \cellcolor{lightgreen}{0.01} & \cellcolor{lightcoral}{-0.05} & \cellcolor{lightcoral}{-0.09} & \cellcolor{lightcoral}{-0.02} \\

\midrule
\multirow{4}{*}{\texttt{Gemma3 27B}} & ZS & \cellcolor{lightgreen}{0.03} & \cellcolor{lightcoral}{-0.09} & 0.00 & \cellcolor{lightcoral}{-0.02} & \cellcolor{lightcoral}{-0.05} & 0.00 & \cellcolor{lightcoral}{-0.03} & \cellcolor{lightcoral}{-0.06} & \cellcolor{lightgreen}{0.11} & \cellcolor{lightgreen}{0.04} & \cellcolor{lightcoral}{-0.06} & \cellcolor{lightgreen}{0.05} & \cellcolor{lightgreen}{0.04} & \cellcolor{lightgreen}{0.01} & \cellcolor{lightcoral}{-0.18} & \cellcolor{lightcoral}{-0.05} & \cellcolor{lightcoral}{-0.49} & \cellcolor{lightcoral}{-0.04} & \cellcolor{lightcoral}{-0.15} & \cellcolor{lightcoral}{-0.03} & \cellcolor{lightcoral}{-0.05} \\
& ZS + Task & \cellcolor{lightgreen}{0.03} & \cellcolor{lightgreen}{0.20} & 0.00 & \cellcolor{lightgreen}{0.03} & \cellcolor{lightgreen}{0.13} & 0.00 & \cellcolor{lightgreen}{0.05} & \cellcolor{lightgreen}{0.12} & 0.00 & \cellcolor{lightcoral}{-0.02} & \cellcolor{lightgreen}{0.14} & \cellcolor{lightgreen}{0.03} & \cellcolor{lightgreen}{0.04} & \cellcolor{lightgreen}{0.03} & \cellcolor{lightgreen}{0.07} & \cellcolor{lightgreen}{0.03} & \cellcolor{lightgreen}{0.03} & \cellcolor{lightgreen}{0.05} & \cellcolor{lightcoral}{-0.03} & \cellcolor{lightgreen}{0.12} & \cellcolor{lightgreen}{0.05} \\
& FS + Task & \cellcolor{lightcoral}{-0.06} & \cellcolor{lightcoral}{-0.01} & 0.00 & \cellcolor{lightcoral}{-0.02} & 0.00 & 0.00 & \cellcolor{lightcoral}{-0.02} & \cellcolor{lightcoral}{-0.08} & \cellcolor{lightcoral}{-0.07} & \cellcolor{lightcoral}{-0.04} & \cellcolor{lightgreen}{0.03} & \cellcolor{lightcoral}{-0.09} & \cellcolor{lightcoral}{-0.02} & \cellcolor{lightgreen}{0.03} & 0.00 & \cellcolor{lightcoral}{-0.08} & \cellcolor{lightgreen}{0.04} & \cellcolor{lightcoral}{-0.03} & \cellcolor{lightcoral}{-0.06} & \cellcolor{lightgreen}{0.01} & \cellcolor{lightcoral}{-0.02} \\
& CoT & \cellcolor{lightgreen}{0.01} & \cellcolor{lightgreen}{0.17} & 0.00 & \cellcolor{lightcoral}{-0.06} & \cellcolor{lightgreen}{0.21} & 0.00 & \cellcolor{lightgreen}{0.04} & \cellcolor{lightgreen}{0.25} & \cellcolor{lightgreen}{0.04} & \cellcolor{lightgreen}{0.03} & \cellcolor{lightgreen}{0.07} & \cellcolor{lightgreen}{0.02} & \cellcolor{lightcoral}{-0.04} & \cellcolor{lightgreen}{0.04} & \cellcolor{lightgreen}{0.03} & \cellcolor{lightcoral}{-0.04} & \cellcolor{lightcoral}{-0.02} & \cellcolor{lightgreen}{0.04} & \cellcolor{lightgreen}{0.11} & \cellcolor{lightgreen}{0.10} & \cellcolor{lightgreen}{0.05} \\
\midrule
\multicolumn{2}{l}{\textit{Average}} & 0.00 & \cellcolor{lightgreen}{0.07} & 0.00 & \cellcolor{lightcoral}{-0.02} & \cellcolor{lightgreen}{0.07} & 0.00 & \cellcolor{lightgreen}{0.01} & \cellcolor{lightgreen}{0.06} & \cellcolor{lightgreen}{0.02} & 0.00 & \cellcolor{lightgreen}{0.04} & 0.00 & 0.00 & \cellcolor{lightgreen}{0.03} & \cellcolor{lightcoral}{-0.02} & \cellcolor{lightcoral}{-0.03} & \cellcolor{lightcoral}{-0.11} & \cellcolor{lightgreen}{0.01} & \cellcolor{lightcoral}{-0.03} & \cellcolor{lightgreen}{0.05} & \cellcolor{lightgreen}{0.01} \\

\midrule
\multirow{4}{*}{\texttt{Llama3.1 8B}} & ZS & \cellcolor{lightcoral}{-0.07} & \cellcolor{lightcoral}{-0.15} & 0.00 & \cellcolor{lightcoral}{-0.01} & 0.00 & 0.00 & \cellcolor{lightcoral}{-0.24} & \cellcolor{lightgreen}{0.13} & \cellcolor{lightgreen}{0.46} & \cellcolor{lightgreen}{0.20} & \cellcolor{lightgreen}{0.17} & \cellcolor{lightgreen}{0.04} & \cellcolor{lightgreen}{0.09} & 0.00 & \cellcolor{lightcoral}{-0.06} & \cellcolor{lightgreen}{0.02} & \cellcolor{lightcoral}{-0.24} & \cellcolor{lightgreen}{0.13} & \cellcolor{lightcoral}{-0.15} & \cellcolor{lightcoral}{-0.53} & \cellcolor{lightcoral}{-0.01} \\
& ZS + Task & \cellcolor{lightgreen}{0.02} & \cellcolor{lightcoral}{-0.30} & 0.00 & \cellcolor{lightgreen}{0.07} & \cellcolor{lightgreen}{0.18} & 0.00 & \cellcolor{lightgreen}{0.01} & \cellcolor{lightgreen}{0.19} & \cellcolor{lightgreen}{0.36} & \cellcolor{lightgreen}{0.24} & \cellcolor{lightgreen}{0.13} & \cellcolor{lightcoral}{-0.02} & \cellcolor{lightgreen}{0.02} & \cellcolor{lightgreen}{0.15} & \cellcolor{lightcoral}{-0.20} & \cellcolor{lightcoral}{-0.75} & \cellcolor{lightcoral}{-0.11} & \cellcolor{lightgreen}{0.05} & \cellcolor{lightgreen}{0.06} & \cellcolor{lightcoral}{-0.04} & 0.00 \\
& FS + Task & \cellcolor{lightgreen}{0.06} & \cellcolor{lightgreen}{0.09} & 0.00 & \cellcolor{lightgreen}{0.03} & \cellcolor{lightgreen}{0.06} & 0.00 & \cellcolor{lightgreen}{0.04} & \cellcolor{lightgreen}{0.06} & \cellcolor{lightgreen}{0.09} & \cellcolor{lightgreen}{0.05} & \cellcolor{lightgreen}{0.03} & \cellcolor{lightcoral}{-0.21} & \cellcolor{lightcoral}{-0.16} & 0.00 & \cellcolor{lightcoral}{-0.01} & \cellcolor{lightcoral}{-0.01} & \cellcolor{lightcoral}{-0.11} & \cellcolor{lightgreen}{0.11} & \cellcolor{lightgreen}{0.04} & \cellcolor{lightcoral}{-0.03} & \cellcolor{lightgreen}{0.01} \\
& CoT & \cellcolor{lightcoral}{-0.13} & \cellcolor{lightcoral}{-0.37} & 0.00 & \cellcolor{lightcoral}{-0.10} & \cellcolor{lightgreen}{0.08} & 0.00 & \cellcolor{lightcoral}{-0.13} & \cellcolor{lightgreen}{0.11} & \cellcolor{lightgreen}{0.23} & \cellcolor{lightgreen}{0.09} & \cellcolor{lightgreen}{0.03} & \cellcolor{lightcoral}{-0.09} & \cellcolor{lightgreen}{0.01} & \cellcolor{lightgreen}{0.07} & \cellcolor{lightcoral}{-0.28} & \cellcolor{lightcoral}{-0.22} & \cellcolor{lightcoral}{-0.17} & \cellcolor{lightgreen}{0.11} & \cellcolor{lightgreen}{0.02} & \cellcolor{lightcoral}{-0.06} & \cellcolor{lightcoral}{-0.04} \\
\midrule
\multicolumn{2}{l}{\textit{Average}} & \cellcolor{lightcoral}{-0.03} & \cellcolor{lightcoral}{-0.18} & 0.00 & 0.00 & \cellcolor{lightgreen}{0.08} & 0.00 & \cellcolor{lightcoral}{-0.08} & \cellcolor{lightgreen}{0.12} & \cellcolor{lightgreen}{0.29} & \cellcolor{lightgreen}{0.15} & \cellcolor{lightgreen}{0.09} & \cellcolor{lightcoral}{-0.07} & \cellcolor{lightcoral}{-0.01} & \cellcolor{lightgreen}{0.06} & \cellcolor{lightcoral}{-0.14} & \cellcolor{lightcoral}{-0.24} & \cellcolor{lightcoral}{-0.16} & \cellcolor{lightgreen}{0.10} & \cellcolor{lightcoral}{-0.01} & \cellcolor{lightcoral}{-0.16} & \cellcolor{lightcoral}{-0.01} \\

\midrule
\multirow{4}{*}{\texttt{Llama3.1 70B}} & ZS & \cellcolor{lightcoral}{-0.02} & \cellcolor{lightgreen}{0.06} & 0.00 & \cellcolor{lightcoral}{-0.03} & \cellcolor{lightcoral}{-0.06} & 0.00 & \cellcolor{lightgreen}{0.01} & 0.00 & \cellcolor{lightgreen}{0.23} & \cellcolor{lightgreen}{0.07} & \cellcolor{lightgreen}{0.07} & \cellcolor{lightgreen}{0.15} & \cellcolor{lightgreen}{0.23} & 0.00 & \cellcolor{lightcoral}{-0.16} & \cellcolor{lightcoral}{-0.09} & \cellcolor{lightgreen}{0.06} & \cellcolor{lightgreen}{0.05} & \cellcolor{lightgreen}{0.05} & \cellcolor{lightcoral}{-0.27} & \cellcolor{lightgreen}{0.02} \\
& ZS + Task & \cellcolor{lightgreen}{0.02} & \cellcolor{lightgreen}{0.05} & 0.00 & \cellcolor{lightgreen}{0.07} & \cellcolor{lightgreen}{0.09} & 0.00 & \cellcolor{lightgreen}{0.02} & \cellcolor{lightgreen}{0.12} & \cellcolor{lightgreen}{0.24} & \cellcolor{lightgreen}{0.19} & \cellcolor{lightgreen}{0.14} & \cellcolor{lightgreen}{0.10} & \cellcolor{lightgreen}{0.11} & \cellcolor{lightgreen}{0.05} & \cellcolor{lightcoral}{-0.03} & \cellcolor{lightgreen}{0.01} & \cellcolor{lightgreen}{0.01} & \cellcolor{lightgreen}{0.05} & \cellcolor{lightgreen}{0.14} & \cellcolor{lightgreen}{0.09} & \cellcolor{lightgreen}{0.07} \\
& FS + Task & \cellcolor{lightcoral}{-0.16} & \cellcolor{lightcoral}{-0.18} & 0.00 & \cellcolor{lightcoral}{-0.03} & \cellcolor{lightcoral}{-0.11} & 0.00 & \cellcolor{lightcoral}{-0.21} & \cellcolor{lightcoral}{-0.08} & \cellcolor{lightcoral}{-0.29} & \cellcolor{lightcoral}{-0.33} & \cellcolor{lightcoral}{-0.12} & \cellcolor{lightcoral}{-0.16} & \cellcolor{lightcoral}{-0.11} & \cellcolor{lightcoral}{-0.05} & \cellcolor{lightcoral}{-0.20} & \cellcolor{lightcoral}{-0.01} & \cellcolor{lightcoral}{-0.03} & \cellcolor{lightcoral}{-0.42} & \cellcolor{lightcoral}{-0.25} & \cellcolor{lightcoral}{-0.05} & \cellcolor{lightcoral}{-0.14} \\
& CoT & \cellcolor{lightcoral}{-0.10} & \cellcolor{lightcoral}{-0.03} & 0.00 & \cellcolor{lightcoral}{-0.08} & \cellcolor{lightgreen}{0.08} & 0.00 & \cellcolor{lightcoral}{-0.08} & \cellcolor{lightgreen}{0.14} & \cellcolor{lightgreen}{0.08} & \cellcolor{lightgreen}{0.18} & \cellcolor{lightcoral}{-0.01} & \cellcolor{lightcoral}{-0.05} & 0.00 & \cellcolor{lightgreen}{0.06} & \cellcolor{lightcoral}{-0.07} & \cellcolor{lightcoral}{-0.23} & \cellcolor{lightgreen}{0.08} & \cellcolor{lightgreen}{0.01} & \cellcolor{lightgreen}{0.01} & \cellcolor{lightgreen}{0.06} & 0.00 \\
\midrule
\multicolumn{2}{l}{\textit{Average}} & \cellcolor{lightcoral}{-0.06} & \cellcolor{lightcoral}{-0.03} & 0.00 & \cellcolor{lightcoral}{-0.02} & 0.00 & 0.00 & \cellcolor{lightcoral}{-0.06} & \cellcolor{lightgreen}{0.04} & \cellcolor{lightgreen}{0.07} & \cellcolor{lightgreen}{0.03} & \cellcolor{lightgreen}{0.02} & \cellcolor{lightgreen}{0.01} & \cellcolor{lightgreen}{0.06} & \cellcolor{lightgreen}{0.02} & \cellcolor{lightcoral}{-0.12} & \cellcolor{lightcoral}{-0.08} & \cellcolor{lightgreen}{0.03} & \cellcolor{lightcoral}{-0.08} & \cellcolor{lightcoral}{-0.01} & \cellcolor{lightcoral}{-0.04} & \cellcolor{lightcoral}{-0.01} \\

\bottomrule
\end{tabular}}
\caption{Difference in TNR when using post-language instructions versus English in cross-lingual settings. Each column represents a language pair (post–claim language), with the first language indicating the instruction language. Positive values (\colorbox{lightgreen}{green}) reflect improved filtering when using post-language instructions. The table presents results for six models and four prompting techniques, with average values computed for each model–prompting combination as well as for each language pair.}
\end{table*}

\begin{table*}[]
\resizebox{\textwidth}{!}{%
\begin{tabular}{llrrrrrrrrrrrrrrrrrrrrr}
\toprule
\textbf{Model} & \textbf{Technique} & \textbf{spa-eng} & \textbf{hin-eng} & \textbf{eng-ara} & \textbf{fra-eng} & \textbf{deu-eng} & \textbf{eng-por} & \textbf{spa-por} & \textbf{deu-fra} & \textbf{slk-ces} & \textbf{slk-eng} & \textbf{pol-hbs} & \textbf{ces-eng} & \textbf{ces-pol} & \textbf{nld-deu} & \textbf{msa-ara} & \textbf{kor-eng} & \textbf{mya-msa} & \textbf{ara-fra} & \textbf{hun-pol} & \textbf{tha-por} & \textbf{Avg.} \\
\midrule
\multirow{4}{*}{\texttt{Qwen3 8B}} & ZS & 0.00 & 0.00 & \cellcolor{lightcoral}{-0.02} & 0.00 & 0.00 & \cellcolor{lightcoral}{-0.03} & \cellcolor{lightgreen}{0.03} & \cellcolor{lightgreen}{0.24} & \cellcolor{lightgreen}{0.03} & 0.00 & \cellcolor{lightcoral}{-0.04} & 0.00 & 0.00 & \cellcolor{lightgreen}{0.12} & \cellcolor{lightcoral}{-0.04} & 0.00 & \cellcolor{lightcoral}{-0.32} & \cellcolor{lightgreen}{0.04} & \cellcolor{lightcoral}{-0.12} & \cellcolor{lightcoral}{-0.08} & \cellcolor{lightcoral}{-0.01} \\
& ZS + Task & 0.00 & 0.00 & \cellcolor{lightcoral}{-0.05} & 0.00 & 0.00 & \cellcolor{lightcoral}{-0.07} & \cellcolor{lightcoral}{-0.02} & \cellcolor{lightcoral}{-0.02} & \cellcolor{lightgreen}{0.08} & 0.00 & \cellcolor{lightgreen}{0.01} & 0.00 & \cellcolor{lightgreen}{0.03} & \cellcolor{lightgreen}{0.09} & \cellcolor{lightcoral}{-0.06} & 0.00 & \cellcolor{lightcoral}{-0.14} & \cellcolor{lightcoral}{-0.02} & \cellcolor{lightgreen}{0.01} & \cellcolor{lightcoral}{-0.10} & \cellcolor{lightcoral}{-0.01} \\
& FS + Task & 0.00 & 0.00 & \cellcolor{lightgreen}{0.02} & 0.00 & 0.00 & \cellcolor{lightgreen}{0.13} & \cellcolor{lightgreen}{0.03} & \cellcolor{lightgreen}{0.15} & \cellcolor{lightcoral}{-0.19} & 0.00 & \cellcolor{lightgreen}{0.15} & 0.00 & \cellcolor{lightcoral}{-0.03} & \cellcolor{lightgreen}{0.13} & \cellcolor{lightcoral}{-0.05} & 0.00 & 0.00 & \cellcolor{lightgreen}{0.01} & 0.00 & \cellcolor{lightgreen}{0.18} & \cellcolor{lightgreen}{0.03} \\
& CoT & 0.00 & 0.00 & \cellcolor{lightcoral}{-0.01} & 0.00 & 0.00 & \cellcolor{lightcoral}{-0.03} & \cellcolor{lightcoral}{-0.03} & \cellcolor{lightcoral}{-0.07} & \cellcolor{lightcoral}{-0.10} & 0.00 & \cellcolor{lightcoral}{-0.07} & 0.00 & \cellcolor{lightcoral}{-0.04} & 0.00 & \cellcolor{lightgreen}{0.02} & 0.00 & \cellcolor{lightcoral}{-0.20} & \cellcolor{lightcoral}{-0.02} & \cellcolor{lightgreen}{0.01} & \cellcolor{lightcoral}{-0.05} & \cellcolor{lightcoral}{-0.03} \\
\midrule
\multicolumn{2}{l}{\textit{Average}} & 0.00 & 0.00 & \cellcolor{lightcoral}{-0.02} & 0.00 & 0.00 & 0.00 & 0.00 & \cellcolor{lightgreen}{0.07} & \cellcolor{lightcoral}{-0.05} & 0.00 & \cellcolor{lightgreen}{0.02} & 0.00 & \cellcolor{lightcoral}{-0.01} & \cellcolor{lightgreen}{0.09} & \cellcolor{lightcoral}{-0.03} & 0.00 & \cellcolor{lightcoral}{-0.17} & 0.00 & \cellcolor{lightcoral}{-0.02} & \cellcolor{lightcoral}{-0.01} & \cellcolor{lightcoral}{-0.01} \\

\midrule
\multirow{4}{*}{\texttt{Qwen3 14B}} & ZS & 0.00 & 0.00 & \cellcolor{lightgreen}{0.01} & 0.00 & 0.00 & \cellcolor{lightcoral}{-0.11} & \cellcolor{lightcoral}{-0.12} & \cellcolor{lightcoral}{-0.07} & \cellcolor{lightgreen}{0.06} & 0.00 & \cellcolor{lightgreen}{0.14} & 0.00 & \cellcolor{lightcoral}{-0.30} & \cellcolor{lightgreen}{0.07} & \cellcolor{lightgreen}{0.02} & 0.00 & \cellcolor{lightcoral}{-0.21} & \cellcolor{lightcoral}{-0.04} & \cellcolor{lightcoral}{-0.23} & 0.00 & \cellcolor{lightcoral}{-0.04} \\
& ZS + Task & 0.00 & 0.00 & \cellcolor{lightgreen}{0.03} & 0.00 & 0.00 & \cellcolor{lightcoral}{-0.01} & \cellcolor{lightcoral}{-0.01} & \cellcolor{lightgreen}{0.02} & \cellcolor{lightgreen}{0.14} & 0.00 & \cellcolor{lightgreen}{0.06} & 0.00 & \cellcolor{lightgreen}{0.07} & \cellcolor{lightgreen}{0.02} & \cellcolor{lightgreen}{0.01} & 0.00 & \cellcolor{lightcoral}{-0.02} & \cellcolor{lightgreen}{0.01} & \cellcolor{lightgreen}{0.07} & \cellcolor{lightgreen}{0.03} & \cellcolor{lightgreen}{0.02} \\
& FS + Task & 0.00 & 0.00 & \cellcolor{lightcoral}{-0.01} & 0.00 & 0.00 & \cellcolor{lightcoral}{-0.05} & \cellcolor{lightcoral}{-0.03} & \cellcolor{lightcoral}{-0.02} & \cellcolor{lightcoral}{-0.03} & 0.00 & \cellcolor{lightgreen}{0.02} & 0.00 & \cellcolor{lightgreen}{0.07} & \cellcolor{lightcoral}{-0.11} & \cellcolor{lightgreen}{0.02} & 0.00 & \cellcolor{lightcoral}{-0.03} & 0.00 & \cellcolor{lightgreen}{0.03} & \cellcolor{lightgreen}{0.01} & \cellcolor{lightcoral}{-0.01} \\
& CoT & 0.00 & 0.00 & \cellcolor{lightcoral}{-0.03} & 0.00 & 0.00 & \cellcolor{lightcoral}{-0.04} & \cellcolor{lightcoral}{-0.04} & \cellcolor{lightcoral}{-0.07} & \cellcolor{lightcoral}{-0.04} & 0.00 & \cellcolor{lightcoral}{-0.06} & 0.00 & \cellcolor{lightcoral}{-0.04} & \cellcolor{lightcoral}{-0.01} & \cellcolor{lightcoral}{-0.02} & 0.00 & \cellcolor{lightcoral}{-0.15} & \cellcolor{lightcoral}{-0.01} & \cellcolor{lightcoral}{-0.01} & \cellcolor{lightcoral}{-0.01} & \cellcolor{lightcoral}{-0.03} \\
\midrule
\multicolumn{2}{l}{\textit{Average}} & 0.00 & 0.00 & 0.00 & 0.00 & 0.00 & \cellcolor{lightcoral}{-0.05} & \cellcolor{lightcoral}{-0.05} & \cellcolor{lightcoral}{-0.03} & \cellcolor{lightgreen}{0.03} & 0.00 & \cellcolor{lightgreen}{0.04} & 0.00 & \cellcolor{lightcoral}{-0.05} & \cellcolor{lightcoral}{-0.01} & 0.00 & 0.00 & \cellcolor{lightcoral}{-0.10} & \cellcolor{lightcoral}{-0.01} & \cellcolor{lightcoral}{-0.04} & \cellcolor{lightgreen}{0.01} & \cellcolor{lightcoral}{-0.01} \\

\midrule
\multirow{4}{*}{\texttt{Gemma3 12B}} & ZS & 0.00 & 0.00 & \cellcolor{lightcoral}{-0.02} & 0.00 & 0.00 & \cellcolor{lightgreen}{0.03} & \cellcolor{lightcoral}{-0.03} & \cellcolor{lightgreen}{0.05} & \cellcolor{lightgreen}{0.18} & 0.00 & \cellcolor{lightgreen}{0.12} & 0.00 & \cellcolor{lightcoral}{-0.08} & \cellcolor{lightcoral}{-0.12} & \cellcolor{lightgreen}{0.04} & 0.00 & \cellcolor{lightcoral}{-0.56} & \cellcolor{lightcoral}{-0.02} & \cellcolor{lightcoral}{-0.08} & \cellcolor{lightgreen}{0.02} & \cellcolor{lightcoral}{-0.02} \\
& ZS + Task & 0.00 & 0.00 & \cellcolor{lightgreen}{0.09} & 0.00 & 0.00 & \cellcolor{lightgreen}{0.03} & \cellcolor{lightgreen}{0.05} & \cellcolor{lightcoral}{-0.01} & \cellcolor{lightgreen}{0.03} & 0.00 & \cellcolor{lightgreen}{0.08} & 0.00 & \cellcolor{lightgreen}{0.18} & 0.00 & \cellcolor{lightgreen}{0.12} & 0.00 & \cellcolor{lightgreen}{0.02} & \cellcolor{lightcoral}{-0.02} & \cellcolor{lightgreen}{0.15} & \cellcolor{lightcoral}{-0.02} & \cellcolor{lightgreen}{0.04} \\
 & FS + Task & 0.00 & 0.00 & \cellcolor{lightcoral}{-0.01} & 0.00 & 0.00 & \cellcolor{lightcoral}{-0.06} & \cellcolor{lightcoral}{-0.02} & \cellcolor{lightcoral}{-0.14} & \cellcolor{lightcoral}{-0.13} & 0.00 & \cellcolor{lightcoral}{-0.03} & 0.00 & \cellcolor{lightcoral}{-0.05} & \cellcolor{lightcoral}{-0.11} & \cellcolor{lightcoral}{-0.01} & 0.00 & 0.00 & \cellcolor{lightcoral}{-0.02} & \cellcolor{lightcoral}{-0.01} & \cellcolor{lightcoral}{-0.07} & \cellcolor{lightcoral}{-0.03} \\
 & CoT & 0.00 & 0.00 & \cellcolor{lightgreen}{0.07} & 0.00 & 0.00 & \cellcolor{lightcoral}{-0.10} & \cellcolor{lightcoral}{-0.07} & \cellcolor{lightgreen}{0.01} & \cellcolor{lightgreen}{0.06} & 0.00 & \cellcolor{lightgreen}{0.18} & 0.00 & \cellcolor{lightgreen}{0.03} & \cellcolor{lightgreen}{0.11} & \cellcolor{lightgreen}{0.12} & 0.00 & \cellcolor{lightcoral}{-0.07} & \cellcolor{lightcoral}{-0.03} & \cellcolor{lightgreen}{0.04} & \cellcolor{lightcoral}{-0.05} & \cellcolor{lightgreen}{0.01} \\
\midrule
\multicolumn{2}{l}{\textit{Average}} & 0.00 & 0.00 & \cellcolor{lightgreen}{0.03} & 0.00 & 0.00 & \cellcolor{lightcoral}{-0.02} & \cellcolor{lightcoral}{-0.02} & \cellcolor{lightcoral}{-0.03} & \cellcolor{lightgreen}{0.04} & 0.00 & \cellcolor{lightgreen}{0.09} & 0.00 & \cellcolor{lightgreen}{0.02} & \cellcolor{lightcoral}{-0.03} & \cellcolor{lightgreen}{0.07} & 0.00 & \cellcolor{lightcoral}{-0.15} & \cellcolor{lightcoral}{-0.02} & \cellcolor{lightgreen}{0.03} & \cellcolor{lightcoral}{-0.03} & 0.00 \\

\midrule
\multirow{4}{*}{\texttt{Gemma3 27B}} & ZS & 0.00 & 0.00 & \cellcolor{lightcoral}{-0.01} & 0.00 & 0.00 & \cellcolor{lightgreen}{0.06} & \cellcolor{lightgreen}{0.02} & \cellcolor{lightgreen}{0.19} & \cellcolor{lightgreen}{0.17} & 0.00 & \cellcolor{lightgreen}{0.38} & 0.00 & \cellcolor{lightcoral}{-0.09} & \cellcolor{lightgreen}{0.01} & \cellcolor{lightgreen}{0.01} & 0.00 & \cellcolor{lightcoral}{-0.11} & \cellcolor{lightcoral}{-0.01} & \cellcolor{lightcoral}{-0.03} & \cellcolor{lightgreen}{0.16} & \cellcolor{lightgreen}{0.04} \\
& ZS + Task & 0.00 & 0.00 & \cellcolor{lightgreen}{0.09} & 0.00 & 0.00 & \cellcolor{lightgreen}{0.08} & \cellcolor{lightgreen}{0.04} & \cellcolor{lightgreen}{0.16} & \cellcolor{lightgreen}{0.10} & 0.00 & \cellcolor{lightgreen}{0.10} & 0.00 & \cellcolor{lightgreen}{0.16} & \cellcolor{lightgreen}{0.18} & \cellcolor{lightgreen}{0.08} & 0.00 & \cellcolor{lightgreen}{0.04} & \cellcolor{lightgreen}{0.04} & \cellcolor{lightgreen}{0.12} & \cellcolor{lightgreen}{0.01} & \cellcolor{lightgreen}{0.06} \\
& FS + Task & 0.00 & 0.00 & \cellcolor{lightcoral}{-0.01} & 0.00 & 0.00 & \cellcolor{lightgreen}{0.01} & \cellcolor{lightgreen}{0.01} & \cellcolor{lightcoral}{-0.07} & 0.00 & 0.00 & \cellcolor{lightgreen}{0.01} & 0.00 & \cellcolor{lightgreen}{0.03} & \cellcolor{lightgreen}{0.04} & \cellcolor{lightcoral}{-0.01} & 0.00 & \cellcolor{lightgreen}{0.02} & \cellcolor{lightcoral}{-0.02} & \cellcolor{lightcoral}{-0.01} & \cellcolor{lightcoral}{-0.01} & 0.00 \\
& CoT & 0.00 & 0.00 & \cellcolor{lightgreen}{0.12} & 0.00 & 0.00 & \cellcolor{lightgreen}{0.06} & \cellcolor{lightgreen}{0.05} & \cellcolor{lightgreen}{0.02} & \cellcolor{lightgreen}{0.08} & 0.00 & \cellcolor{lightcoral}{-0.01} & 0.00 & \cellcolor{lightgreen}{0.05} & \cellcolor{lightgreen}{0.28} & \cellcolor{lightgreen}{0.06} & 0.00 & \cellcolor{lightgreen}{0.03} & \cellcolor{lightcoral}{-0.11} & \cellcolor{lightgreen}{0.04} & \cellcolor{lightgreen}{0.01} & \cellcolor{lightgreen}{0.03} \\
\midrule
\multicolumn{2}{l}{\textit{Average}} & 0.00 & 0.00 & \cellcolor{lightgreen}{0.05} & 0.00 & 0.00 & \cellcolor{lightgreen}{0.05} & \cellcolor{lightgreen}{0.03} & \cellcolor{lightgreen}{0.07} & \cellcolor{lightgreen}{0.09} & 0.00 & \cellcolor{lightgreen}{0.12} & 0.00 & \cellcolor{lightgreen}{0.04} & \cellcolor{lightgreen}{0.13} & \cellcolor{lightgreen}{0.04} & 0.00 & \cellcolor{lightcoral}{-0.01} & \cellcolor{lightcoral}{-0.02} & \cellcolor{lightgreen}{0.03} & \cellcolor{lightgreen}{0.04} & \cellcolor{lightgreen}{0.03} \\

\midrule
\multirow{4}{*}{\texttt{Llama3.1 8B}} & ZS & 0.00 & 0.00 & \cellcolor{lightgreen}{0.14} & 0.00 & 0.00 & \cellcolor{lightcoral}{-0.27} & \cellcolor{lightcoral}{-0.50} & \cellcolor{lightgreen}{0.14} & \cellcolor{lightgreen}{0.20} & 0.00 & \cellcolor{lightgreen}{0.49} & 0.00 & \cellcolor{lightgreen}{0.12} & \cellcolor{lightgreen}{0.17} & \cellcolor{lightgreen}{0.33} & 0.00 & \cellcolor{lightcoral}{-0.32} & \cellcolor{lightgreen}{0.01} & \cellcolor{lightgreen}{0.08} & \cellcolor{lightcoral}{-0.24} & \cellcolor{lightgreen}{0.02} \\
& ZS + Task & 0.00 & 0.00 & \cellcolor{lightgreen}{0.07} & 0.00 & 0.00 & \cellcolor{lightcoral}{-0.09} & \cellcolor{lightcoral}{-0.02} & \cellcolor{lightgreen}{0.13} & \cellcolor{lightgreen}{0.16} & 0.00 & \cellcolor{lightcoral}{-0.12} & 0.00 & \cellcolor{lightgreen}{0.06} & \cellcolor{lightgreen}{0.16} & \cellcolor{lightgreen}{0.09} & 0.00 & \cellcolor{lightcoral}{-0.30} & \cellcolor{lightgreen}{0.04} & \cellcolor{lightgreen}{0.04} & \cellcolor{lightcoral}{-0.11} & \cellcolor{lightgreen}{0.01} \\
& FS + Task & 0.00 & 0.00 & \cellcolor{lightgreen}{0.06} & 0.00 & 0.00 & \cellcolor{lightgreen}{0.01} & \cellcolor{lightgreen}{0.05} & \cellcolor{lightcoral}{-0.06} & \cellcolor{lightcoral}{-0.15} & 0.00 & \cellcolor{lightcoral}{-0.13} & 0.00 & \cellcolor{lightcoral}{-0.01} & \cellcolor{lightgreen}{0.03} & \cellcolor{lightgreen}{0.03} & 0.00 & \cellcolor{lightcoral}{-0.05} & \cellcolor{lightgreen}{0.04} & \cellcolor{lightcoral}{-0.04} & \cellcolor{lightgreen}{0.02} & \cellcolor{lightcoral}{-0.01} \\
& CoT & 0.00 & 0.00 & \cellcolor{lightgreen}{0.08} & 0.00 & 0.00 & \cellcolor{lightgreen}{0.12} & \cellcolor{lightgreen}{0.13} & \cellcolor{lightcoral}{-0.01} & \cellcolor{lightgreen}{0.04} & 0.00 & \cellcolor{lightcoral}{-0.08} & 0.00 & \cellcolor{lightgreen}{0.24} & \cellcolor{lightgreen}{0.09} & \cellcolor{lightgreen}{0.15} & 0.00 & \cellcolor{lightcoral}{-0.32} & \cellcolor{lightgreen}{0.02} & \cellcolor{lightgreen}{0.19} & \cellcolor{lightgreen}{0.11} & \cellcolor{lightgreen}{0.04} \\
\midrule
\multicolumn{2}{l}{\textit{Average}} & 0.00 & 0.00 & \cellcolor{lightgreen}{0.09} & 0.00 & 0.00 & \cellcolor{lightcoral}{-0.06} & \cellcolor{lightcoral}{-0.08} & \cellcolor{lightgreen}{0.05} & \cellcolor{lightgreen}{0.06} & 0.00 & \cellcolor{lightgreen}{0.04} & 0.00 & \cellcolor{lightgreen}{0.11} & \cellcolor{lightgreen}{0.11} & \cellcolor{lightgreen}{0.15} & 0.00 & \cellcolor{lightcoral}{-0.25} & \cellcolor{lightgreen}{0.03} & \cellcolor{lightgreen}{0.07} & \cellcolor{lightcoral}{-0.06} & \cellcolor{lightgreen}{0.01} \\
\midrule
\multirow{4}{*}{\texttt{Llama3.1 70B}} & ZS & 0.00 & 0.00 & \cellcolor{lightgreen}{0.05} & 0.00 & 0.00 & \cellcolor{lightgreen}{0.01} & \cellcolor{lightgreen}{0.03} & \cellcolor{lightcoral}{-0.05} & \cellcolor{lightgreen}{0.19} & 0.00 & \cellcolor{lightgreen}{0.30} & 0.00 & \cellcolor{lightgreen}{0.18} & \cellcolor{lightcoral}{-0.03} & \cellcolor{lightgreen}{0.08} & 0.00 & \cellcolor{lightcoral}{-0.24} & \cellcolor{lightcoral}{-0.04} & \cellcolor{lightgreen}{0.03} & \cellcolor{lightgreen}{0.01} & \cellcolor{lightgreen}{0.03} \\
 & ZS + Task & 0.00 & 0.00 & \cellcolor{lightgreen}{0.08} & 0.00 & 0.00 & \cellcolor{lightgreen}{0.03} & 0.00 & \cellcolor{lightgreen}{0.17} & \cellcolor{lightgreen}{0.15} & 0.00 & \cellcolor{lightgreen}{0.21} & 0.00 & \cellcolor{lightgreen}{0.15} & \cellcolor{lightgreen}{0.13} & \cellcolor{lightgreen}{0.09} & 0.00 & \cellcolor{lightcoral}{-0.24} & \cellcolor{lightgreen}{0.02} & \cellcolor{lightgreen}{0.10} & \cellcolor{lightgreen}{0.04} & \cellcolor{lightgreen}{0.05} \\
 & FS + Task & 0.00 & 0.00 & \cellcolor{lightcoral}{-0.33} & 0.00 & 0.00 & \cellcolor{lightcoral}{-0.04} & \cellcolor{lightcoral}{-0.10} & \cellcolor{lightcoral}{-0.05} & \cellcolor{lightcoral}{-0.16} & 0.00 & \cellcolor{lightcoral}{-0.18} & 0.00 & \cellcolor{lightcoral}{-0.10} & \cellcolor{lightcoral}{-0.13} & \cellcolor{lightcoral}{-0.32} & 0.00 & \cellcolor{lightcoral}{-0.18} & \cellcolor{lightcoral}{-0.08} & \cellcolor{lightcoral}{-0.08} & \cellcolor{lightcoral}{-0.08} & \cellcolor{lightcoral}{-0.09} \\
 & CoT & 0.00 & 0.00 & \cellcolor{lightgreen}{0.01} & 0.00 & 0.00 & 0.00 & \cellcolor{lightcoral}{-0.01} & \cellcolor{lightcoral}{-0.01} & \cellcolor{lightcoral}{-0.01} & 0.00 & 0.00 & 0.00 & \cellcolor{lightcoral}{-0.05} & \cellcolor{lightgreen}{0.08} & \cellcolor{lightcoral}{-0.10} & 0.00 & \cellcolor{lightcoral}{-0.32} & \cellcolor{lightcoral}{-0.03} & \cellcolor{lightgreen}{0.02} & \cellcolor{lightgreen}{0.04} & \cellcolor{lightcoral}{-0.02} \\
\midrule
\multicolumn{2}{l}{\textit{Average}} & 0.00 & 0.00 & \cellcolor{lightcoral}{-0.05} & 0.00 & 0.00 & 0.00 & \cellcolor{lightcoral}{-0.02} & \cellcolor{lightgreen}{0.02} & \cellcolor{lightgreen}{0.04} & 0.00 & \cellcolor{lightgreen}{0.08} & 0.00 & \cellcolor{lightgreen}{0.05} & \cellcolor{lightgreen}{0.01} & \cellcolor{lightcoral}{-0.06} & 0.00 & \cellcolor{lightcoral}{-0.24} & \cellcolor{lightcoral}{-0.03} & \cellcolor{lightgreen}{0.02} & 0.00 & \cellcolor{lightcoral}{-0.01} \\

\bottomrule
\end{tabular}}
\caption{Difference in TNR when using claim-language instructions versus English in cross-lingual settings. Each column represents a language pair (post–claim language), with the second language indicating the instruction language. Positive values (\colorbox{lightgreen}{green}) reflect improved filtering when using claim-language instructions. The table presents results for six models and four prompting techniques, with average values computed for each model–prompting combination as well as for each language pair.}
\label{tab:tnr-claim-diff}
\end{table*}

\begin{table*}
\resizebox{\textwidth}{!}{%
\begin{tabular}{lcrrrrrrrrrrrrrrrrrrrrr}
\toprule
\textbf{Technique} & \textbf{Thinking} & \textbf{ara} & \textbf{bul} & \textbf{ces} & \textbf{deu} & \textbf{ell} & \textbf{eng} & \textbf{fra} & \textbf{hbs} & \textbf{hin} & \textbf{hun} & \textbf{kor} & \textbf{msa} & \textbf{mya} & \textbf{nld} & \textbf{pol} & \textbf{por} & \textbf{ron} & \textbf{slk} & \textbf{spa} & \textbf{tha} & \textbf{Avg.} \\
\midrule
\multirow{2}{*}{\makecell[l]{Zero-Shot}} & \xmark & 0.79 & 0.85 & 0.68 & 0.58 & 0.84 & 0.76 &0.80& 0.61 & 0.76 & 0.84 & 0.90& 0.66 & 0.78 & 0.74 & 0.61 & 0.64 & 0.90& 0.78 & 0.71 & 0.86 & 0.75 \\
& \checkmark & 0.85 & 0.97 & 0.78 & 0.70& 0.86 & 0.76 & 0.86 & 0.61 & 0.87 & 0.85 & 0.94 & 0.86 & 0.85 & 0.79 & 0.72 & 0.67 & 0.86 & 0.84 & 0.78 & 0.94 & 0.82 \\
\midrule
\multirow{2}{*}{\makecell[l]{Zero-Shot +\\Task Description}} & \xmark & 0.91 & 0.95 & 0.73 & 0.77 & 0.86 & 0.72 & 0.83 & 0.65 & 0.82 & 0.87 & 0.86 & 0.67 & 0.94 & 0.81 & 0.68 & 0.77 & 0.84 & 0.84 & 0.80& 0.89 & 0.81 \\
& \checkmark & 0.93 & 0.99 & 0.90 & 0.86 & 0.92 & 0.87 & 0.92 & 0.79 & 0.92 & 0.91 & 0.95 & 0.92 & 0.94 & 0.89 & 0.82 & 0.85 & 0.95 & 0.93 & 0.88 & 0.95 & 0.90 \\
\midrule
\multirow{2}{*}{\makecell[l]{Few-Shot +\\Task Description}} & \xmark & 0.86 & 0.91 & 0.71 & 0.62 & 0.81 & 0.57 & 0.87 & 0.68 & 0.81 & 0.75 & 0.85 & 0.59 & 0.92 & 0.68 & 0.62 & 0.67 & 0.82 & 0.73 & 0.75 & 0.79 & 0.75 \\
& \checkmark & 0.97 & 0.99 & 0.91 & 0.91 & 0.96 & 0.88 & 0.93 & 0.83 & 0.97 & 0.93 & 0.96 & 0.92 & 0.95 & 0.95 & 0.87 & 0.86 & 0.95 & 0.93 & 0.94 & 0.95 & 0.93 \\
\bottomrule
\end{tabular}}
\caption{TNR using English for the instruction with (\checkmark) and without (\xmark) thinking mode for the \texttt{Qwen3 8B} in amonolingual setting.}
\end{table*}

\begin{table*}[]
\resizebox{\textwidth}{!}{%
\begin{tabular}{lcrrrrrrrrrrrrrrrrrrrrr}
\toprule
\textbf{Technique} & \textbf{Thinking} & \textbf{ara} & \textbf{bul} & \textbf{ces} & \textbf{deu} & \textbf{ell} & \textbf{eng} & \textbf{fra} & \textbf{hbs} & \textbf{hin} & \textbf{hun} & \textbf{kor} & \textbf{msa} & \textbf{mya} & \textbf{nld} & \textbf{pol} & \textbf{por} & \textbf{ron} & \textbf{slk} & \textbf{spa} & \textbf{tha} & \textbf{Avg.} \\
\midrule
\multirow{2}{*}{\makecell[l]{Zero-Shot}} & \xmark & \cellcolor{lightcoral}{-0.01} & \cellcolor{lightcoral}{-0.10} & \cellcolor{lightgreen}{0.01} & \cellcolor{lightgreen}{0.17} & \cellcolor{lightcoral}{-0.10} & 0.00 & \cellcolor{lightgreen}{0.17} & \cellcolor{lightgreen}{0.11} & \cellcolor{lightgreen}{0.03} & \cellcolor{lightcoral}{-0.05} & \cellcolor{lightcoral}{-0.29} & \cellcolor{lightcoral}{-0.13} & \cellcolor{lightcoral}{-0.32} & \cellcolor{lightcoral}{-0.11} & \cellcolor{lightgreen}{0.02} & \cellcolor{lightgreen}{0.12} & \cellcolor{lightgreen}{0.04} & \cellcolor{lightgreen}{0.07} & \cellcolor{lightcoral}{-0.08} & \cellcolor{lightcoral}{-0.35} & \cellcolor{lightcoral}{-0.04} \\
& \checkmark & \cellcolor{lightgreen}{0.03} & 0.00 & \cellcolor{lightgreen}{0.03} & \cellcolor{lightgreen}{0.09} & \cellcolor{lightcoral}{-0.07} & 0.00 & \cellcolor{lightgreen}{0.01} & \cellcolor{lightgreen}{0.02} & \cellcolor{lightgreen}{0.03} & \cellcolor{lightgreen}{0.03} & 0.00 & \cellcolor{lightcoral}{-0.07} & \cellcolor{lightcoral}{-0.28} & \cellcolor{lightcoral}{-0.02} & \cellcolor{lightcoral}{-0.03} & \cellcolor{lightcoral}{-0.01} & \cellcolor{lightcoral}{-0.01} & \cellcolor{lightgreen}{0.05} & \cellcolor{lightcoral}{-0.05} & \cellcolor{lightcoral}{-0.05} & \cellcolor{lightcoral}{-0.02} \\
\midrule
\multirow{2}{*}{\makecell[l]{Zero-Shot +\\Task Description}} & \xmark & \cellcolor{lightcoral}{-0.02} & 0.00 & \cellcolor{lightgreen}{0.05} & \cellcolor{lightgreen}{0.14} & \cellcolor{lightcoral}{-0.04} & 0.00 & \cellcolor{lightgreen}{0.06} & \cellcolor{lightgreen}{0.03} & \cellcolor{lightcoral}{-0.05} & \cellcolor{lightcoral}{-0.06} & \cellcolor{lightcoral}{-0.09} & \cellcolor{lightcoral}{-0.04} & \cellcolor{lightcoral}{-0.18} & \cellcolor{lightgreen}{0.10} & \cellcolor{lightgreen}{0.09} & \cellcolor{lightcoral}{-0.03} & \cellcolor{lightgreen}{0.02} & \cellcolor{lightcoral}{-0.01} & \cellcolor{lightgreen}{0.09} & \cellcolor{lightcoral}{-0.08} & 0.00 \\
& \checkmark & \cellcolor{lightgreen}{0.03} & 0.00 & \cellcolor{lightcoral}{-0.01} & \cellcolor{lightgreen}{0.05} & \cellcolor{lightcoral}{-0.01} & 0.00 & \cellcolor{lightcoral}{-0.01} & \cellcolor{lightcoral}{-0.01} & \cellcolor{lightgreen}{0.01} & 0.00 & \cellcolor{lightcoral}{-0.01} & \cellcolor{lightcoral}{-0.01} & \cellcolor{lightcoral}{-0.17} & \cellcolor{lightgreen}{0.03} & \cellcolor{lightgreen}{0.09} & \cellcolor{lightcoral}{-0.04} & \cellcolor{lightcoral}{-0.03} & 0.00 & \cellcolor{lightgreen}{0.02} & \cellcolor{lightgreen}{0.01} & 0.00 \\
\midrule
\multirow{2}{*}{\makecell[l]{Few-Shot +\\Task Description}} & \xmark & \cellcolor{lightgreen}{0.03} & \cellcolor{lightgreen}{0.05} & \cellcolor{lightcoral}{-0.08} & \cellcolor{lightgreen}{0.19} & \cellcolor{lightcoral}{-0.21} & 0.00 & \cellcolor{lightgreen}{0.08} & \cellcolor{lightgreen}{0.09} & \cellcolor{lightgreen}{0.11} & \cellcolor{lightgreen}{0.09} & \cellcolor{lightcoral}{-0.10} & \cellcolor{lightgreen}{0.30} & \cellcolor{lightcoral}{-0.36} & \cellcolor{lightgreen}{0.22} & \cellcolor{lightgreen}{0.01} & \cellcolor{lightgreen}{0.11} & \cellcolor{lightgreen}{0.11} & \cellcolor{lightcoral}{-0.02} & \cellcolor{lightgreen}{0.16} & \cellcolor{lightgreen}{0.15} & \cellcolor{lightgreen}{0.05} \\
 & \checkmark & \cellcolor{lightgreen}{0.01} & \cellcolor{lightgreen}{0.01} & 0.00 & \cellcolor{lightgreen}{0.02} & 0.00 & 0.00 & \cellcolor{lightcoral}{-0.01} & \cellcolor{lightgreen}{0.01} & \cellcolor{lightgreen}{0.01} & \cellcolor{lightgreen}{0.04} & \cellcolor{lightcoral}{-0.01} & \cellcolor{lightgreen}{0.01} & \cellcolor{lightcoral}{-0.15} & \cellcolor{lightgreen}{0.01} & \cellcolor{lightgreen}{0.04} & \cellcolor{lightgreen}{0.04} & \cellcolor{lightcoral}{-0.01} & \cellcolor{lightgreen}{0.01} & \cellcolor{lightgreen}{0.03} & \cellcolor{lightcoral}{-0.01} & 0.00 \\
\bottomrule
\end{tabular}}
\setlength{\fboxsep}{1pt}
\caption{Impact of the thinking mode on TNR in monolingual settings across three prompting techniques for the \texttt{Qwen3 8B}. The table shows the difference in TNR when using target language instructions versus English. Each row compares performance with (\checkmark) and without (\xmark) "thinking mode", across 20 languages. Positive values (\colorbox{lightgreen}{green}) indicate improved performance with target language instructions. Average scores are reported in the final column.}
\label{tab:my_label}
\end{table*}

\begin{table*}
\resizebox{\textwidth}{!}{%
\begin{tabular}{lcrrrrrrrrrrrrrrrrrrrrr}
\toprule
\textbf{Technique} & \textbf{Thinking} & \textbf{spa-eng} & \textbf{hin-eng} & \textbf{eng-ara} & \textbf{fra-eng} & \textbf{deu-eng} & \textbf{eng-por} & \textbf{spa-por} & \textbf{deu-fra} & \textbf{slk-ces} & \textbf{slk-eng} & \textbf{pol-hbs} & \textbf{ces-eng} & \textbf{ces-pol} & \textbf{nld-deu} & \textbf{msa-ara} & \textbf{kor-eng} & \textbf{mya-msa} & \textbf{ara-fra} & \textbf{hun-pol} & \textbf{tha-por} & \textbf{Avg.} \\
\midrule
\multirow{2}{*}{\makecell[l]{Zero-Shot}} & \xmark & 0.78 & 0.88 & 0.89 & 0.70& 0.71 & 0.85 & 0.90 & 0.68 & 0.70& 0.60& 0.70& 0.69 & 0.59 & 0.68 & 0.77 & 0.84 & 0.76 & 0.93 & 0.81 & 0.84 & 0.76 \\
& \checkmark & 0.84 & 0.88 & 0.89 & 0.79 & 0.76 & 0.89 & 0.88 & 0.70& 0.81 & 0.72 & 0.74 & 0.73 & 0.70& 0.72 & 0.86 & 0.79 & 0.95 & 0.94 & 0.79 & 0.86 & 0.81 \\
\midrule
\multirow{2}{*}{\makecell[l]{Zero-Shot +\\Task Description}} & \xmark & 0.86 & 0.96 & 0.96 & 0.80& 0.79 & 0.89 & 0.95 & 0.83 & 0.71 & 0.66 & 0.73 & 0.66 & 0.76 & 0.81 & 0.91 & 0.81 & 0.98 & 0.97 & 0.87 & 0.87 & 0.84 \\
& \checkmark & 0.91 & 0.97 & 0.96 & 0.89 & 0.86 & 0.95 & 0.95 & 0.87 & 0.88 & 0.85 & 0.85 & 0.82 & 0.84 & 0.86 & 0.94 & 0.86 & 0.98 & 0.97 & 0.88 & 0.92 & 0.90 \\
\midrule
\multirow{2}{*}{\makecell[l]{Few-Shot +\\Task Description}} & \xmark & 0.73 & 0.92 & 0.97 & 0.71 & 0.67 & 0.78 & 0.91 & 0.77 & 0.76 & 0.58 & 0.65 & 0.60& 0.80& 0.70& 0.88 & 0.71 & 1.00 & 0.98 & 0.80& 0.75 & 0.78 \\
& \checkmark & 0.92 & 0.98 & 0.97 & 0.90 & 0.88 & 0.95 & 0.97 & 0.91 & 0.92 & 0.83 & 0.86 & 0.83 & 0.88 & 0.92 & 0.96 & 0.88 & 0.96 & 0.98 & 0.92 & 0.90 & 0.91 \\
\bottomrule
\end{tabular}}
\caption{TNR using English for the instruction with (\checkmark) and without (\xmark) thinking mode for the \texttt{Qwen3 8B} in a cross-lingual setting.}
\end{table*}

\begin{table*}
\resizebox{\textwidth}{!}{%
\begin{tabular}{lcrrrrrrrrrrrrrrrrrrrrr}
\toprule
\textbf{Technique} & \textbf{Thinking} & \textbf{spa-eng} & \textbf{hin-eng} & \textbf{eng-ara} & \textbf{fra-eng} & \textbf{deu-eng} & \textbf{eng-por} & \textbf{spa-por} & \textbf{deu-fra} & \textbf{slk-ces} & \textbf{slk-eng} & \textbf{pol-hbs} & \textbf{ces-eng} & \textbf{ces-pol} & \textbf{nld-deu} & \textbf{msa-ara} & \textbf{kor-eng} & \textbf{mya-msa} & \textbf{ara-fra} & \textbf{hun-pol} & \textbf{tha-por} & \textbf{Avg.} \\
\midrule
\multirow{2}{*}{\makecell[l]{Zero-Shot}} & \xmark & \cellcolor{lightcoral}{-0.05} & \cellcolor{lightcoral}{-0.04} & 0.00 & \cellcolor{lightgreen}{0.21} & \cellcolor{lightgreen}{0.08} & 0.00 & \cellcolor{lightcoral}{-0.03} & \cellcolor{lightgreen}{0.11} & \cellcolor{lightgreen}{0.09} & \cellcolor{lightgreen}{0.04} & \cellcolor{lightcoral}{-0.04} & 0.00 & \cellcolor{lightgreen}{0.02} & \cellcolor{lightcoral}{-0.05} & \cellcolor{lightcoral}{-0.41} & \cellcolor{lightcoral}{-0.40} & \cellcolor{lightcoral}{-0.39} & \cellcolor{lightgreen}{0.03} & \cellcolor{lightcoral}{-0.13} & \cellcolor{lightcoral}{-0.46} & \cellcolor{lightcoral}{-0.07} \\
& \checkmark & \cellcolor{lightcoral}{-0.05} & \cellcolor{lightgreen}{0.04} & 0.00 & \cellcolor{lightcoral}{-0.03} & \cellcolor{lightgreen}{0.04} & 0.00 & \cellcolor{lightcoral}{-0.04} & \cellcolor{lightgreen}{0.04} & 0.00 & \cellcolor{lightcoral}{-0.02} & \cellcolor{lightcoral}{-0.02} & \cellcolor{lightcoral}{-0.05} & \cellcolor{lightgreen}{0.03} & \cellcolor{lightcoral}{-0.01} & \cellcolor{lightcoral}{-0.13} & 0.00 & \cellcolor{lightcoral}{-0.43} & \cellcolor{lightgreen}{0.01} & \cellcolor{lightcoral}{-0.03} & \cellcolor{lightcoral}{-0.14} & \cellcolor{lightcoral}{-0.04} \\
\midrule
\multirow{2}{*}{\makecell[l]{Zero-Shot +\\Task Description}} & \xmark & \cellcolor{lightgreen}{0.01} & 0.00 & 0.00 & \cellcolor{lightgreen}{0.02} & \cellcolor{lightgreen}{0.13} & 0.00 & \cellcolor{lightgreen}{0.03} & \cellcolor{lightgreen}{0.07} & \cellcolor{lightgreen}{0.01} & \cellcolor{lightcoral}{-0.02} & \cellcolor{lightgreen}{0.07} & \cellcolor{lightgreen}{0.02} & \cellcolor{lightgreen}{0.05} & \cellcolor{lightgreen}{0.11} & \cellcolor{lightcoral}{-0.14} & \cellcolor{lightcoral}{-0.10} & \cellcolor{lightcoral}{-0.14} & \cellcolor{lightcoral}{-0.01} & \cellcolor{lightcoral}{-0.16} & \cellcolor{lightcoral}{-0.14} & \cellcolor{lightcoral}{-0.01} \\
 & \checkmark  & \cellcolor{lightcoral}{-0.02} & \cellcolor{lightgreen}{0.01} & 0.00 & \cellcolor{lightcoral}{-0.05} & \cellcolor{lightgreen}{0.01} & 0.00 & 0.00 & \cellcolor{lightgreen}{0.01} & \cellcolor{lightgreen}{0.03} & \cellcolor{lightcoral}{-0.05} & \cellcolor{lightgreen}{0.04} & \cellcolor{lightcoral}{-0.04} & \cellcolor{lightcoral}{-0.02} & \cellcolor{lightgreen}{0.05} & \cellcolor{lightcoral}{-0.05} & \cellcolor{lightcoral}{-0.08} & \cellcolor{lightcoral}{-0.07} & \cellcolor{lightgreen}{0.01} & \cellcolor{lightcoral}{-0.05} & \cellcolor{lightgreen}{0.01} & \cellcolor{lightcoral}{-0.01} \\
\midrule
\multirow{2}{*}{\makecell[l]{Few-Shot +\\Task Description}} & \xmark & \cellcolor{lightgreen}{0.10} & \cellcolor{lightgreen}{0.08} & 0.00 & \cellcolor{lightgreen}{0.17} & \cellcolor{lightgreen}{0.14} & 0.00 & \cellcolor{lightgreen}{0.05} & \cellcolor{lightgreen}{0.13} & \cellcolor{lightcoral}{-0.08} & \cellcolor{lightcoral}{-0.08} & \cellcolor{lightgreen}{0.13} & \cellcolor{lightcoral}{-0.12} & \cellcolor{lightcoral}{-0.16} & \cellcolor{lightgreen}{0.19} & \cellcolor{lightgreen}{0.07} & \cellcolor{lightcoral}{-0.19} & \cellcolor{lightcoral}{-0.41} & \cellcolor{lightcoral}{-0.02} & \cellcolor{lightgreen}{0.03} & \cellcolor{lightgreen}{0.16} & \cellcolor{lightgreen}{0.01} \\
 & \checkmark  & \cellcolor{lightgreen}{0.01} & \cellcolor{lightgreen}{0.01} & 0.00 & \cellcolor{lightgreen}{0.01} & \cellcolor{lightgreen}{0.01} & 0.00 & \cellcolor{lightgreen}{0.02} & \cellcolor{lightgreen}{0.03} & 0.00 & \cellcolor{lightcoral}{-0.01} & \cellcolor{lightgreen}{0.05} & \cellcolor{lightcoral}{-0.04} & \cellcolor{lightgreen}{0.03} & \cellcolor{lightgreen}{0.04} & \cellcolor{lightcoral}{-0.05} & \cellcolor{lightcoral}{-0.03} & \cellcolor{lightcoral}{-0.11} & \cellcolor{lightgreen}{0.02} & \cellcolor{lightgreen}{0.01} & \cellcolor{lightcoral}{-0.03} & 0.00 \\
\bottomrule
\end{tabular}}
\setlength{\fboxsep}{1pt}
\caption{Impact of the thinking mode on performance in cross-lingual settings across three prompting techniques for the \texttt{Qwen3 8B}. The table shows the difference in the TNR when using post-language instructions (the first language in the column name) versus English. Each row compares performance with (\checkmark) and without (\xmark) "thinking mode", across 20 language pairs. Positive values (\colorbox{lightgreen}{green}) indicate improved performance with post-language instructions. Average scores are reported in the final column.}
\end{table*}

\begin{table*}
\resizebox{\textwidth}{!}{%
\begin{tabular}{lcrrrrrrrrrrrrrrrrrrrrr}
\toprule
\textbf{Technique} & \textbf{Thinking} & \textbf{spa-eng} & \textbf{hin-eng} & \textbf{eng-ara} & \textbf{fra-eng} & \textbf{deu-eng} & \textbf{eng-por} & \textbf{spa-por} & \textbf{deu-fra} & \textbf{slk-ces} & \textbf{slk-eng} & \textbf{pol-hbs} & \textbf{ces-eng} & \textbf{ces-pol} & \textbf{nld-deu} & \textbf{msa-ara} & \textbf{kor-eng} & \textbf{mya-msa} & \textbf{ara-fra} & \textbf{hun-pol} & \textbf{tha-por} & \textbf{Avg.} \\
\midrule
\multirow{2}{*}{\makecell[l]{Zero-Shot}} & \xmark & 0.00 & 0.00 & \cellcolor{lightcoral}{-0.02} & 0.00 & 0.00 & \cellcolor{lightcoral}{-0.03} & \cellcolor{lightgreen}{0.03} & \cellcolor{lightgreen}{0.24} & \cellcolor{lightgreen}{0.03} & 0.00 & \cellcolor{lightcoral}{-0.04} & 0.00 & 0.00 & \cellcolor{lightgreen}{0.12} & \cellcolor{lightcoral}{-0.04} & 0.00 & \cellcolor{lightcoral}{-0.32} & \cellcolor{lightgreen}{0.04} & \cellcolor{lightcoral}{-0.12} & \cellcolor{lightcoral}{-0.08} & \cellcolor{lightcoral}{-0.01} \\
& \checkmark & 0.00 & 0.00 & \cellcolor{lightgreen}{0.01} & 0.00 & 0.00 & \cellcolor{lightcoral}{-0.03} & \cellcolor{lightcoral}{-0.01} & \cellcolor{lightgreen}{0.02} & \cellcolor{lightgreen}{0.01} & 0.00 & \cellcolor{lightcoral}{-0.02} & 0.00 & \cellcolor{lightcoral}{-0.02} & \cellcolor{lightgreen}{0.05} & 0.00 & 0.00 & \cellcolor{lightcoral}{-0.08} & \cellcolor{lightcoral}{-0.01} & 0.00 & \cellcolor{lightcoral}{-0.04} & \cellcolor{lightcoral}{-0.01} \\
\midrule
\multirow{2}{*}{\makecell[l]{Zero-Shot +\\Task Description}} & \xmark & 0.00 & 0.00 & \cellcolor{lightcoral}{-0.05} & 0.00 & 0.00 & \cellcolor{lightcoral}{-0.07} & \cellcolor{lightcoral}{-0.02} & \cellcolor{lightcoral}{-0.02} & \cellcolor{lightgreen}{0.08} & 0.00 & \cellcolor{lightgreen}{0.01} & 0.00 & \cellcolor{lightgreen}{0.03} & \cellcolor{lightgreen}{0.09} & \cellcolor{lightcoral}{-0.06} & 0.00 & \cellcolor{lightcoral}{-0.14} & \cellcolor{lightcoral}{-0.02} & \cellcolor{lightgreen}{0.01} & \cellcolor{lightcoral}{-0.10} & \cellcolor{lightcoral}{-0.01} \\
 & \checkmark  & 0.00 & 0.00 & \cellcolor{lightcoral}{-0.01} & 0.00 & 0.00 & \cellcolor{lightcoral}{-0.04} & \cellcolor{lightcoral}{-0.03} & \cellcolor{lightcoral}{-0.03} & \cellcolor{lightcoral}{-0.02} & 0.00 & \cellcolor{lightcoral}{-0.04} & 0.00 & \cellcolor{lightgreen}{0.06} & \cellcolor{lightgreen}{0.03} & \cellcolor{lightgreen}{0.01} & 0.00 & \cellcolor{lightcoral}{-0.02} & 0.00 & \cellcolor{lightgreen}{0.04} & \cellcolor{lightcoral}{-0.01} & 0.00 \\
\midrule
\multirow{2}{*}{\makecell[l]{Few-Shot +\\Task Description}} & \xmark & 0.00 & 0.00 & \cellcolor{lightgreen}{0.02} & 0.00 & 0.00 & \cellcolor{lightgreen}{0.13} & \cellcolor{lightgreen}{0.03} & \cellcolor{lightgreen}{0.15} & \cellcolor{lightcoral}{-0.19} & 0.00 & \cellcolor{lightgreen}{0.15} & 0.00 & \cellcolor{lightcoral}{-0.03} & \cellcolor{lightgreen}{0.13} & \cellcolor{lightcoral}{-0.05} & 0.00 & 0.00 & \cellcolor{lightgreen}{0.01} & 0.00 & \cellcolor{lightgreen}{0.18} & \cellcolor{lightgreen}{0.03} \\
& \checkmark  & 0.00 & 0.00 & \cellcolor{lightgreen}{0.02} & 0.00 & 0.00 & \cellcolor{lightgreen}{0.03} & \cellcolor{lightgreen}{0.02} & \cellcolor{lightgreen}{0.03} & \cellcolor{lightcoral}{-0.04} & 0.00 & \cellcolor{lightgreen}{0.02} & 0.00 & \cellcolor{lightgreen}{0.03} & \cellcolor{lightgreen}{0.01} & 0.00 & 0.00 & \cellcolor{lightcoral}{-0.01} & \cellcolor{lightgreen}{0.01} & \cellcolor{lightgreen}{0.04} & \cellcolor{lightgreen}{0.02} & \cellcolor{lightgreen}{0.01} \\
\bottomrule
\end{tabular}}
\caption{Impact of the thinking mode on performance in cross-lingual settings across three prompting techniques for the \texttt{Qwen3 8B}. The table shows the difference in the TNR when using claim-language instructions (the second language in the column name) versus English. Each row compares performance with (\checkmark) and without (\xmark) "thinking mode", across 20 language pairs. Positive values (\colorbox{lightgreen}{green}) indicate improved performance with claim-language instructions. Average scores are reported in the final column.}
\end{table*}

\section{Retrieval Bias}

Table~\ref{tab:top20claims} shows the most frequently retrieved claims across the top-20 results for the \texttt{Multilingual E5} model. Unlike the top-1 results (Table~\ref{tab:top1_claims}), this broader view reveals a heavier presence of noisy or structurally non-claim-like content. For example, the most frequent entry is a JavaScript snippet retrieved in 159 different cases. Similarly, raw URLs—including Facebook, Instagram, and Google Drive links—appear multiple times with high retrieval frequency. These items generally lack informative or verifiable content and suggest that the model may favor certain boilerplate or template-like artifacts during dense retrieval. This pattern highlights a retrieval bias toward recurring surface forms, particularly in the absence of strong semantic grounding. The average similarity scores for these claims are relatively high, indicating that the model considers them relevant despite their limited factual content.

\begin{table}[h]
\centering
\small
\begin{tabular}{p{4cm} r r}
\toprule
\textbf{Claim (truncated)} & \textbf{Freq} & \textbf{CosSim} \\
\midrule
(function(d, s, id) \{ var js, fjs... & 159 & 0.778 \\
https://scontent.ftxl2-1.fna.fbcdn... & 64 & 0.780 \\
https://drive.google.com/file/d/1T... & 63 & 0.780 \\
https://www.instagram.com/tv/CTjM5... & 61 & 0.784 \\
1. All calls will be recorded... & 44 & 0.779 \\
https://www.facebook.com/photo.php... & 37 & 0.779 \\
The corona virus is large in size... & 33 & 0.816 \\
https://www.facebook.com/10007599... & 30 & 0.786 \\
Urgent information on Covid... & 26 & 0.802 \\
https://archive.ph/p0fQB & 25 & 0.785 \\
\bottomrule
\end{tabular}
\caption{Top-20 retrieved claims: truncated content with frequency (freq) and average cosine similarity score (CosSim).}
\label{tab:top20claims}
\end{table}

\begin{figure*}
    \centering
    \includegraphics[width=0.99\linewidth]{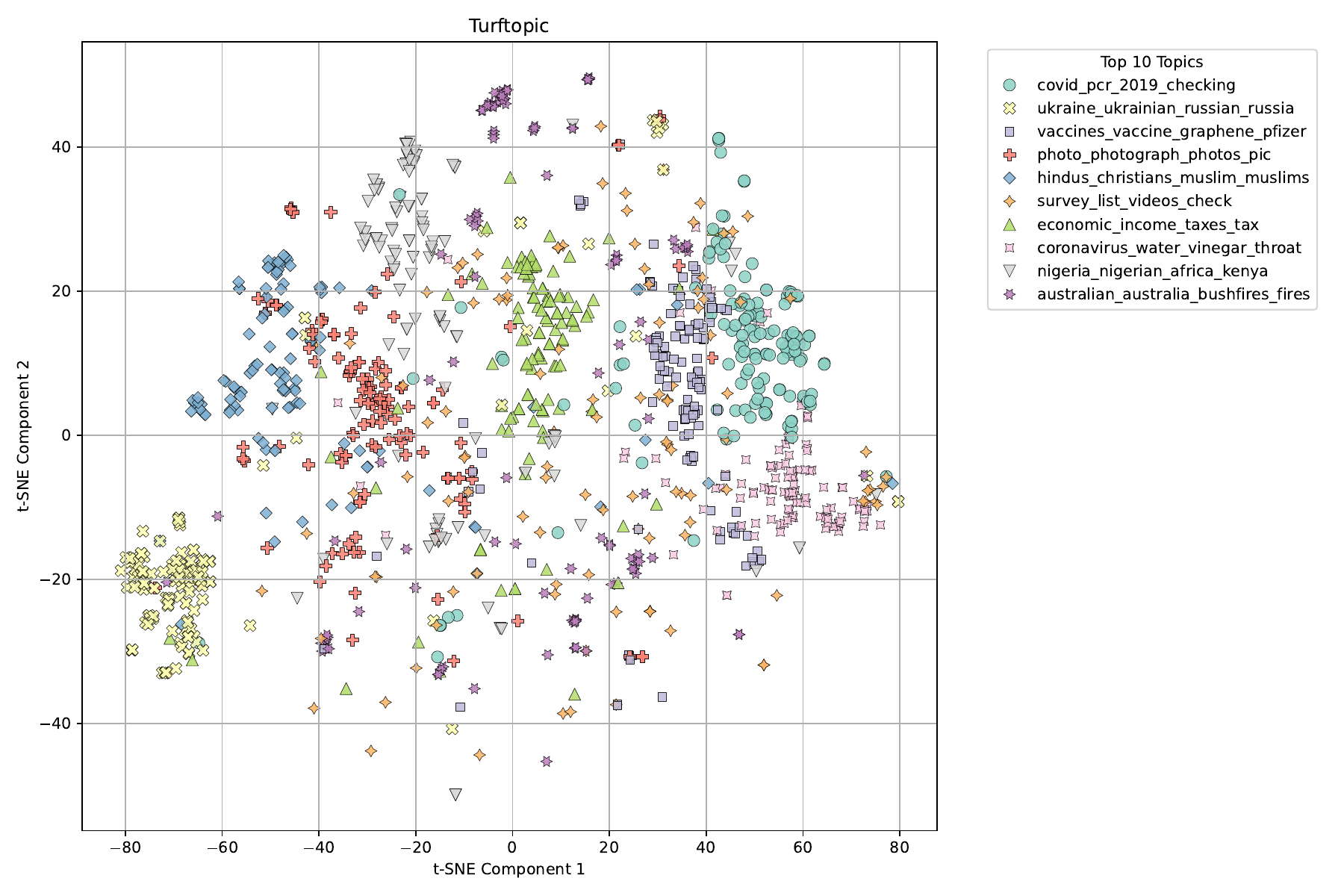}
\caption{t-SNE projection of fact-check claim embeddings colored by their assigned topics from \texttt{turftopic}. The plot shows the 10 most frequent topics in the corpus. Each point represents a fact-checked claim, and clusters correspond to semantically coherent topic groupings. Despite overlapping content dimensions, visually separable clusters reflect the model's ability to distinguish major thematic areas such as health, geopolitics, and misinformation trends.}
    \label{fig:topics}
\end{figure*}

\end{document}